\documentclass[10pt,twocolumn,letterpaper]{article}

\usepackage{iccv}
\usepackage{times}
\usepackage{epsfig}
\usepackage{graphicx}
\usepackage{amsmath}
\usepackage{amssymb}

\usepackage{balance}
\usepackage{bm} 
\usepackage{booktabs}
\usepackage{color}
\usepackage{dsfont}
\usepackage{lscape} 
\usepackage{nicefrac}
\usepackage[super]{nth} 
\usepackage{subfigure}
\usepackage{xspace}
\usepackage[table]{xcolor}    

\usepackage{tabularx}
\newcolumntype{Y}{>{\centering\arraybackslash}X}

\usepackage{nicefrac}

\usepackage{pifont}
\newcommand{\cmark}{\ding{51}}%
\newcommand{\xmark}{\ding{55}}%

\usepackage{etoolbox}
\usepackage[binary-units]{siunitx}
\robustify\bfseries
\definecolor{note-color}{RGB}{220,20,60}

\newcommand*{\red}[1]{\textcolor{red}{#1}}

\newcommand*{\rom}[1]{\expandafter\@slowromancap\romannumeral #1@}

\newcommand*{\finish}{
    {\small
    \balance
    \bibliographystyle{ieee}
    \bibliography{short,papers,external,macros,local}}
    \end{document}
}

\newcommand*{\myparagraph}[1]{\vspace{0.5em}\noindent\textbf{#1}}

\newcommand*{\verdict}{\vspace{0.25em}\noindent\textit{\underline{Verdict:}~}}

\DeclareMathOperator{\linear}{linear}

\DeclareMathOperator{\relu}{relu}
\DeclareMathOperator{\sigmoid}{sigmoid}
\DeclareMathOperator{\sigmoids}{sigmoids}

\DeclareMathOperator{\fanin}{fan\_in}
\DeclareMathOperator{\fanout}{fan\_out}
\DeclareMathOperator{\fanmax}{fan\_max}
\DeclareMathOperator{\msra}{msra}

\DeclareMathOperator{\sse}{\text{SSE}}

\definecolor{mygreen}{rgb}{0.2, 0.7, 0.1}

\usepackage[breaklinks=true,bookmarks=false,colorlinks=true,citecolor={mygreen},filecolor={mygreen}]{hyperref}

\usepackage[capitalize]{cleveref}
\crefname{section}{Sec.}{Section}

\iccvfinalcopy 

\def\iccvPaperID{***} 

\usepackage{setspace}
\usepackage{fancyhdr}

\begin{document}

\title{Deep Video Deblurring: The Devil is in the Details}

\author{Jochen Gast \hspace{1cm} Stefan Roth\\
Department of Computer Science, TU Darmstadt\vspace{1em}}

\maketitle

\thispagestyle{fancy} 

\newcommand{\change}{}

%
%
\begin{abstract}
Video deblurring for hand-held cameras is a challenging task, since the underlying blur is caused by both camera shake and object motion.
State-of-the-art deep networks exploit temporal information from neighboring frames, either by means of spatio-temporal transformers or by recurrent architectures.
In contrast to these involved models, we found that a simple baseline CNN can perform astonishingly well when particular care is taken \wrt the details of model and training procedure.
To that end, we conduct a comprehensive study regarding these crucial details, uncovering extreme differences in quantitative and qualitative performance.
Exploiting these details allows us to boost the architecture and training procedure of a simple baseline CNN by a staggering $3.15$\emph{dB}, such that it becomes highly competitive \wrt cutting-edge networks.
This raises the question whether the reported accuracy difference between models is always due to technical contributions or also subject to such orthogonal, but crucial details.
\end{abstract}

\section{Introduction}
\label{sec:introduction}
Blind image deblurring -- the recovery of a sharp image given a blurry one -- has been studied extensively \cite{Krishnan:2011:BDN,Levin:2006:BMD,Perrone:2014:TVB,Shan:2008:HQM,Sun:2013:EBK,Xu:2010:TPK,Xu:2013:ULS}.
However, more recently and perhaps with the increasing popularity of hand-held video cameras, attention has shifted towards deblurring videos \cite{Nah:2017:DMC,Su:2017:DVD}.
With the (re-)emergence of deep learning and the availability of large amounts of data, the best performing methods today are usually discriminatively trained CNNs
\cite{Chen:2018:R2D}, RNNs \cite{Tao:2018:SRN}, or a mixture thereof \cite{Kim:2017:OVD,Kim:2018:STT}.
While the ``zoo'' of video deblurring models differs quite significantly, explanations as to why one network works better than another often remain at an unsatisfactory level.
While the performance of state-of-the-art video deblurring methods is usually validated by training \textit{within}-paper models under the same conditions, the specifics of the training settings \textit{between} papers remain rather different.

In this work we show that some of these seemingly small details in the model setup and training procedure add up to astonishing quantitative and visual differences.
In fact, our quantitative evaluation raises the question whether the benefit for some state-of-the-art models comes from the proposed architectures or perhaps the setup details.
This mirrors observations in other areas of computer vision, where the significance of choosing the right training setup is crucial to achieve highly competitive models \cite{Sun:2019:MMT}.
Henceforth, we conduct a study on how the model setup and training details of a comparatively simple baseline CNN drastically influence the resulting image quality in video deblurring.
By finding the right settings, we unlock a significant amount of hidden power of this baseline, and achieve state-of-the art results on popular benchmarks.

Our systematic analysis considers the following variations:
\emph{(1)} We investigate the use of $\linear$ output layers instead of the typical $\sigmoids$ and consider different initialization methods.
Our new $\fanmax$ initialization combined with $\linear$ outputs already yields a substantial $2$dB benefit over our $\sigmoid$ baseline.
\emph{(2)} While recent work proposes to deblur in YCbCr color space \cite{Zhang:2019:ASL}, we show that there is no significant benefit over RGB.
Instead, a simple extension of the training schedule can lead to an additional $0.4$dB benefit.
\emph{(3)} We uncover that both photometric augmentations as well as random image scaling in training hurt deblurring results due to the mismatch of training \vs test data statistics.
The misuse of augmentations can diminish the generalization performance up to a severe 0.44dB.
\emph{(4)} We explore the benefits of using optical flow networks for pre-warping the inputs, which yields another 0.4dB gain.
Concatenating pre-warped images to the inputs improves over a simple replacement of the temporal neighbors by up to $0.27$dB.
This is in contrast to previous work, which either claimed no benefit from using pre-warping \cite{Su:2017:DVD}, or applied a complex spatio-temporal subnetwork with additional trainable weights \cite{Kim:2018:STT}.
\emph{(5)} We explore the influence of training patch size and sequence length.
Longer sequences yield only a minor benefit, but large patch sizes significantly improve over small ones by up to $0.9$dB.
Taken together, we improve our baseline by a striking $3.15$dB and the published results of \cite{Su:2017:DVD} by $2.11$dB, reaching and even surpassing the quality of complex state-of-the-art networks on standard datasets.

\section{Related Work}
\label{sec:relatedwork}
\paragraph{Classic uniform and non-uniform deblurring.}
Classic uniform deblurring methods that restore a sharp image under the assumption of a single blur kernel usually enforce
sparse image statistics, and are often combined with probabilistic, variational frameworks \cite{Cho:2009:FMD,Fergus:2006:RCS,Levin:2009:UEB,Levin:2011:EML,Miskin:2000:ELB}.
Less common approaches include the use of self-similarity \cite{Michaeli:2014:BDI}, discriminatively trained regression tree fields \cite{Schelten:2015:IRT}, a dark-channel prior \cite{Pan:2016:BID}, or scale normalization \cite{Jin:2018:NBD}.

Moving objects or a moving camera, on the other hand, significantly complicate deblurring, since the motion varies across the image domain.
Here, usually restricting assumptions are enforced on the generative blur, either in the form of a candidate blur model \cite{Couzinie:2013:LER,Kim:2013:DSD}, a linear blur model \cite{Gast:2016:POM,Kim:2014:SFD}, or a more generic blur basis \cite{Gupta:2010:SID,Hirsch:2010:FRN,Whyte:2010:NDS,Zheng:2013:FMD}.

\myparagraph{Classic video deblurring.}
Early work on video blurring \cite{Cho:2012:VDH,Matsushita:2006:FVS} proposes to transfer sharp pixels from neighboring frames to the central reference frame.
While Matsushita~\etal \cite{Matsushita:2006:FVS} apply a global homography, Cho~\etal \cite{Cho:2012:VDH} improve on this by local patch search.
Overall, the averaging nature of these approaches tends to overly smooth results \cite{Delbracio:2015:HVD}.
Delbracio~\etal \cite{Delbracio:2015:HVD} overcome this via a weighted average in the Fourier domain, but rely on a registration of neighboring frames, which may fail for large blurs.
Kim~\etal \cite{Kim:2015:GVD} propose an energy-based approach to jointly estimate optical flow along a latent sharp image using piece-wise linear blur kernels.
Later, Ren~\etal \cite{Ren:2017:VDS} incorporate semantic segmentation into the energy.
Both approaches rely on primal-dual optimization, which is computationally demanding.

\myparagraph{Deep image deblurring.}
Among the first deblurring methods in the light of the recent renaissance of deep learning has been the work by Sun~\etal~\cite{Sun:2015:LCN} who train a CNN to predict pixelwise candidate blur kernels.
Later, Gong~\etal \cite{Gong:2017:FMB} extend this from image patches to a fully convolutional approach.
Chakrabarti~\cite{Chakrabarti:2016:ANA} tackles uniform deblurring in the frequency domain by predicting Fourier coefficients of patch-wise deconvolution filters.
Note that, as in the classical case, all aforementioned methods are still followed by a standard non-blind deconvolution pipeline.
This restriction is lifted by Schuler~\etal \cite{Schuler:2016:LTD} who replace both the kernel and image estimator module of classic pipelines by neural network blocks, respectively.
Noroozi~\etal \cite{Noroozi:2017:MDW} propose a multi-scale CNN, which directly regresses a sharp image from a blurry one.
Tao~\etal \cite{Tao:2018:SRN} suggest a scale-recurrent neural network (RNN) to solve the deblurring problem at multiple resolutions in conjunction with a multi-scale loss.

\myparagraph{Deep image deblurring via GANs.}
Other approaches draw from the recent progress on generative adversarial networks (GANs).
Ramakrishnan~\etal \cite{Ramakrishnan:2017:DGF} propose a GAN for recovering a sharp image from a given blurry one; the generator aims to output a visually plausible, sharp image, which fools the discriminator into thinking it comes from the true sharp image distribution.
Nah~\etal \cite{Nah:2017:DMC} propose a multi-scale CNN accompanied by an adversarial loss in order to mimic traditional course-to-fine deblurring techniques.
Similarly, Kupyn~\etal \cite{Kupyn:2018:DGB} apply a conditional GAN, where the content (or perceptual) loss is notably defined in the domain of CNN feature maps rather than output color space.
We do not consider the use of adversarial networks here, as we argue that the accuracy of feed-forward CNNs is not yet saturated on the deblurring task.
Note that despite the simplicity of our baseline, we outperform the model of \cite{Kupyn:2018:DGB} by a large margin, \cf \cref{sec:experiments}.

\myparagraph{Deep video deblurring.}
Deep learning approaches to video deblurring have yielded tremendous progress in speed and image quality.
Kim~\etal \cite{Kim:2017:OVD} focus on the temporal nature of the problem by applying a temporal feature blending layer within an RNN.
{\change
Similarly, Nah~\etal \cite{Nah:2019:RNN} apply an RNN to propagate intra-frame information.
}
While RNNs are promising, we note that these are often difficult to train in practice \cite{Pascanu:2013:OTT}.
We do not rely on a recurrent architecture, but a plain CNN, achieving very competitive results.
Zhang~\etal \cite{Zhang:2019:ASL} use spatio-temporal 3D convolutions in the early stages of a deep residual network.
Chen~\etal \cite{Chen:2018:R2D} extend \cite{Kupyn:2018:DGB} with a physics-based reblurring pipeline, which constructs a reblurred image from the sharp predictions using optical flow, and subsequently enforces consistency between the reblurred image and the blurry input image.
{\change
Wang~\etal \cite{Wang:2019:VRE} apply deformable convolutions along an attention module to tackle general video restoration tasks.
}

The DBN model of Su~\etal \cite{Su:2017:DVD} serves as baseline model in our study.
DBN is a simple encoder-decoder CNN with symmetric skip connections; its input is simply the concatenation of the temporal window of the video input sequence.
Later, Kim~\etal \cite{Kim:2018:STT} extend the DBN model by a 3D spatio-temporal transformer, which transforms the inputs to the reference frame.
Note that this requires training an additional subnetwork that finds 3D correspondences of the inputs to the reference frame.
We find that we can outperform \cite{Kim:2018:STT} based on the same backbone network without the need of a spatial transformer network.
More generally, we uncover crucial details in the model and training procedure, which strikingly boost the accuracy by several dB in PSNR, yielding a method that is highly competitive.

\begin{figure*}
\centering
\subfigure[Input]{%
\label{fig:ablation-outputinit-input}%
\begin{minipage}{4.0cm}
\centering
\includegraphics[height=2.2cm]{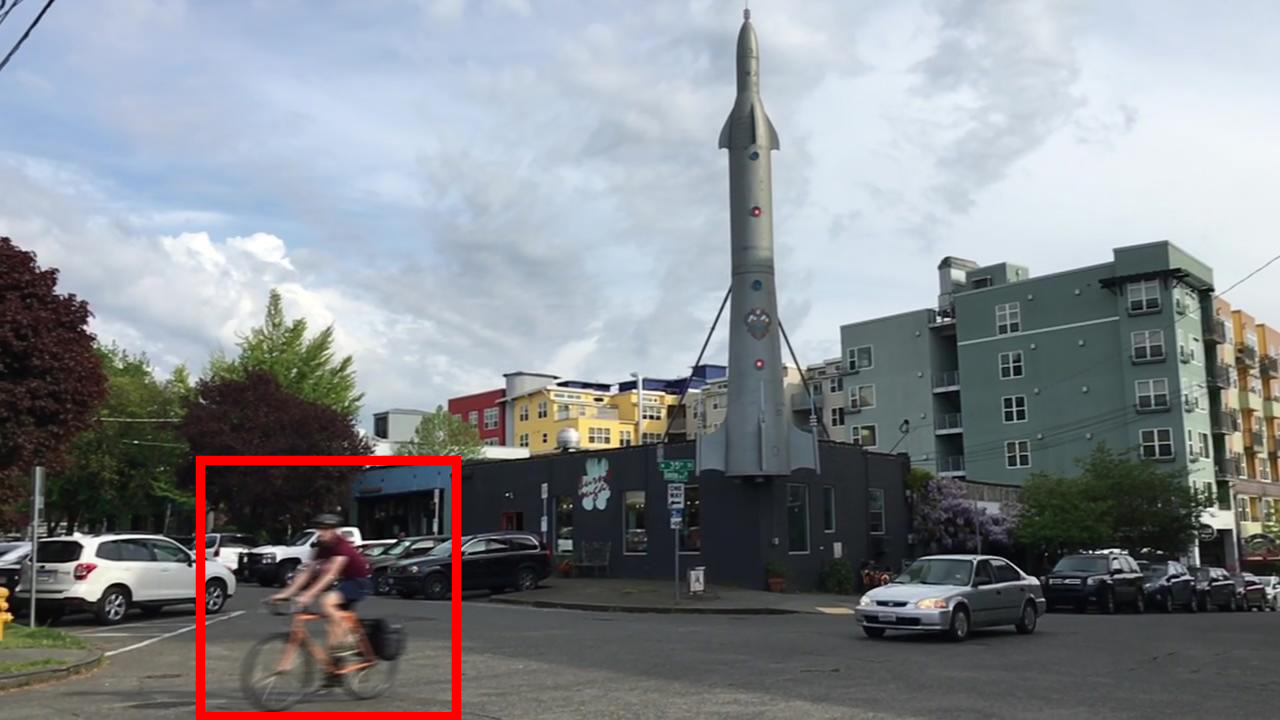}\\
\vspace{0.3em}
\end{minipage}}
%
%
\subfigure[lin+$\fanmax$]{
\label{fig:ablation-outputinit-linearfanmax}
\begin{minipage}{2.2cm}
\centering
\includegraphics[height=2.2cm]{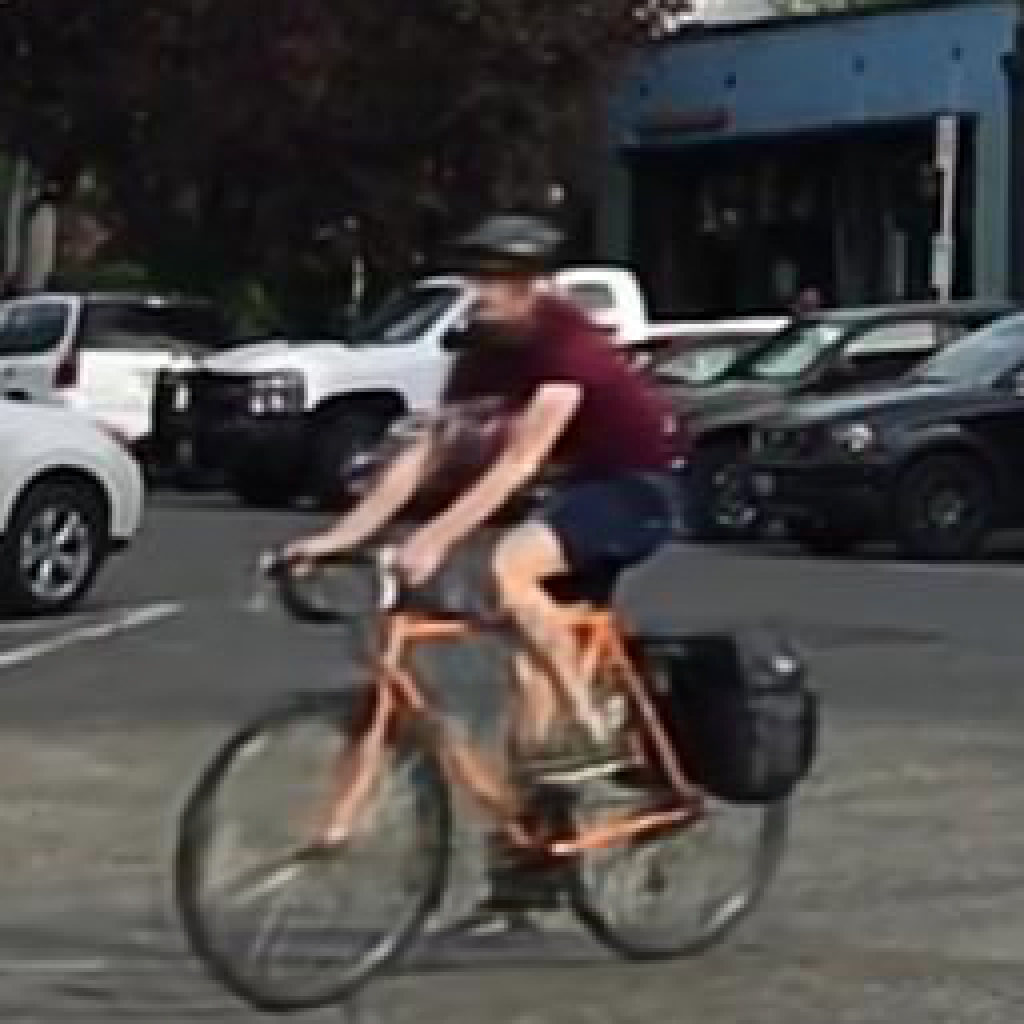}\\
\vspace{0.3em}
\end{minipage}}
\subfigure[sigm+$\fanmax$]{
\label{fig:ablation-outputinit-sigmoidfanmax}
\begin{minipage}{2.2cm}
\centering
\includegraphics[height=2.2cm]{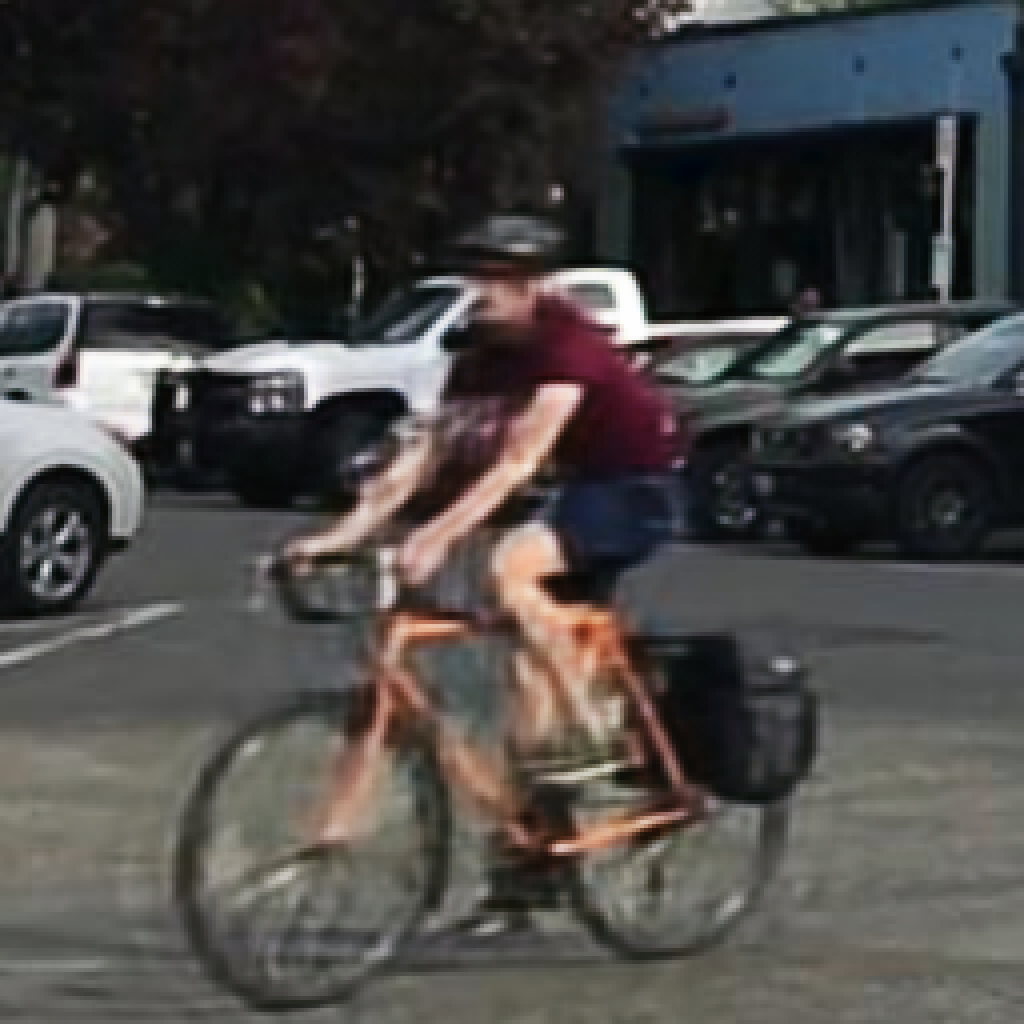} \\
\vspace{0.3em}
\end{minipage}}
\subfigure[sigm+$\fanin$]{
\label{fig:ablation-outputinit-sigmoidfanin}
\begin{minipage}{2.2cm}
\centering
\includegraphics[height=2.2cm]{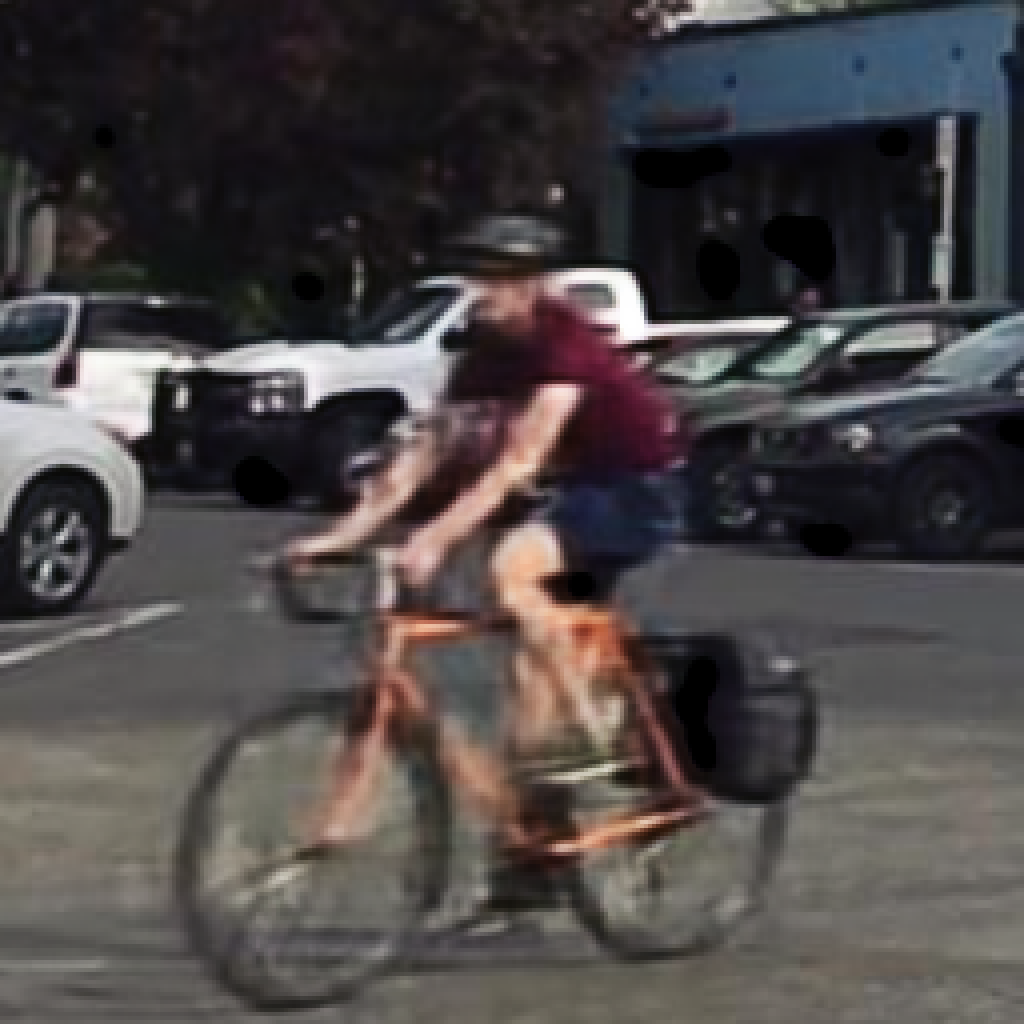}\\
\vspace{0.3em}
\end{minipage}}
\subfigure[sigm+$\fanout$]{
\label{fig:ablation-outputinit-sigmoidfanout}
\begin{minipage}{2.2cm}
\centering
\includegraphics[height=2.2cm]{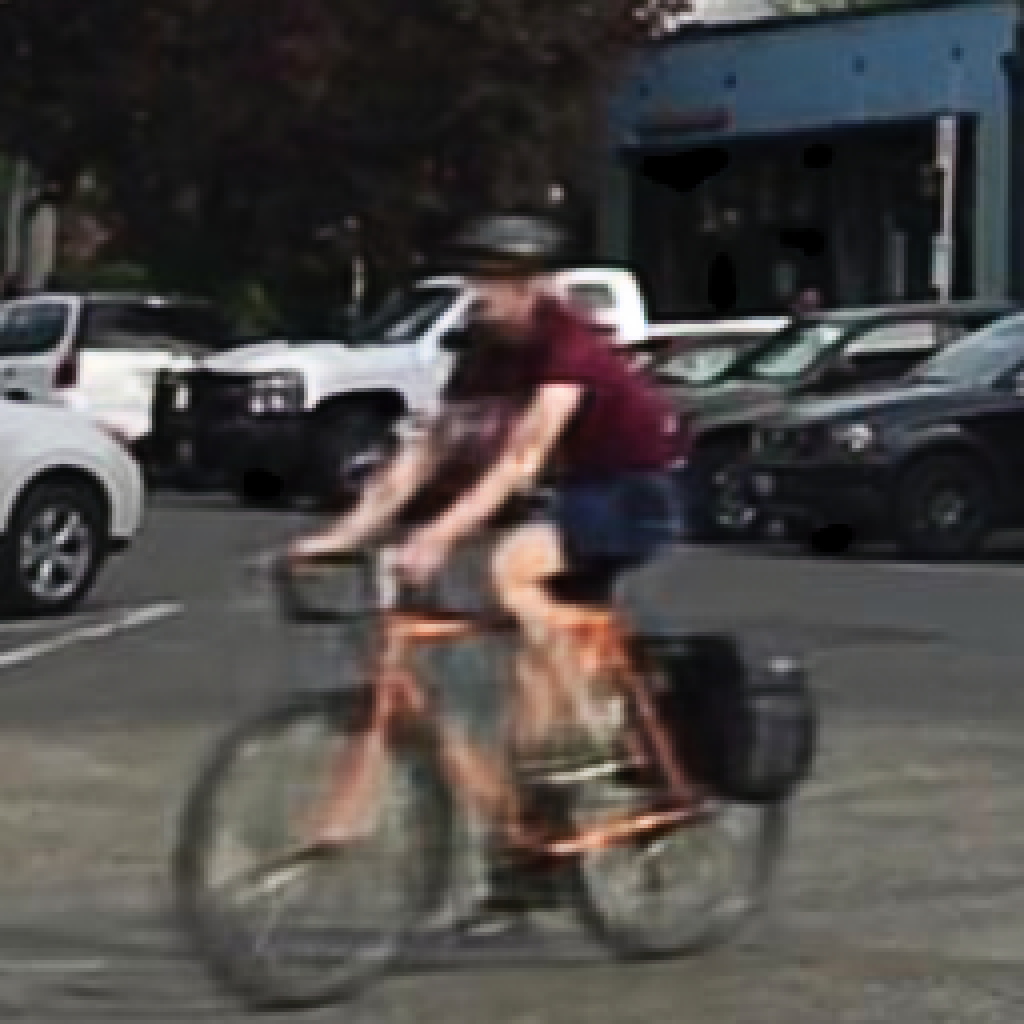}\\
\vspace{0.3em}
\end{minipage}}
%
%
%
\subfigure[gt]{
\label{fig:ablation-outputinit-gt}
\begin{minipage}{2.2cm}
\centering
\includegraphics[height=2.2cm]{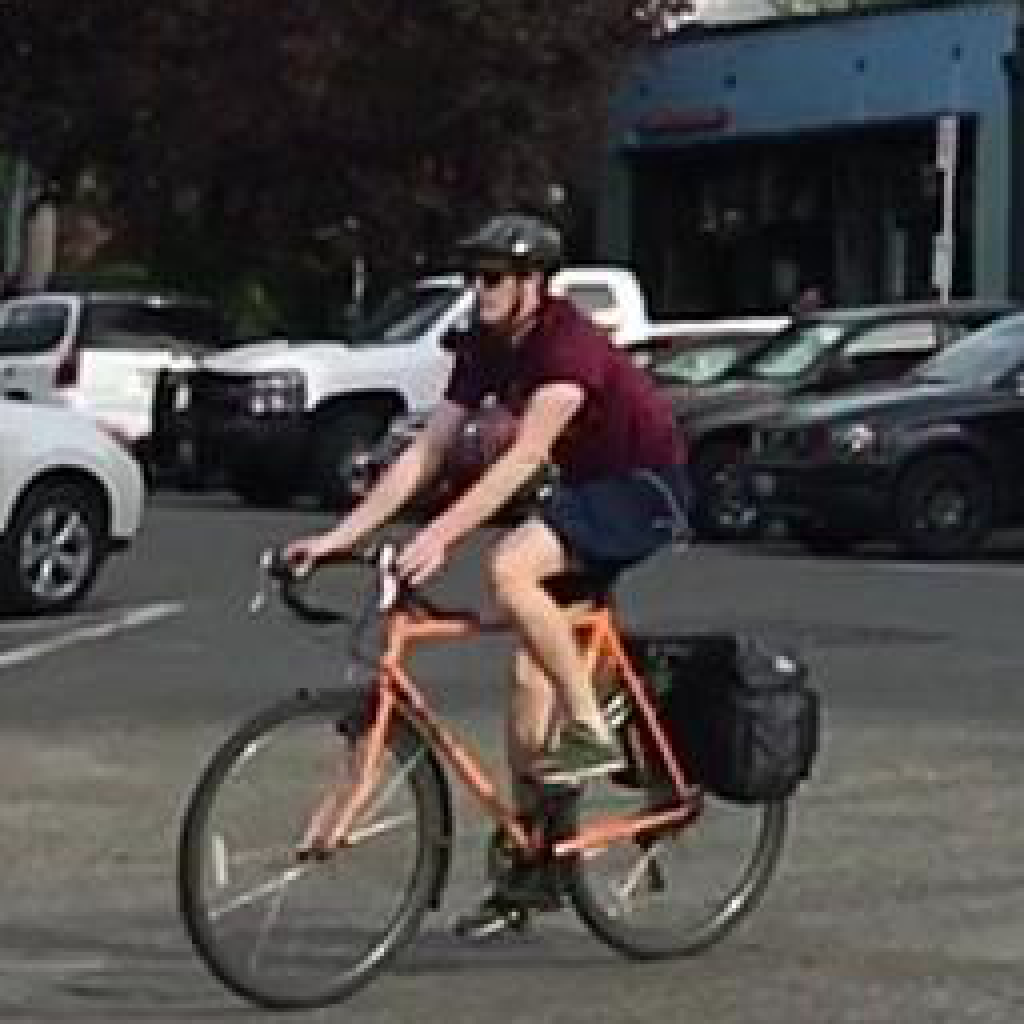}\\
\vspace{0.3em}
\end{minipage}}
\caption{{\bf Varying output layers and initializations.}
For the input \subref{fig:ablation-outputinit-input}, a $\linear$ output and $\fanmax$ initialization \subref{fig:ablation-outputinit-linearfanmax} visually yields better results than a $\sigmoid$ layer, independent of the fan-type used in the initialization.
Note the artifacts on the wheel in \subref{fig:ablation-outputinit-sigmoidfanmax} -- \subref{fig:ablation-outputinit-sigmoidfanout}.
}
\label{fig:ablation-outputinit}
\vspace{-0.5em}
\end{figure*}

\section{The Details of Deep Video Deblurring}
\label{sec:deep-video-deblurring}
As has been observed in papers in several areas of deep learning and beyond, careful choices of the architecture, (hyper-)parameters, training procedure, and more can significantly affect the final accuracy \cite{Chatfield:2014:RDD,Lucic:2018:AGC,Perrone:2014:TVB,Sun:2019:MMT}.
We show that the same holds true in deep video deblurring.
Specifically, we revisit the basic deep video deblurring network of Su~\etal \cite{Su:2017:DVD} and will uncover step-by-step, how choices made in mode, training, and preprocessing affect the deblurring accuracy.
All together, these details add up to a very significant $3.15$dB difference on the test dataset.

\subsection{Baseline network}
The basis architecture of our study is the DBN network of Su \etal~\cite{Su:2017:DVD} (\cf Table~1 therein), a fairly standard CNN with symmetric skip connections.
We closely follow the original training procedure in as far as it is specified in the paper \cite{Su:2017:DVD}.
Since we focus on details including the training procedure here, we first summarize the basic setup.
The baseline model and all subsequent refinements are trained on the 61 training sequences and tested on the 10 test sequences of the GOPRO dataset \cite{Su:2017:DVD}.
The sum of squared error (SSE) loss
is used for training and minimized with Adam \cite{Kingma:2015:AAM}, starting at a learning rate of $0.005$.
Following \cite{Su:2017:DVD}, the batch size is taken as 64 where we draw 8 random crops per example.
For all convolutional and transposed convolutional layers, 2D batch normalization \cite{Ioffe:2015:BNA} is applied and initialized with unit weights and zero biases.
While this simple architecture has led to competitive results when it was published in 2017, more recent methods \cite{Chen:2018:R2D,Kim:2018:STT} have strongly outperformed it.
In the following, we explore the potential to improve this baseline architecture and perform a step-by-step analysis.
\Cref{tab:dbn-ablation} gives an overview.

\begin{figure}[b]
\vspace{-0.5em}
\centering
\includegraphics[width=1\linewidth]{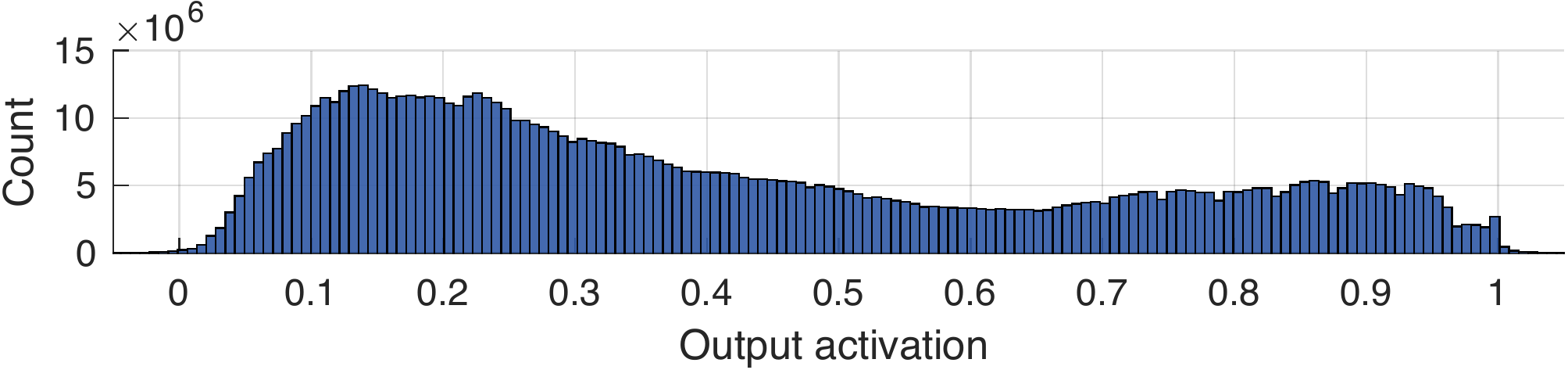}
\caption{%
{\bf Output activation statistics over test dataset.}
Even with linear outputs, the $\sse$ loss confines most activations to $[0,1]$.}
\label{fig:linear-activations}
\end{figure}

\subsection{Detail analysis}
\label{sec:baseline}

\paragraph{Output activation.}
The DBN network \cite{Su:2017:DVD} uses a $\sigmoid$ output layer to yield color values in the range $[0,1]$.
Given the limited range of pixel values in real digital images, this appears to be a prudent choice at first glance.
We question this, however, by recalling that $\sigmoid$ nonlinearities are a common root of optimization issues due to the well-known vanishing gradient problem.
We thus ask whether we need the $\sigmoid$ nonlinearity.

To that end, we replace it with a simple $\linear$ output.
As we can see in \cref{tab:dbn-ablation}\red{(a \vs d)}, this yields a very substantial 1dB accuracy benefit, highlighting again the importance of avoiding vanishing gradients.
In fact, the restriction to the unit range does not pose a significant problem even without output nonlinearity, since the SSE loss largely limits the linear outputs to the correct range anyway.
This is illustrated in \cref{fig:linear-activations}, which shows the linear activations on the test dataset after training with linear output activations under a $\sse$ loss; only very few values lie outside the valid color value range.
This can be easily addressed by clamping the outputs to $[0,1]$ at test time.

\myparagraph{Initialization.}
The choice of initialization is not discussed in \cite{Su:2017:DVD}.
However, as for any nonlinear optimization problem, initialization plays a crucial role.
Indeed, we find that good initialization is necessary to reproduce the results reported in \cite{Su:2017:DVD}.
Perhaps, the most popular initialization strategy for $\relu$-based neural networks today is the $\msra$ method of He~\etal \cite{Kaiming:2015:DDR}.
It ensures that under $\relu$ activations, the magnitudes of the input signal do not exponentially increase or decrease.
The $\msra$ initialization method typically comes in two variants, $\msra + \fanin$, and $\msra + \fanout$, depending on whether signal magnitudes should be preserved in the forward or backward pass.
In practice, $\fanin$ and $\fanout$ correspond to the number of gates connected to the inputs and outputs.
We additionally propose $\fanmax$, which we define as the maximum number of gates connected to either the inputs or outputs,
{\change
providing a trade-off between $\fanin$ and $\fanout$.
For hourglass architectures, it is typical to increase the number of feature maps in the encoding part; here, $\fanmax$ adapts to the increasing number of feature maps via $\fanout$ initialization.
The decoder is effectively initialized by $\fanin$ to accommodate the decreasing number of feature maps.
}

\Cref{tab:dbn-ablation}\red{(a -- f)} evaluates these initializations in conjunction with $\linear$ and $\sigmoid$ outputs layers.
Due to the attenuated gradient, all three $\sigmoid$ variants are worse than any $\linear$ output layer.
On the other hand, $\linear$ in conjunction with $\fanmax$ initialization works much better than the traditional $\fanin$ and $\fanout$ initializations, yielding a $\sim$0.7dB benefit.
The visual results in \cref{fig:ablation-outputinit} also reveal that the $\linear$ output contains fewer visual artifacts.

\verdict
For a color prediction task such as deblurring, $\sigmoids$ should be replaced by $\linear$ outputs.
{\change
We recommend considering a $\fanmax$ initialization as an alternative to $\fanin$ and $\fanout$.
}

\begin{figure}[t]
\centering
\subfigure[GT Y]{\includegraphics[width=0.48\linewidth]{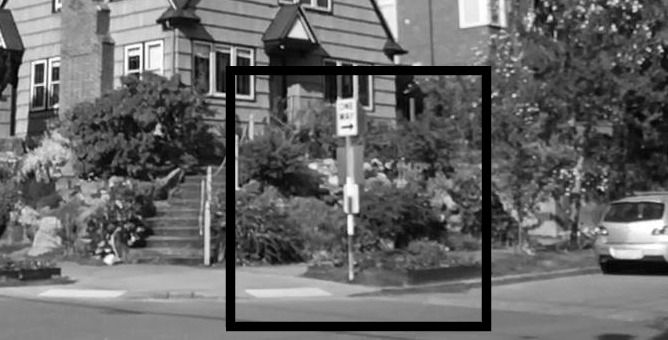}%
\label{fig:ycbcr-oracle-sharp-y}}
\subfigure[Blurry CbCr]{\includegraphics[width=0.48\linewidth]{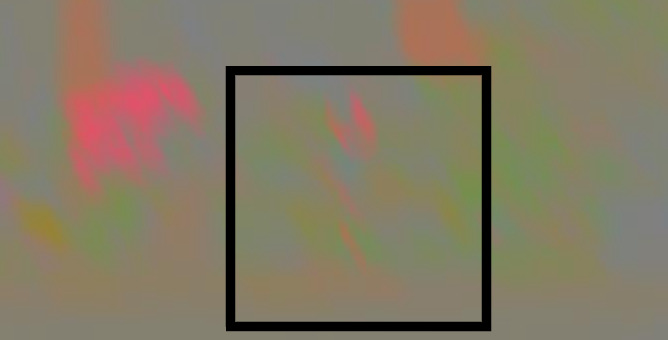}%
\label{fig:ycbcr-oracle-blurry-cbcr}} \\
\subfigure[GT RGB]{\includegraphics[width=0.48\linewidth]{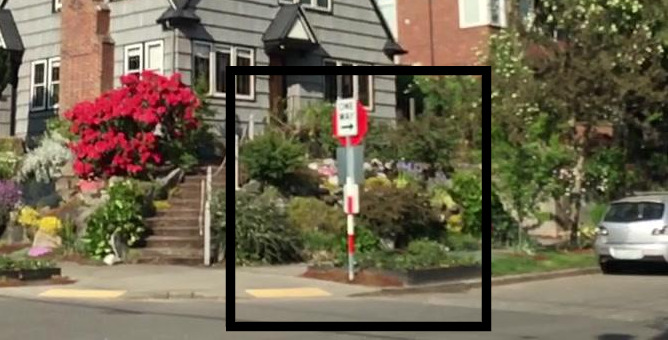}%
\label{fig:ycbcr-oracle-sharp-rgb}}
\subfigure[Reconstructed RGB]{\includegraphics[width=0.48\linewidth]{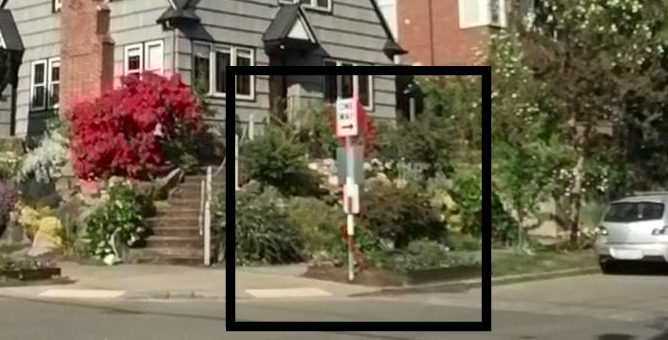}%
\label{fig:ycbcr-oracle-recon}}
\caption{%
{\bf Oracle experiment in YCbCr color space.}
Deblurring in YCbCr color space combines \subref{fig:ycbcr-oracle-sharp-y} the sharp Y channel (here, ground truth) with \subref{fig:ycbcr-oracle-blurry-cbcr} the blurry CbCr channel.
The reconstruction \subref{fig:ycbcr-oracle-recon} is quantitatively close to the RGB ground truth \subref{fig:ycbcr-oracle-sharp-rgb}, yet suffers from halo artifacts for very blurry regions, as highlighted.}
\vspace{-0.5em}
\label{fig:ycbycr-oracle}
\end{figure}

\myparagraph{Color space.}
{\change
In classic deblurring color channels are typically deblurred separately.}
While this is clearly not necessary in deep neural architectures -- we can just output three color channels simultaneously -- the question remains whether the RGB color space is appropriate.
Zhang~\etal \cite{Zhang:2019:ASL} propose to convert the blurry input images to YCbCr space, where Y corresponds to grayscale intensities and CbCr denotes the color components, \cf \cref{fig:ycbycr-oracle}.
The sharp image is subsequently reconstructed from the deblurred Y channels and the blurry input CbCr channels.
{\change
This effectively enforces a natural upper bound on the problem, \ie computing the average PSNR value of the test dataset yields}%
\begin{subequations}
\begin{align}
    \text{PSNR}(\text{RGB}_{\text{input}}, \text{RGB}_{\text{gt}}) &= 27.23\text{dB} \\
    \text{PSNR}(\text{cat}(\text{Y}_\text{gt}, \text{CbCr}_\text{input}), \text{RGB}_{\text{gt}}) &= 56.26\text{dB}.
\end{align}
\end{subequations}
That is, an oracle with access to the ground truth $Y$ channel can achieve at most $56.26$dB PSNR.
Hence, the natural upper bound does not pose a real quantitative limitation, since $56.26$dB is much better than any current method can achieve.
In practice, however, we found that the benefit of solving the problem in YCbCr space is not significant.
\Cref{tab:dbn-ablation}\red{(f, g)} show a minimal $\sim$0.01dB benefit of using YCbCr over RGB.
YCbCr can still be useful as it allows for models with a smaller computational footprint, since fewer weights are required in the first and last layer.
Here, we want to raise another problem of YCbCr deblurring:
For very blurry regions, the reconstruction even from the ground truth Y channel may contain halo artifacts as depicted in \cref{fig:ycbcr-oracle-recon}.

\begin{figure}[t]
\centering
\subfigure[Original statistics]{\includegraphics[width=0.48\linewidth]{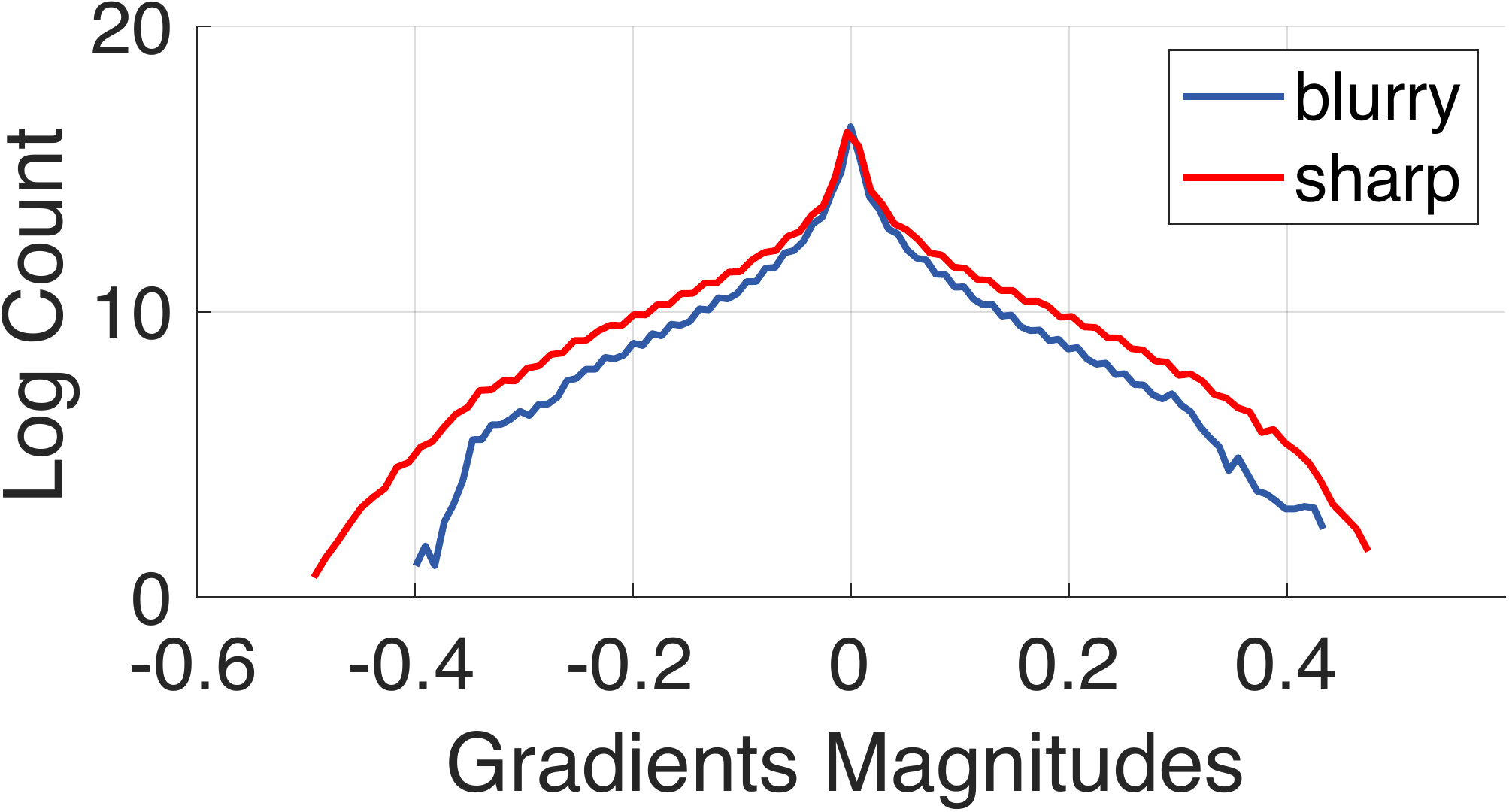}%
\label{fig:scaled-gradient-stats-unscaled}}
\subfigure[Statistics under rescaling]{\includegraphics[width=0.48\linewidth]{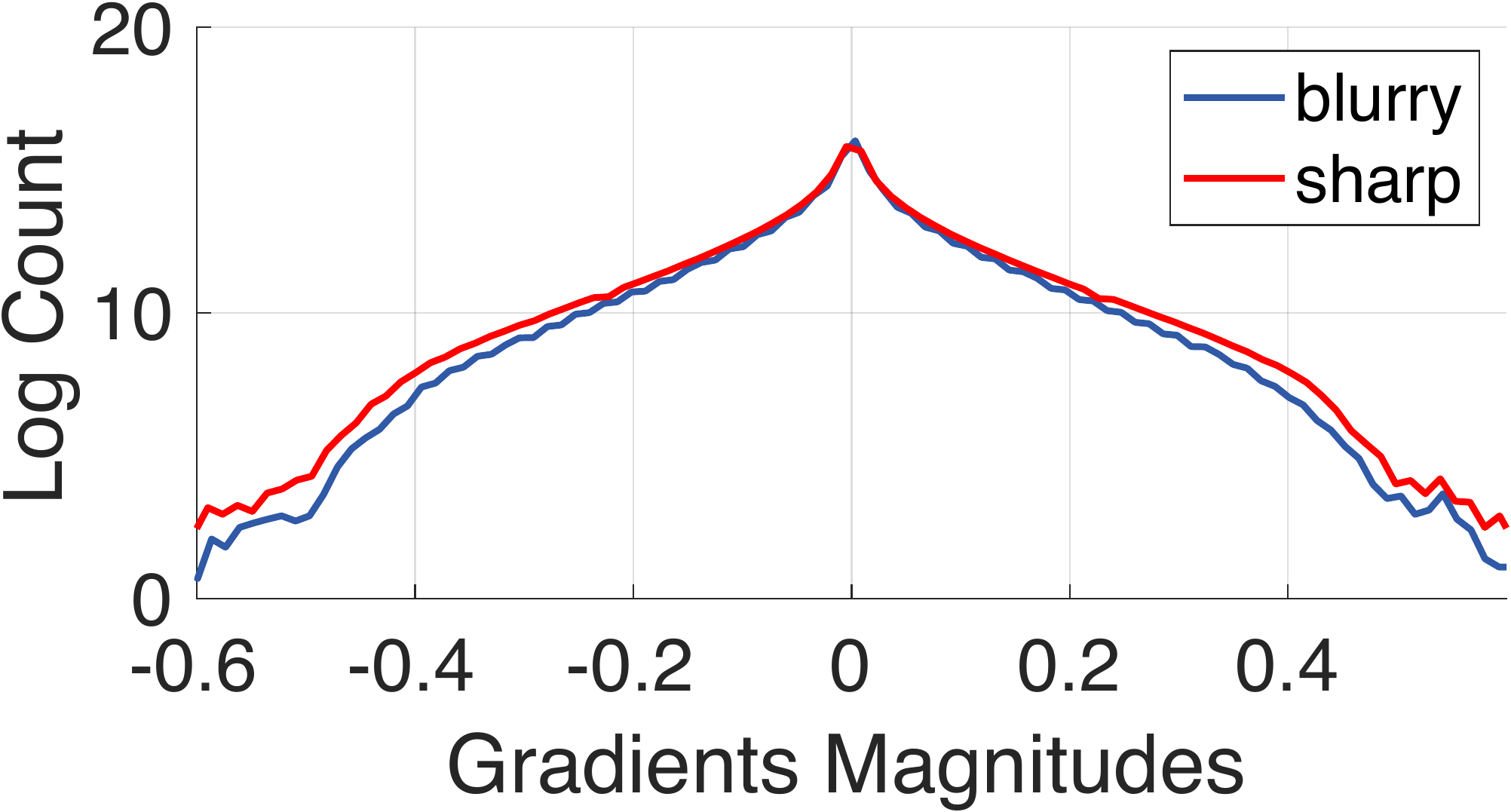}%
\label{fig:scaled-gradient-stats-scaled}}
\caption{%
{\bf Gradient statistics under rescaling.}
Rescaling the images as part of the augmentation is problematic due to the changed degradation statistics (blue -- blurry image statistics, red -- sharp image statistics).
The difference between the plots in unscaled \subref{fig:scaled-gradient-stats-unscaled} \vs rescaled images \subref{fig:scaled-gradient-stats-scaled} is apparent.
}
\vspace{-0.5em}
\label{fig:scaled-gradient-stats}
\end{figure}

\begin{figure*}[t]
\centering
\subfigure[Input]{%
\label{fig:ablation-colorspaceschedule-input}%
\begin{minipage}{4.0cm}
\centering
\includegraphics[height=2.2cm]{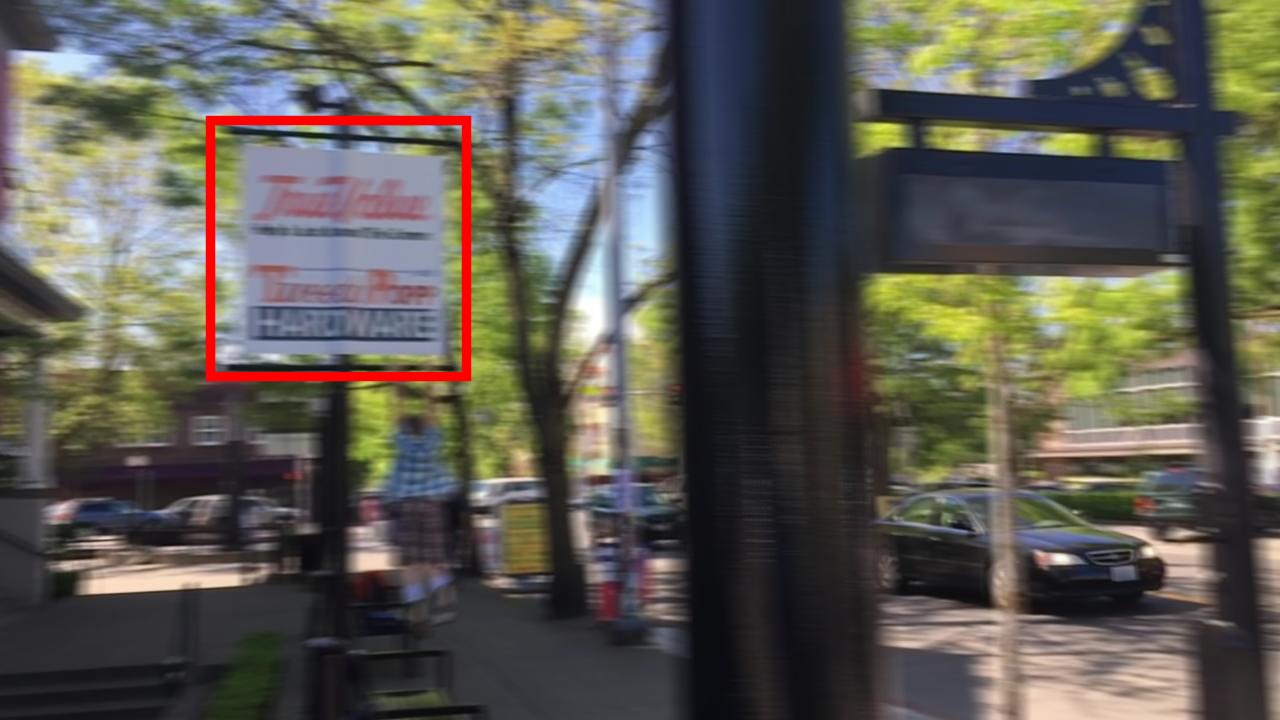}\\
\vspace{0.3em}
\end{minipage}}
%
%
\subfigure[rgb+short]{
\label{fig:ablation-colorspaceschedule-rgbshort}
\begin{minipage}{2.2cm}
\centering
\includegraphics[height=2.2cm]{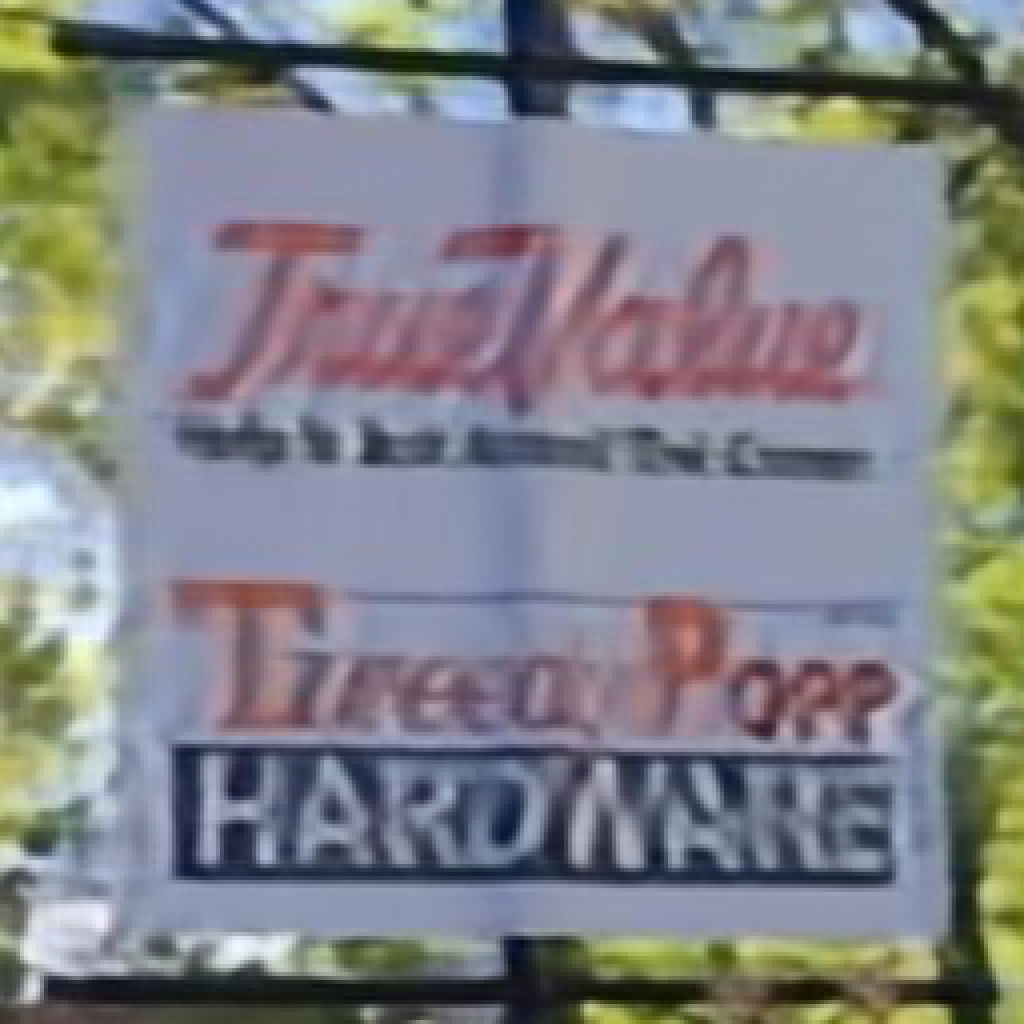}\\
\vspace{0.3em}
\end{minipage}}
%
%
\subfigure[ycbcr+short]{
\label{fig:ablation-colorspaceschedule-ycbcrshort}
\begin{minipage}{2.2cm}
\centering
\includegraphics[height=2.2cm]{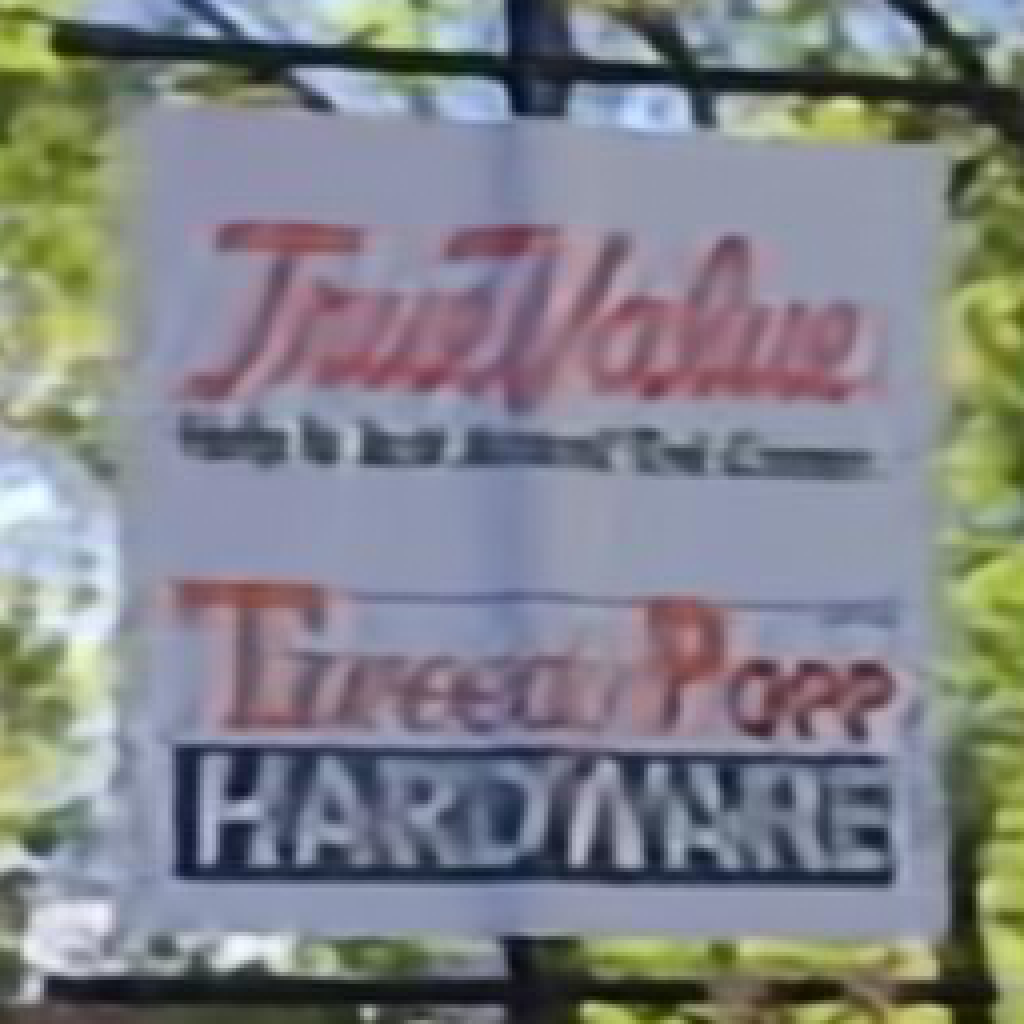}\\
\vspace{0.3em}
\end{minipage}}
%
%
\subfigure[rgb+long]{
\label{fig:ablation-colorspaceschedule-rgblong}
\begin{minipage}{2.2cm}
\centering
\includegraphics[height=2.2cm]{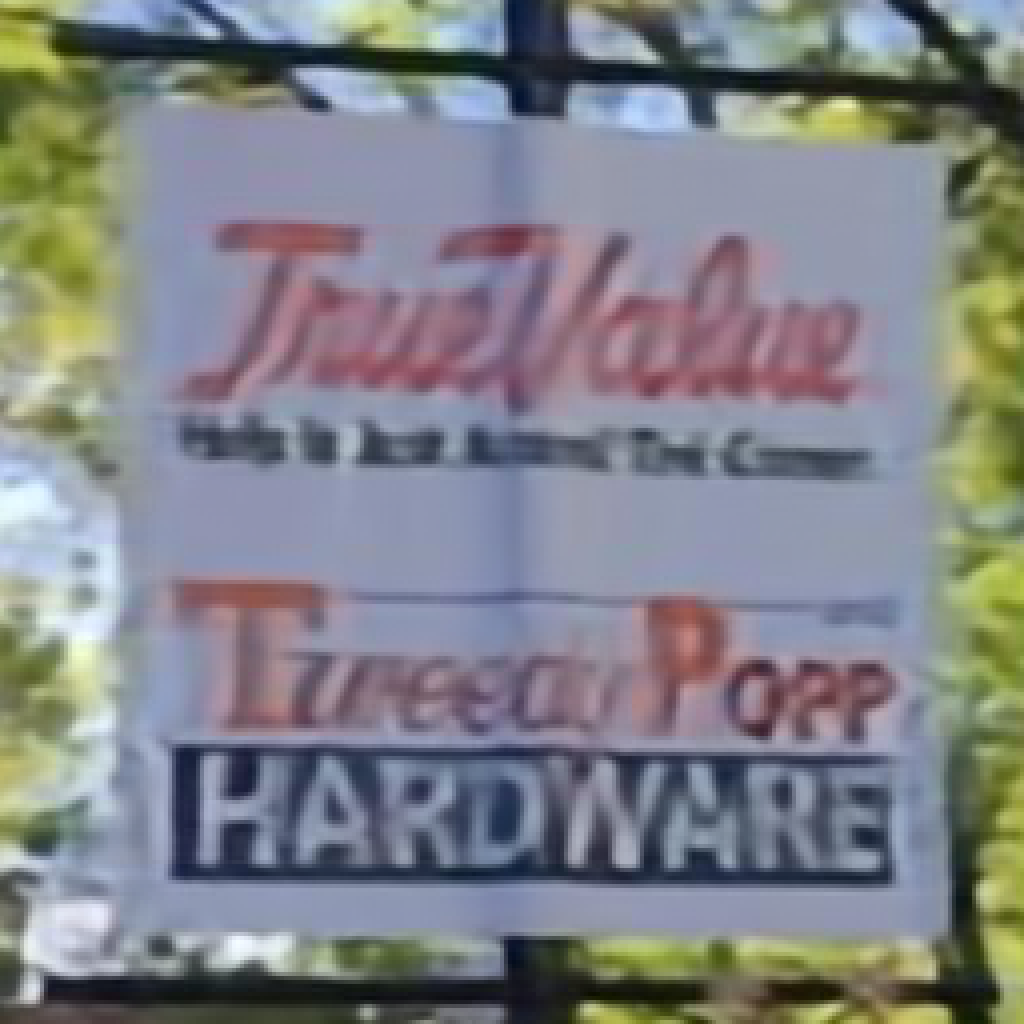} \\
\vspace{0.3em}
\end{minipage}}
%
%
\subfigure[ycbcr+long]{
\label{fig:ablation-colorspaceschedule-ycbcrlong}
\begin{minipage}{2.2cm}
\centering
\includegraphics[height=2.2cm]{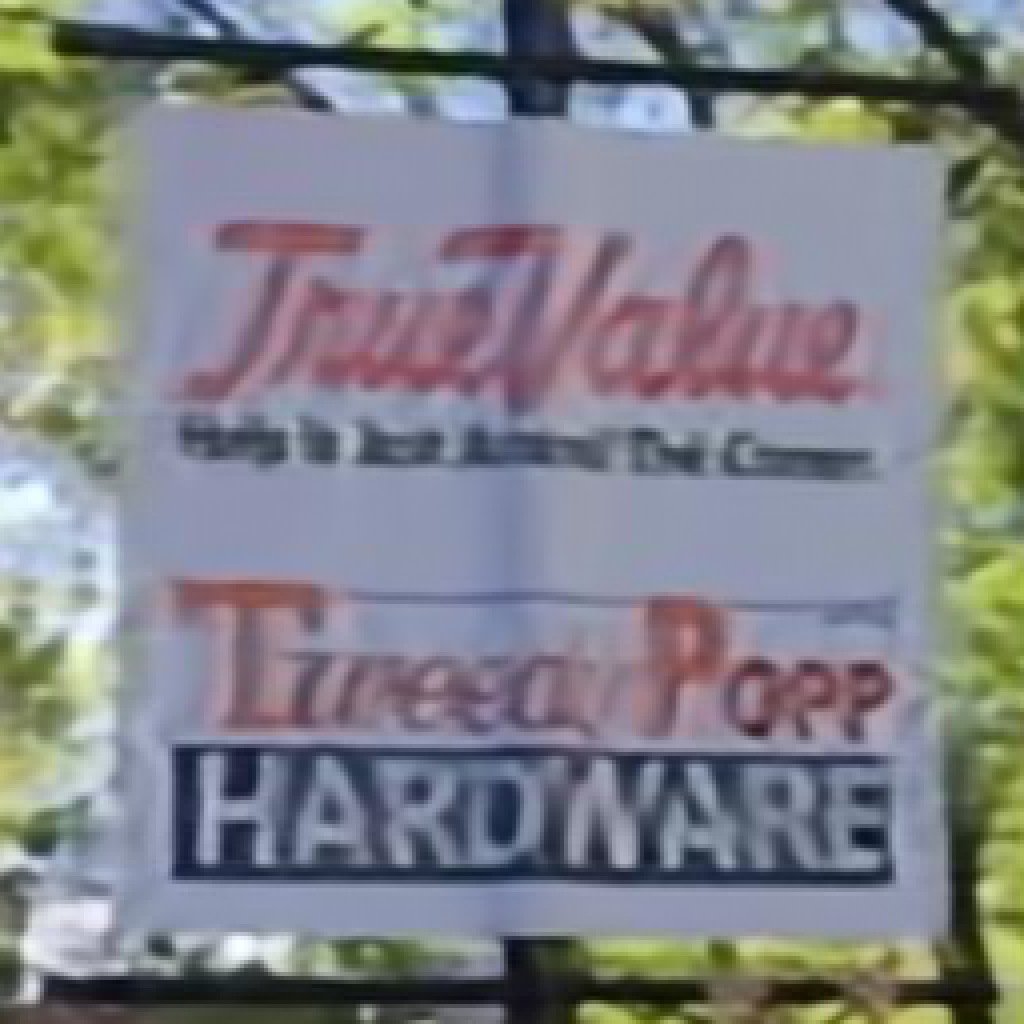}\\
\vspace{0.3em}
\end{minipage}}
%
%
\subfigure[gt]{
\label{fig:ablation-colorspaceschedule-gt}
\begin{minipage}{2.2cm}
\centering
\includegraphics[height=2.2cm]{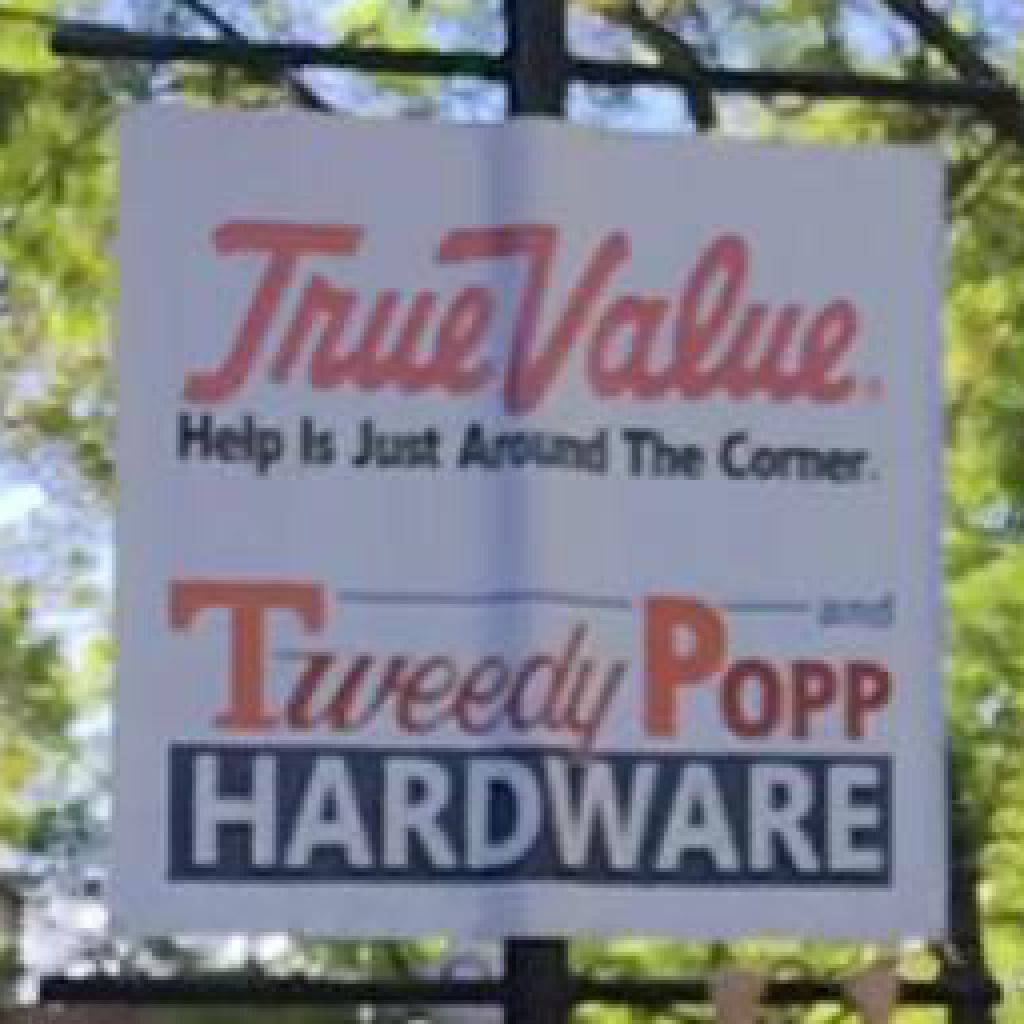}\\
\vspace{0.3em}
\end{minipage}}
\caption{{\bf Color space and training schedule.}
The difference of RGB deblurring \subref{fig:ablation-colorspaceschedule-rgbshort} and YCbCr deblurring \subref{fig:ablation-colorspaceschedule-ycbcrshort} is minimal.
However, using a long training schedule \subref{fig:ablation-colorspaceschedule-rgblong} and
\subref{fig:ablation-colorspaceschedule-ycbcrlong} significantly boosts performance of both.
Note how the last letters of 'HARDWARE' become visibly clearer with the long training schedule.
}
\label{fig:ablation-colorspaceschedule}
\vspace{-0.5em}%
\end{figure*}

\myparagraph{Training schedule.}
As observed in other works, \eg \cite{Ilg:2017:FN2}, longer training schedules can be beneficial for dense prediction tasks.
Here, we apply two different training schedules, a short one with 116 epochs  resembling the original schedule \cite{Su:2017:DVD} by halving the learning rate at epochs $[32, 44, 56, 68, 80, 92, 104]$, as well as a long schedule with 216 epochs, halving the learning rate at epochs $[108, 126, 144, 162, 180, 198]$.
To obtain the long training schedule, we initially inspected the results of running PyTorch's \texttt{ReduceLROnPlateau} scheduler (with patience=10, factor=0.5) for an indefinite time, where we subsequently scheduled the epochs in which learning rates drop in equidistant intervals (here 18).
The longer training schedule improves both the RGB and YCbCr networks roughly by 0.4dB, \cf \cref{tab:dbn-ablation}\red{(f -- i)}.
Since the benefit of YCbCr is rather small for both short and long schedule, we conduct the remaining experiments in RGB space.
\Cref{fig:ablation-colorspaceschedule} shows the visual differences between RGB and YCbCr deblurring.
While the perceptual differences between RGB and YCbCr are not significant, the long schedules improve the readability of the letters over the short ones.

\begin{figure*}[t]
\centering
\subfigure[Input]{%
\label{fig:ablation-augmentations-input}%
\begin{minipage}{4.0cm}
\centering
\includegraphics[height=2.2cm]{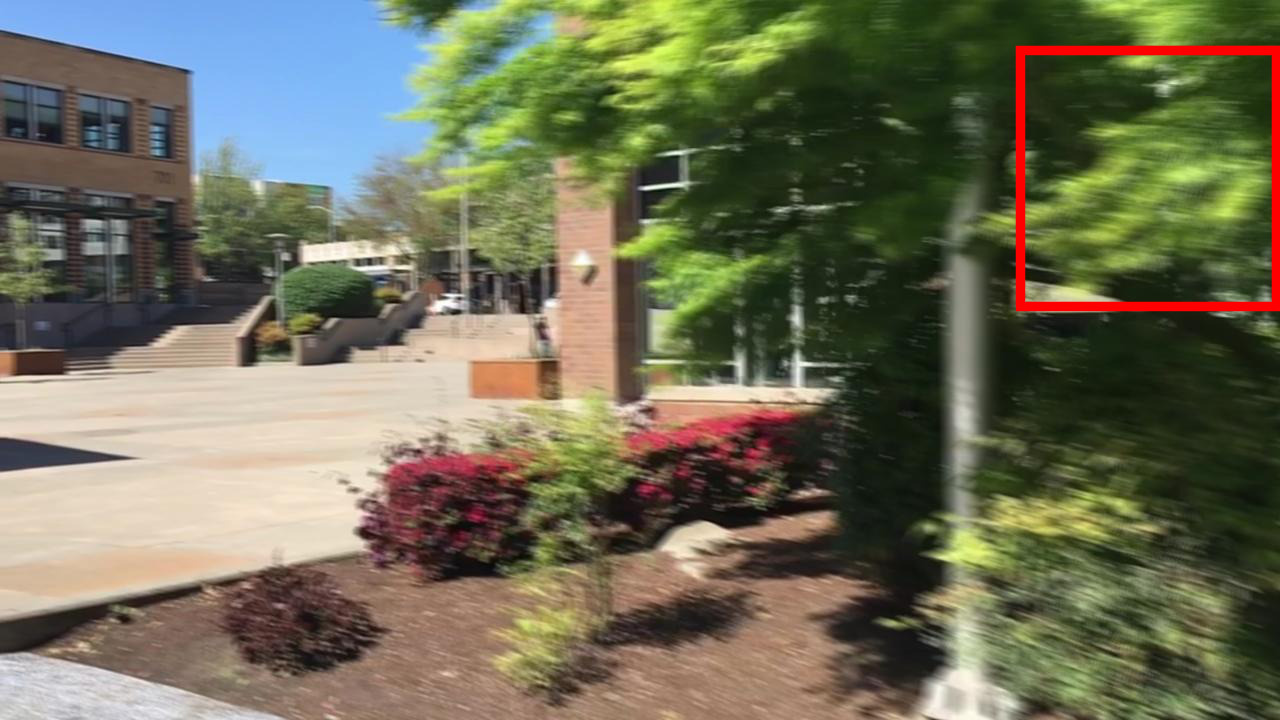}\\
\vspace{0.3em}
\end{minipage}}
%
%
\subfigure[photom.]{
\label{fig:ablation-augmentations-photometrics}
\begin{minipage}{2.2cm}
\centering
\includegraphics[height=2.2cm]{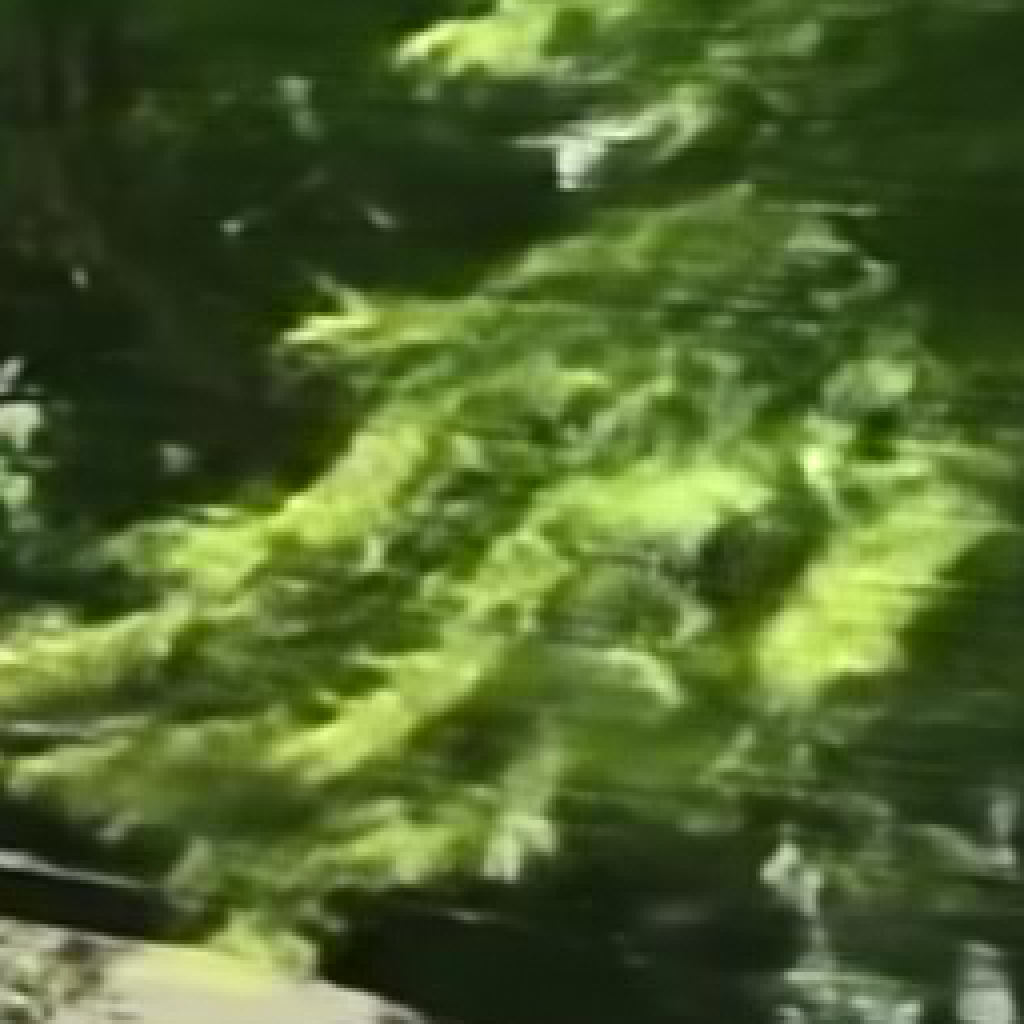}\\
\vspace{0.3em}
\end{minipage}}
%
%
\subfigure[scales]{
\label{fig:ablation-augmentations-scales}
\begin{minipage}{2.2cm}
\centering
\includegraphics[height=2.2cm]{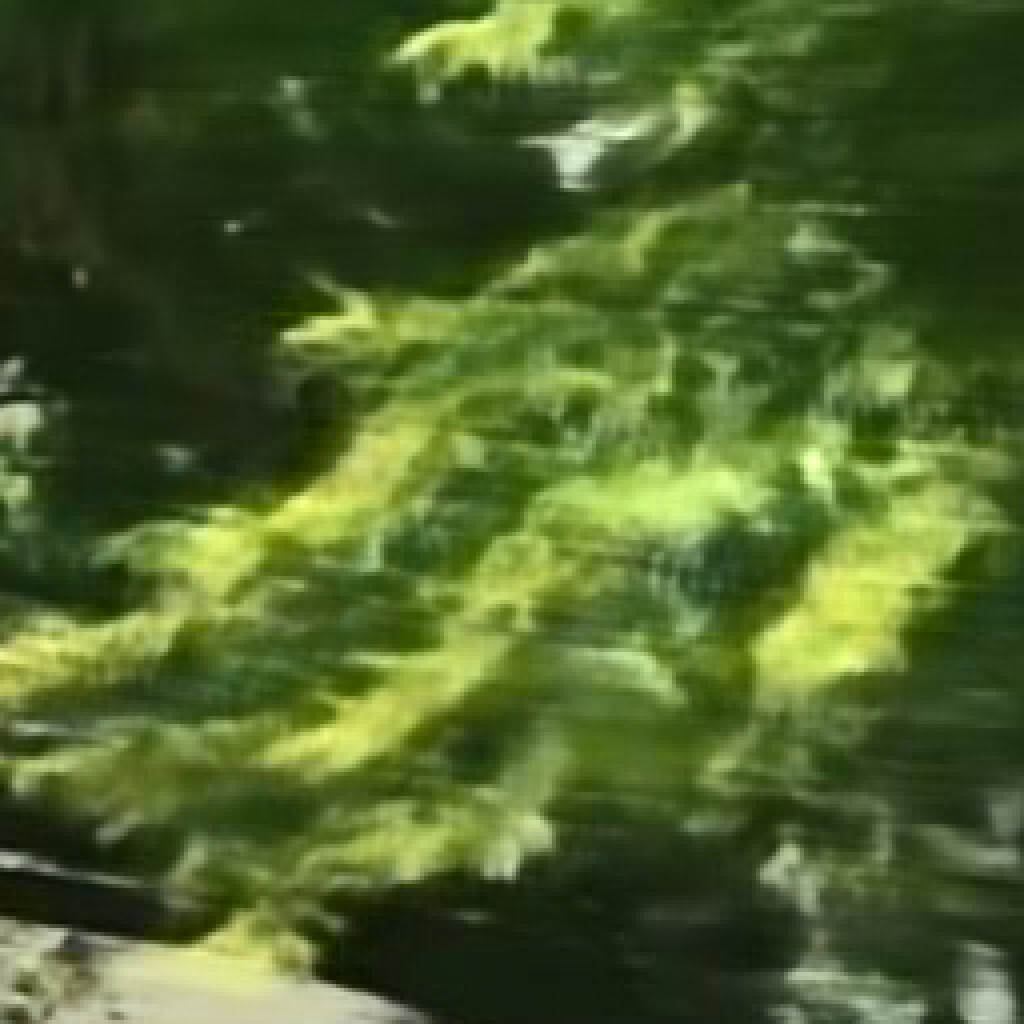} \\
\vspace{0.3em}
\end{minipage}}
%
%
\subfigure[photom.+scales]{
\label{fig:ablation-augmentations-photometrics-scales}
\begin{minipage}{2.2cm}
\centering
\includegraphics[height=2.2cm]{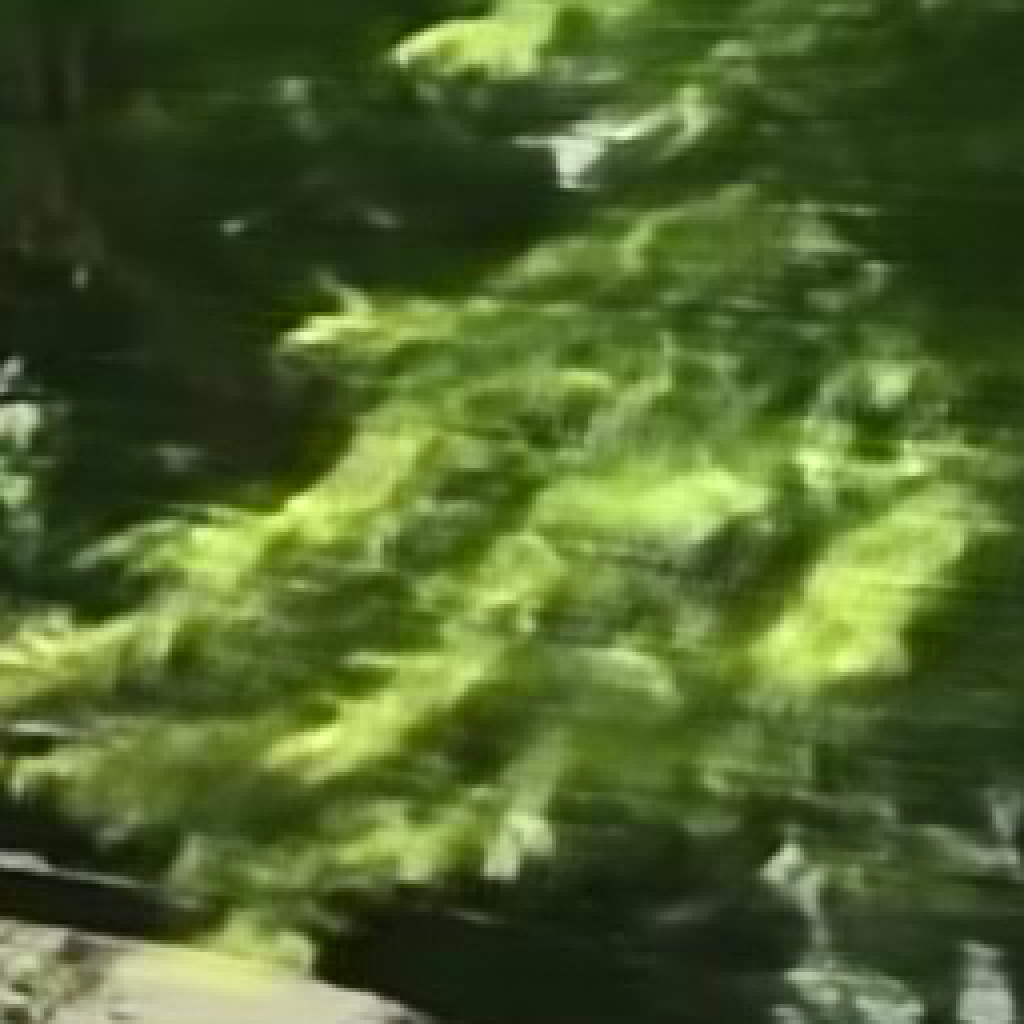}\\
\vspace{0.3em}
\end{minipage}}
%
%
\subfigure[no augm.]{
\label{fig:ablation-augmentations-none}
\begin{minipage}{2.2cm}
\centering
\includegraphics[height=2.2cm]{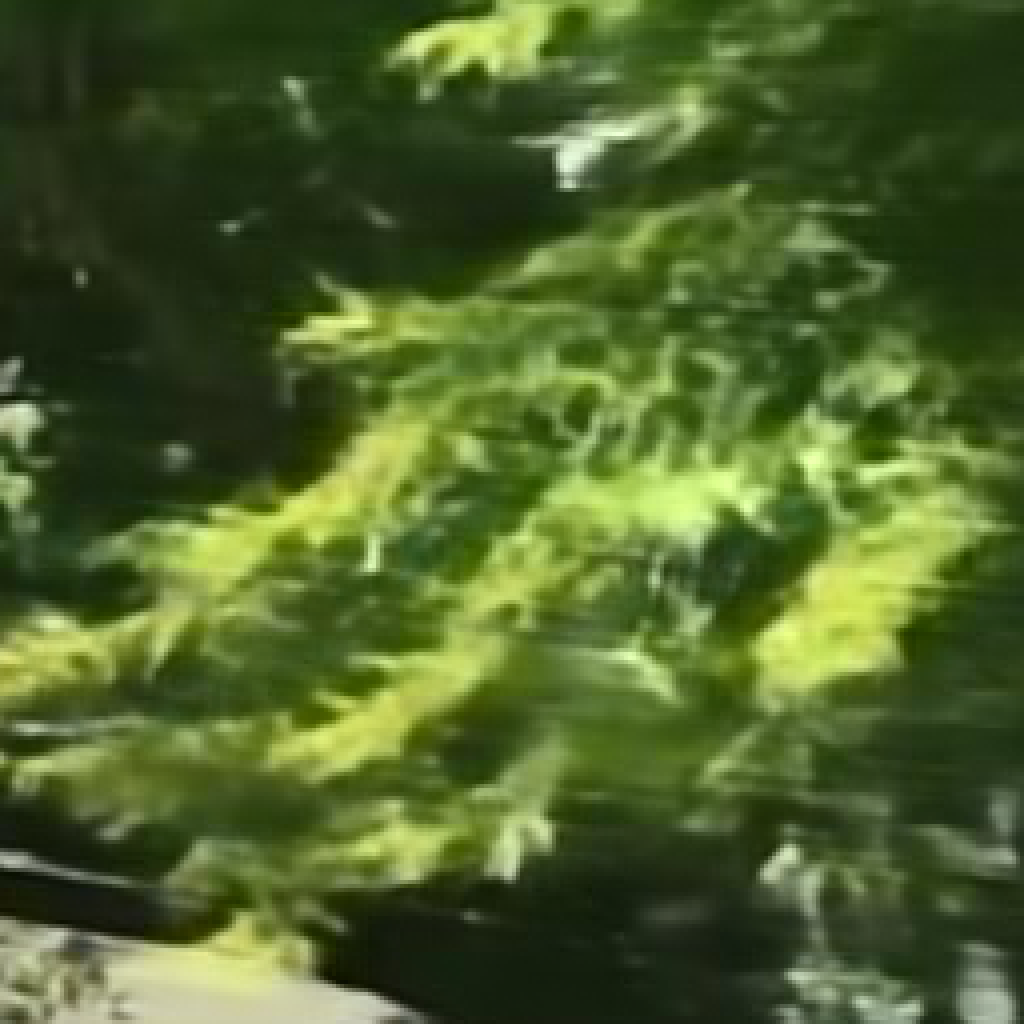}\\
\vspace{0.3em}
\end{minipage}}
%
%
%
\subfigure[gt]{
\label{fig:ablation-augmentations-gt}
\begin{minipage}{2.2cm}
\centering
\includegraphics[height=2.2cm]{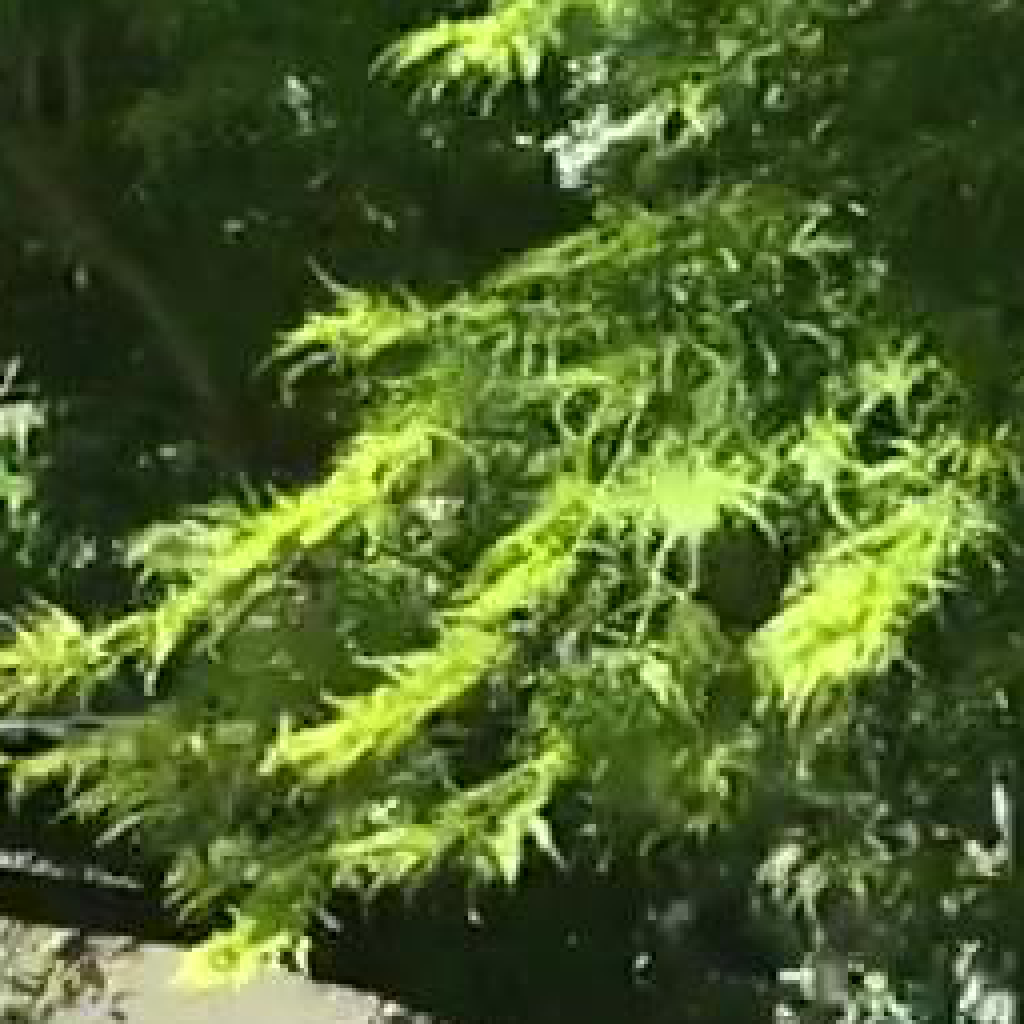}\\
\vspace{0.3em}
\end{minipage}}
\caption{{\bf Varying photometric augmentations and scales.}
Both photometric augmentations and random scales \subref{fig:ablation-augmentations-photometrics}, \subref{fig:ablation-augmentations-scales} have a negative impact on image quality.
The differences are subtle but visually apparent in blobs; compare, \eg, the central part of \subref{fig:ablation-augmentations-photometrics-scales} with \subref{fig:ablation-augmentations-none}.
}
\label{fig:ablation-augmentations}
\vspace{-0.5em}
\end{figure*}

\verdict
YCbCr does not present a significant benefit over RGB; it is, however, viable for very large models, if model size is an issue.
Very blurry training examples may be suboptimal, since even the oracle $Y$ channel yields halo artifacts.
Similar to other dense prediction tasks, long training schedules yield significant benefits.

\myparagraph{Photometric augmentation and random scales.}
Data augmentation plays a crucial role in many dense prediction tasks such as optical flow \cite{Dosovitskiy:2015:FN}.
However, it is often disregarded from the analysis of deblurring methods.
More precisely, while our baseline \cite{Su:2017:DVD} and recent work \cite{Kim:2018:STT, Zhang:2019:ASL} all train under random rotations ($0^\circ, 90^\circ, 180^\circ, 270^\circ$), random horizontal and vertical flips, and random crops (usually of size $128^2$), other types of augmentations such as photometric transformations and random scaling are not agreed upon.
Su \etal~\cite{Su:2017:DVD} train their model under random image scales of $[\nicefrac{1}{4}, \nicefrac{1}{3}, \nicefrac{1}{2}]$, yet Zhang~\etal \cite{Zhang:2019:ASL} do not rescale the training images.
Here, we explore the influence of both random photometric transformations and random scales.

\begin{figure*}
\centering
\subfigure[Input]{%
\label{fig:ablation-flow-input}%
\begin{minipage}{4.0cm}
\centering
\includegraphics[height=2.2cm]{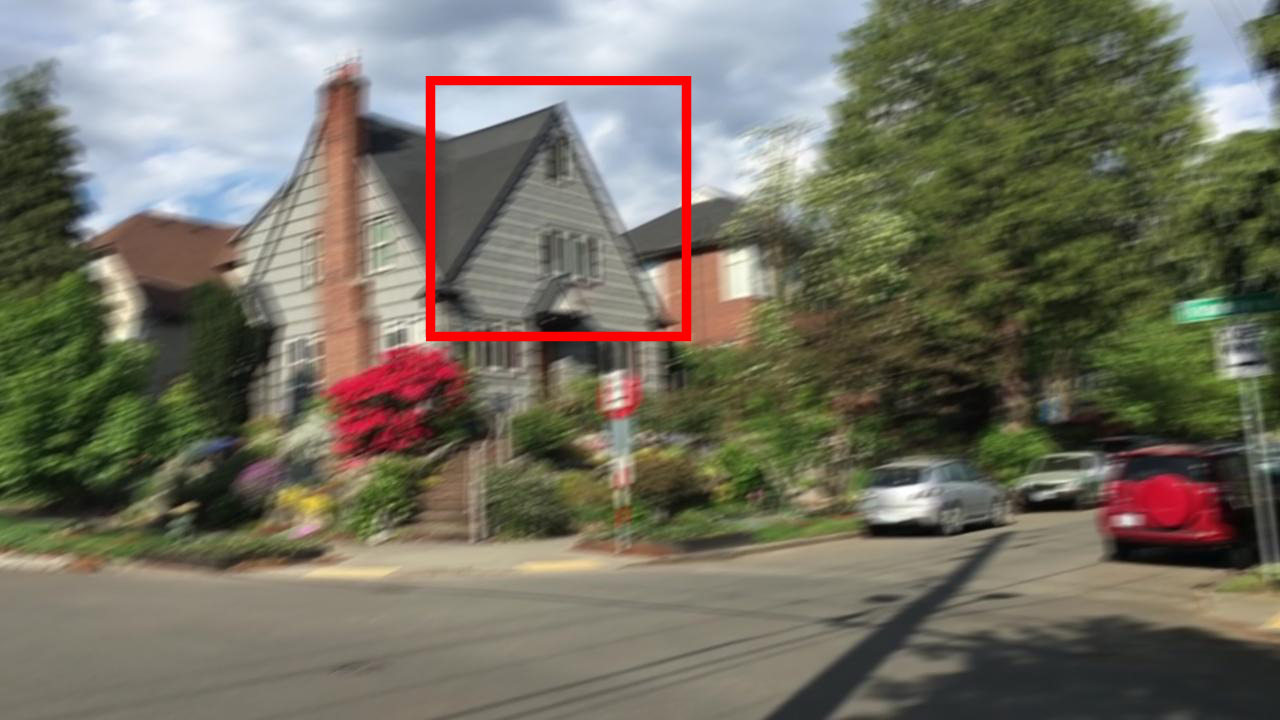}\\
\vspace{0.3em}
\end{minipage}}
%
%
\subfigure[no flow]{
\label{fig:ablation-flow-no}
\begin{minipage}{2.2cm}
\centering
\includegraphics[height=2.2cm]{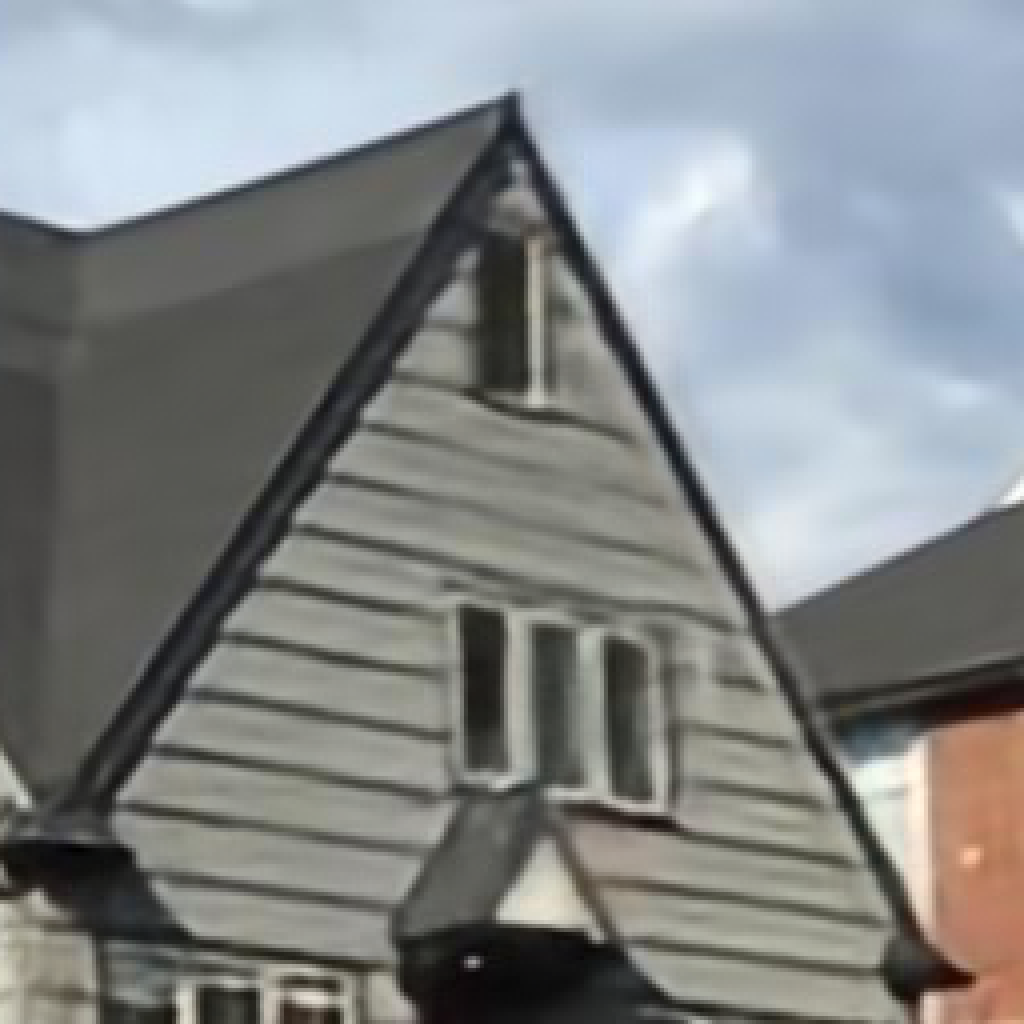}\\
\vspace{0.3em}
\end{minipage}}
%
%
\subfigure[f1s+rep]{
\label{fig:ablation-flow-f1s-rep}
\begin{minipage}{2.2cm}
\centering
\includegraphics[height=2.2cm]{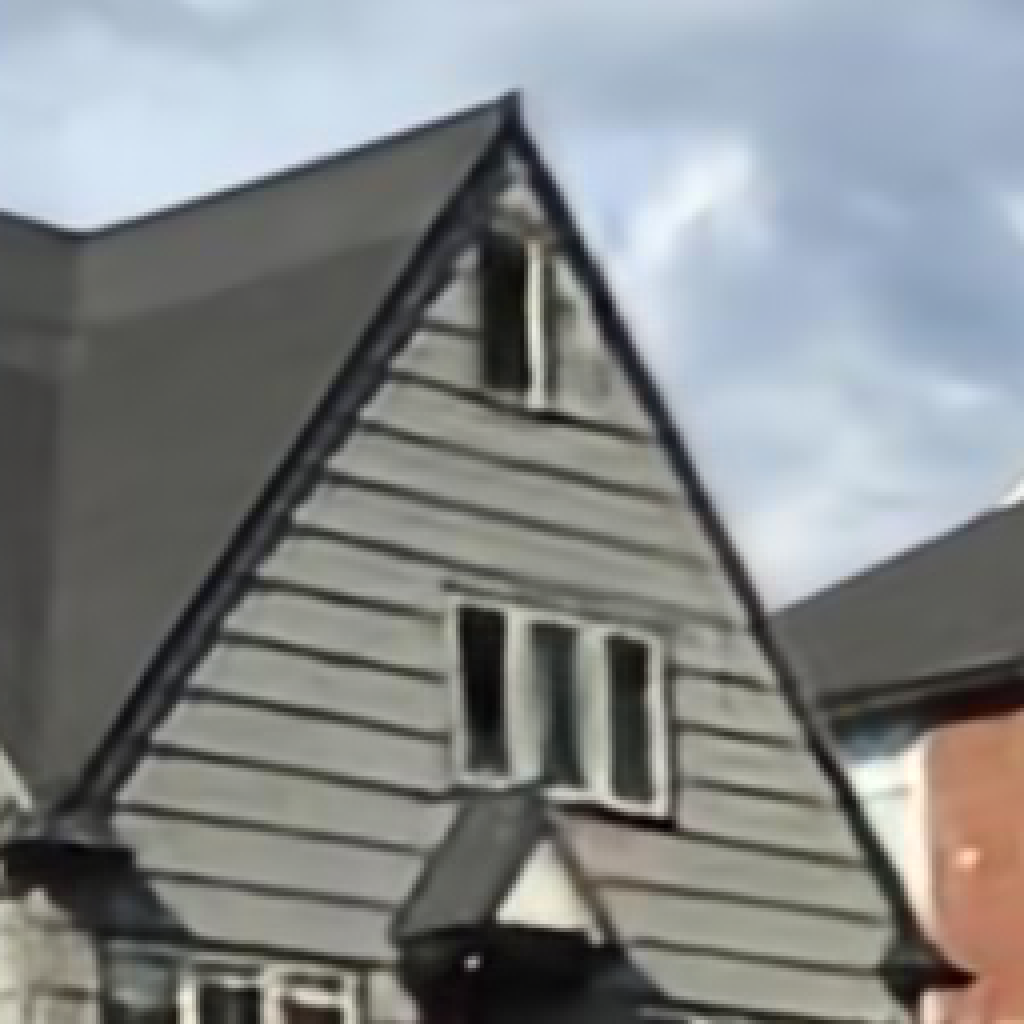}\\
\vspace{0.3em}
\end{minipage}}
%
%
\subfigure[pwc+rep]{
\label{fig:ablation-flow-pwc-rep}
\begin{minipage}{2.2cm}
\centering
\includegraphics[height=2.2cm]{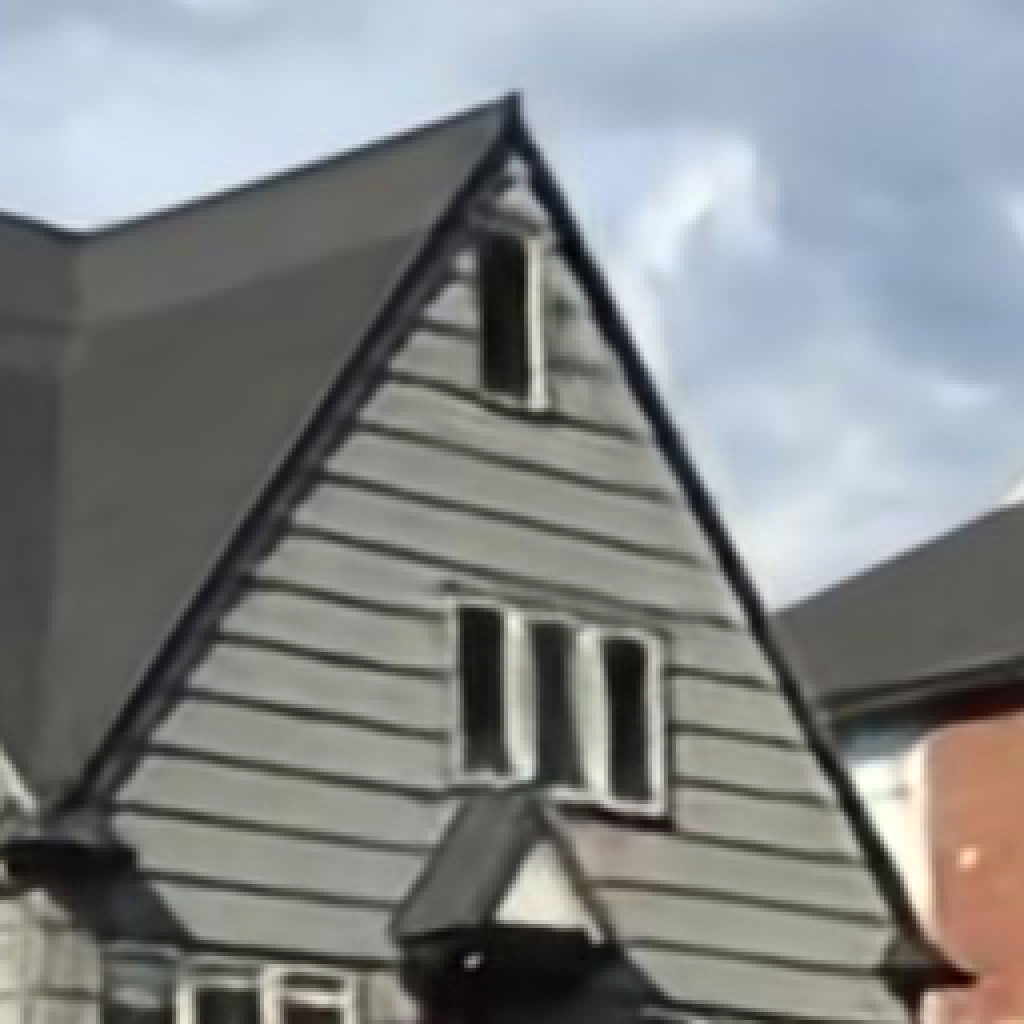} \\
\vspace{0.3em}
\end{minipage}}
%
%
\subfigure[pwc+cat]{
\label{fig:ablation-flow-pwc-cat}
\begin{minipage}{2.2cm}
\centering
\includegraphics[height=2.2cm]{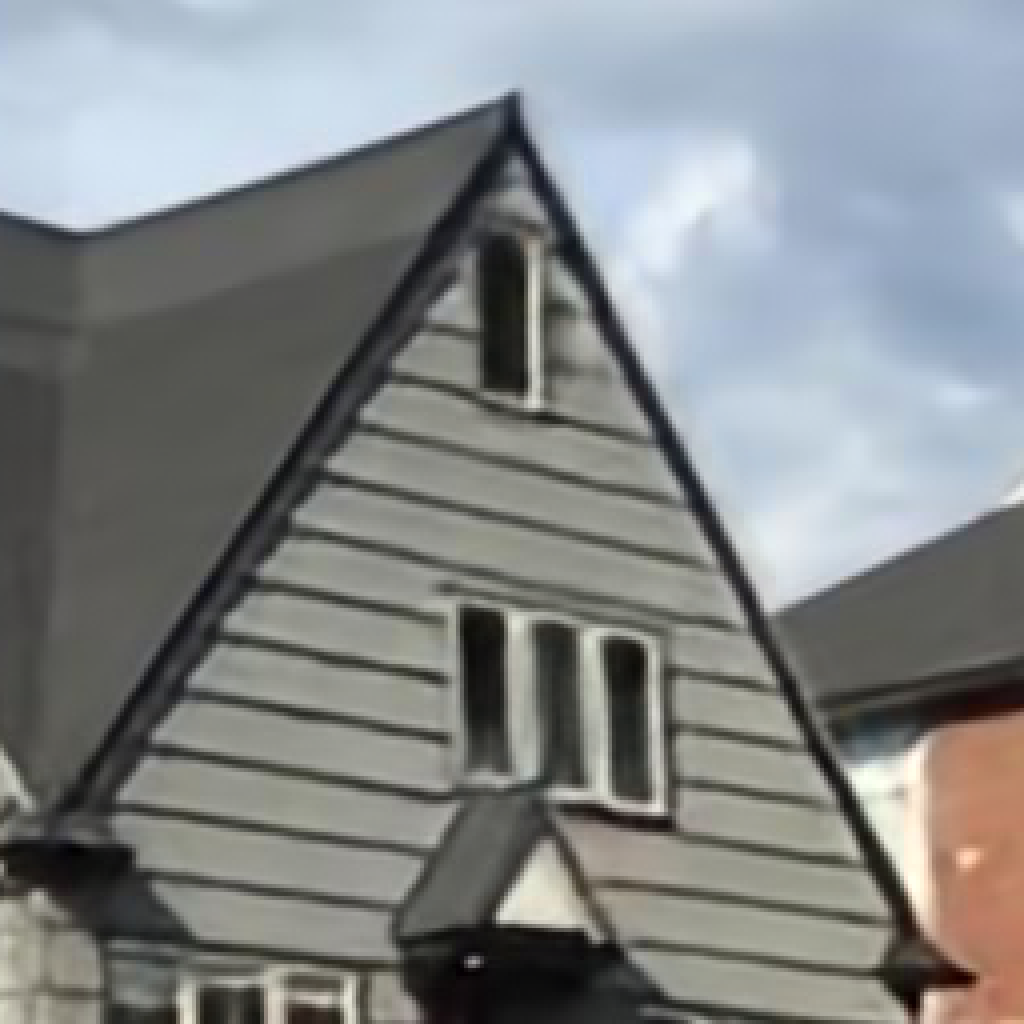}\\
\vspace{0.3em}
\end{minipage}}
%
%
\subfigure[gt]{
\label{fig:ablation-flow-gt}
\begin{minipage}{2.2cm}
\centering
\includegraphics[height=2.2cm]{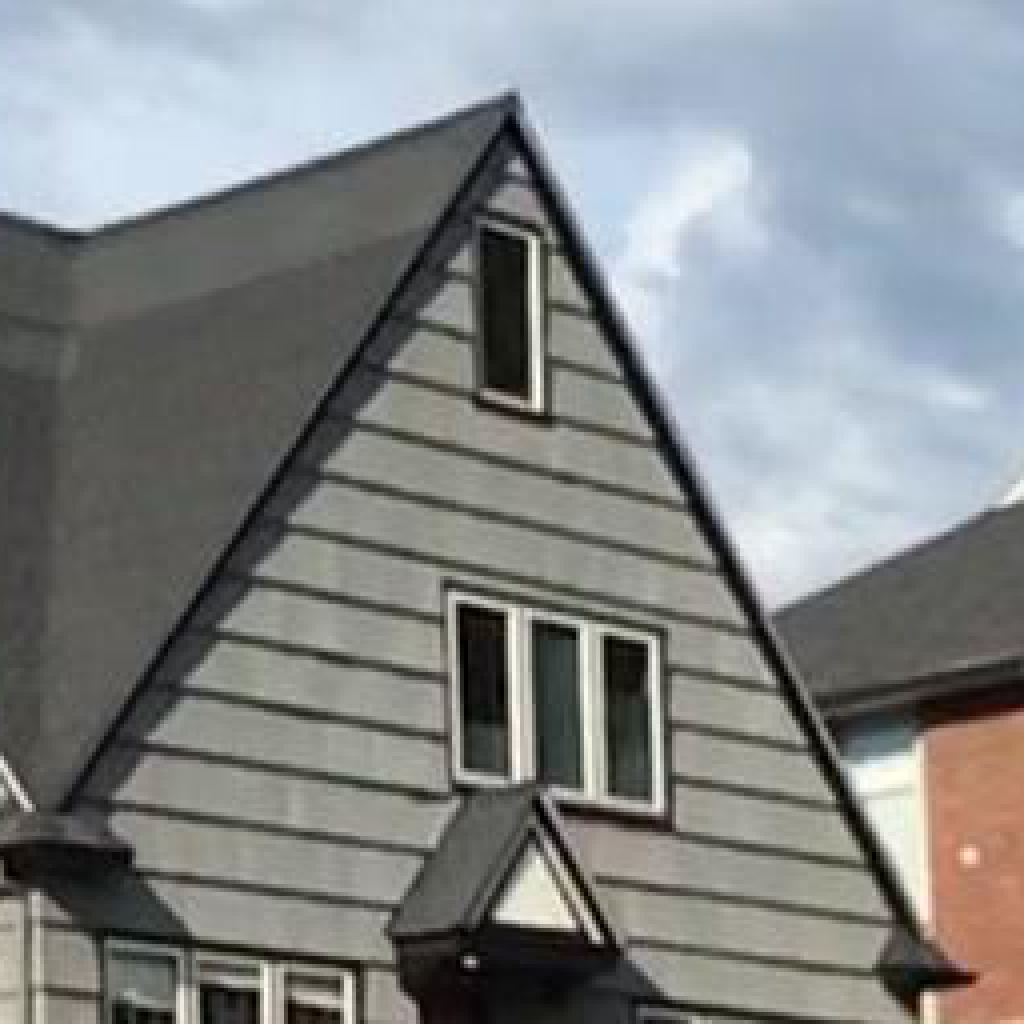}\\
\vspace{0.3em}
\end{minipage}}
\caption{{\bf Optical flow prewarping.}
Prewarping with optical flow positively influences image quality.
We experiment with FlowNet1S \subref{fig:ablation-flow-f1s-rep}, and PWC-Net \subref{fig:ablation-flow-pwc-rep}, \subref{fig:ablation-flow-pwc-cat}.
Concatenating warped images with the inputs \subref{fig:ablation-flow-pwc-cat} produces fewer visual artifacts than just replacing the temporal neighbors \subref{fig:ablation-flow-pwc-rep}.
Here all flow variants reconstruct the horizontal structures much better than the baseline without pre-warping \subref{fig:ablation-flow-no}.
}
\label{fig:ablation-flow}
\vspace{-0.5em}
\end{figure*}

\begin{figure*}
\centering
\subfigure[Input]{%
\label{fig:ablation-crops-input}%
\begin{minipage}{4.0cm}
\centering
\includegraphics[height=2.2cm]{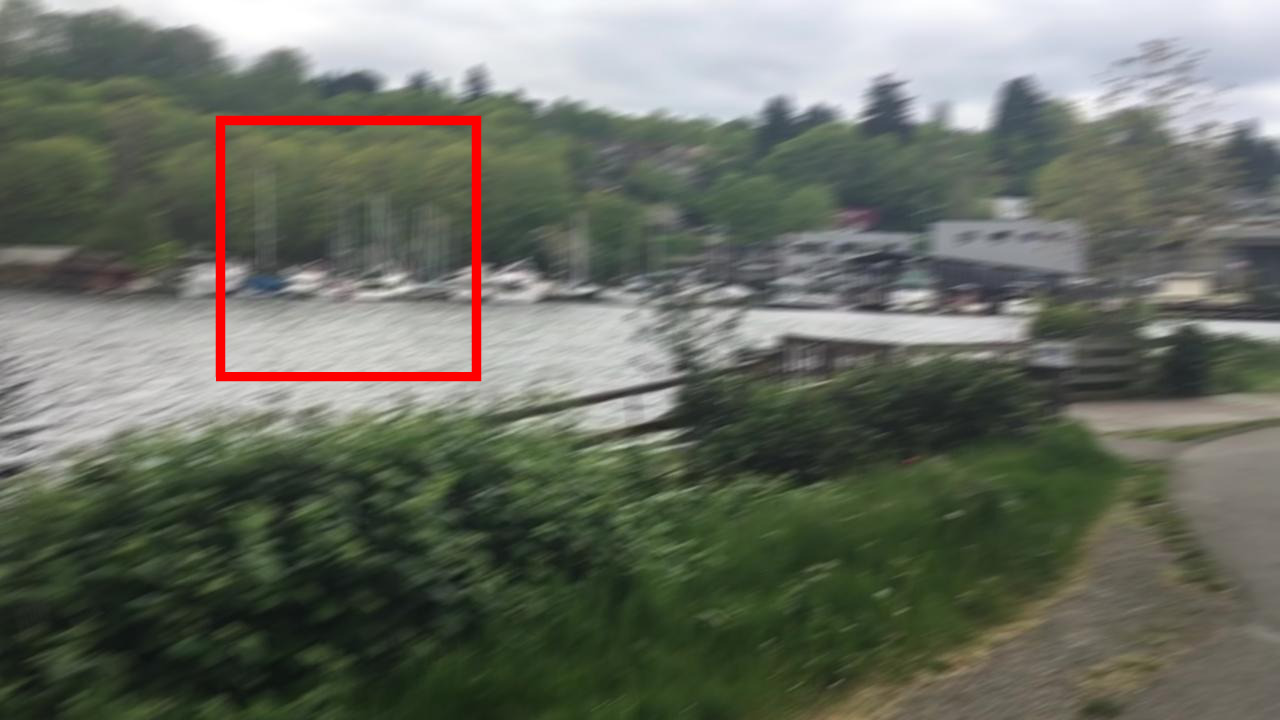}\\
\vspace{0.3em}
\end{minipage}}
%
%
\subfigure[64x64]{
\label{fig:ablation-crops-64}
\begin{minipage}{2.2cm}
\centering
\includegraphics[height=2.2cm]{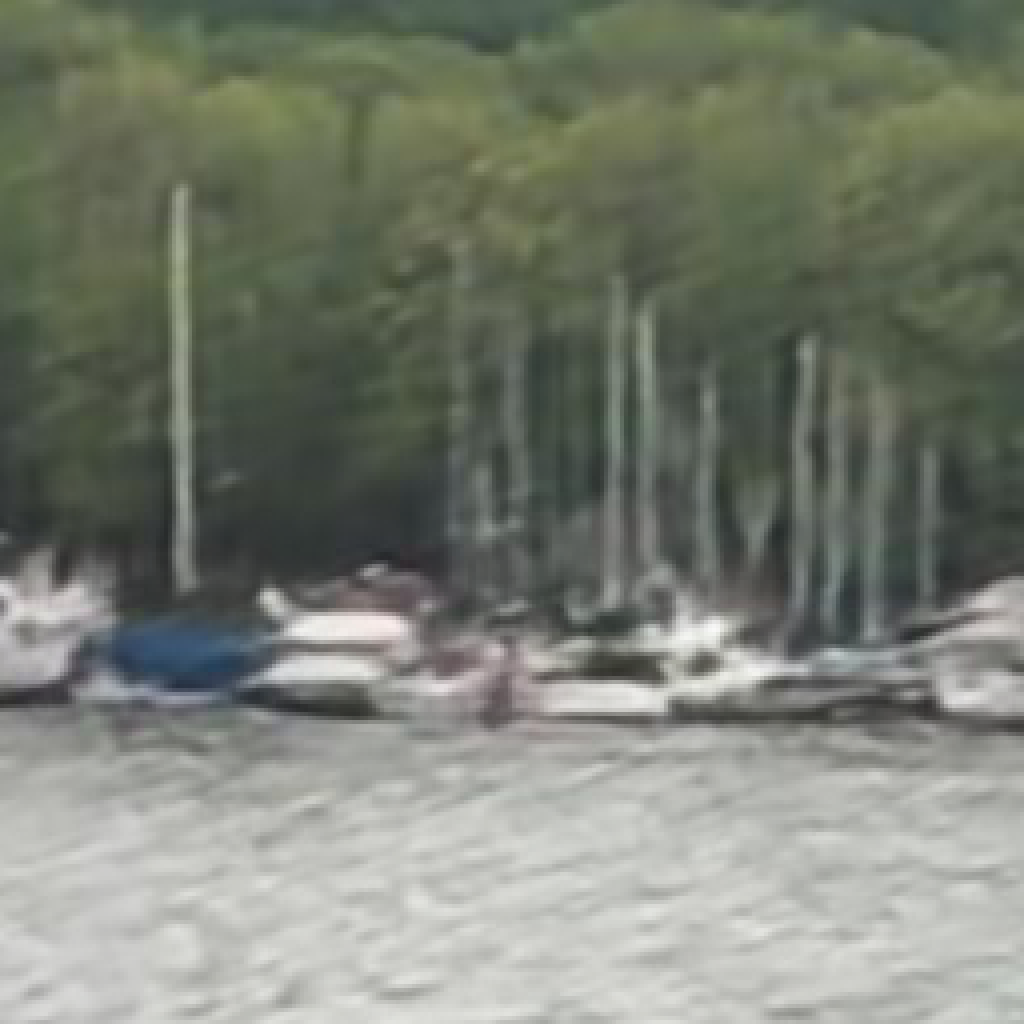}\\
\vspace{0.3em}
\end{minipage}}
%
%
\subfigure[128x128]{
\label{fig:ablation-crops-128}
\begin{minipage}{2.2cm}
\centering
\includegraphics[height=2.2cm]{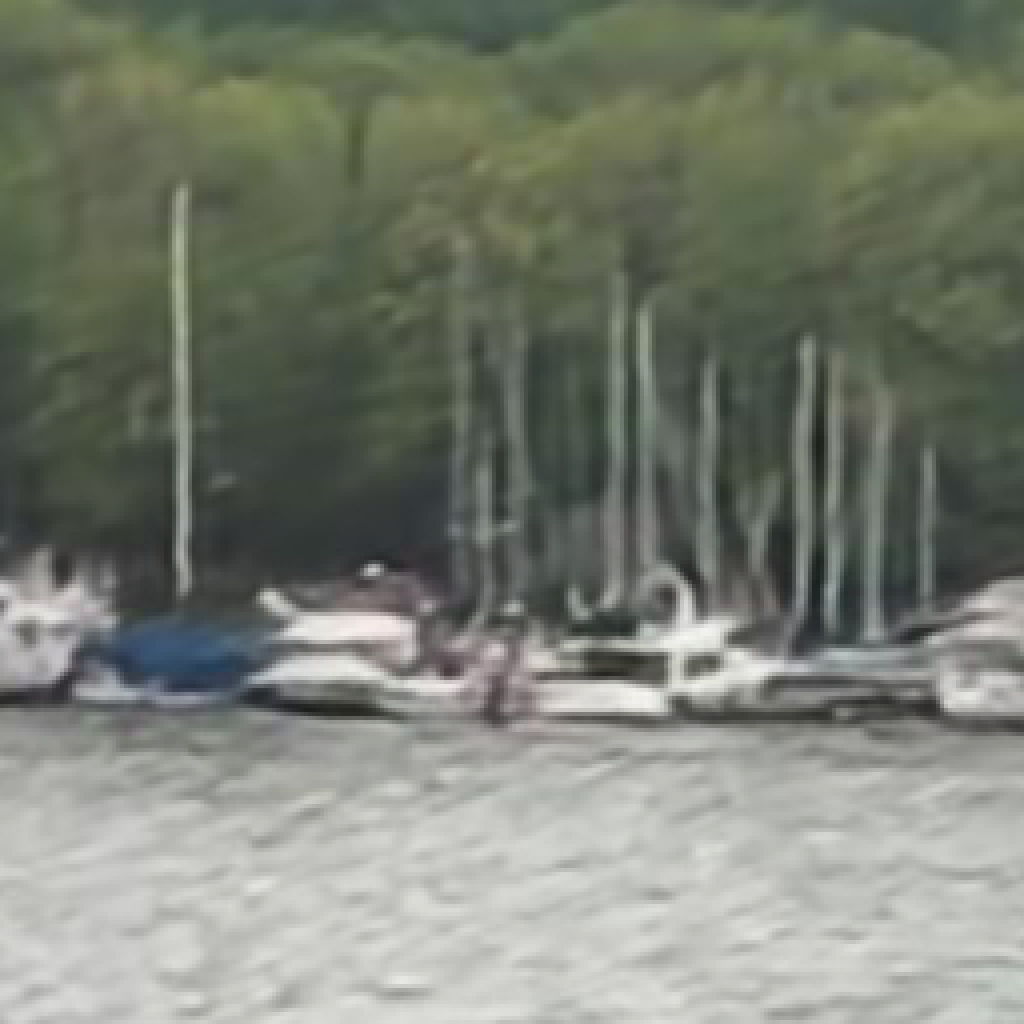} \\
\vspace{0.3em}
\end{minipage}}
%
%
\subfigure[160x160]{
\label{fig:ablation-crops-160}
\begin{minipage}{2.2cm}
\centering
\includegraphics[height=2.2cm]{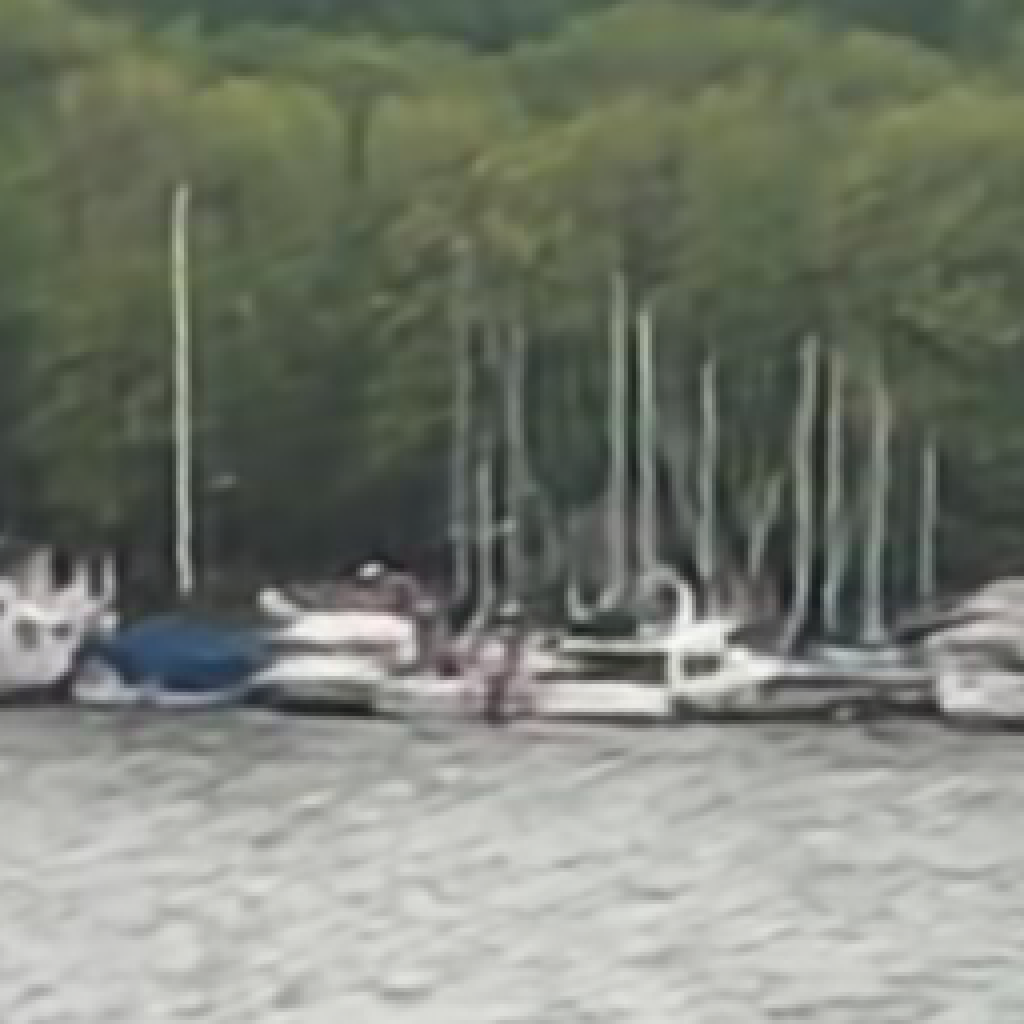}\\
\vspace{0.3em}
\end{minipage}}
%
%
\subfigure[192x192]{
\label{fig:ablation-crops-192}
\begin{minipage}{2.2cm}
\centering
\includegraphics[height=2.2cm]{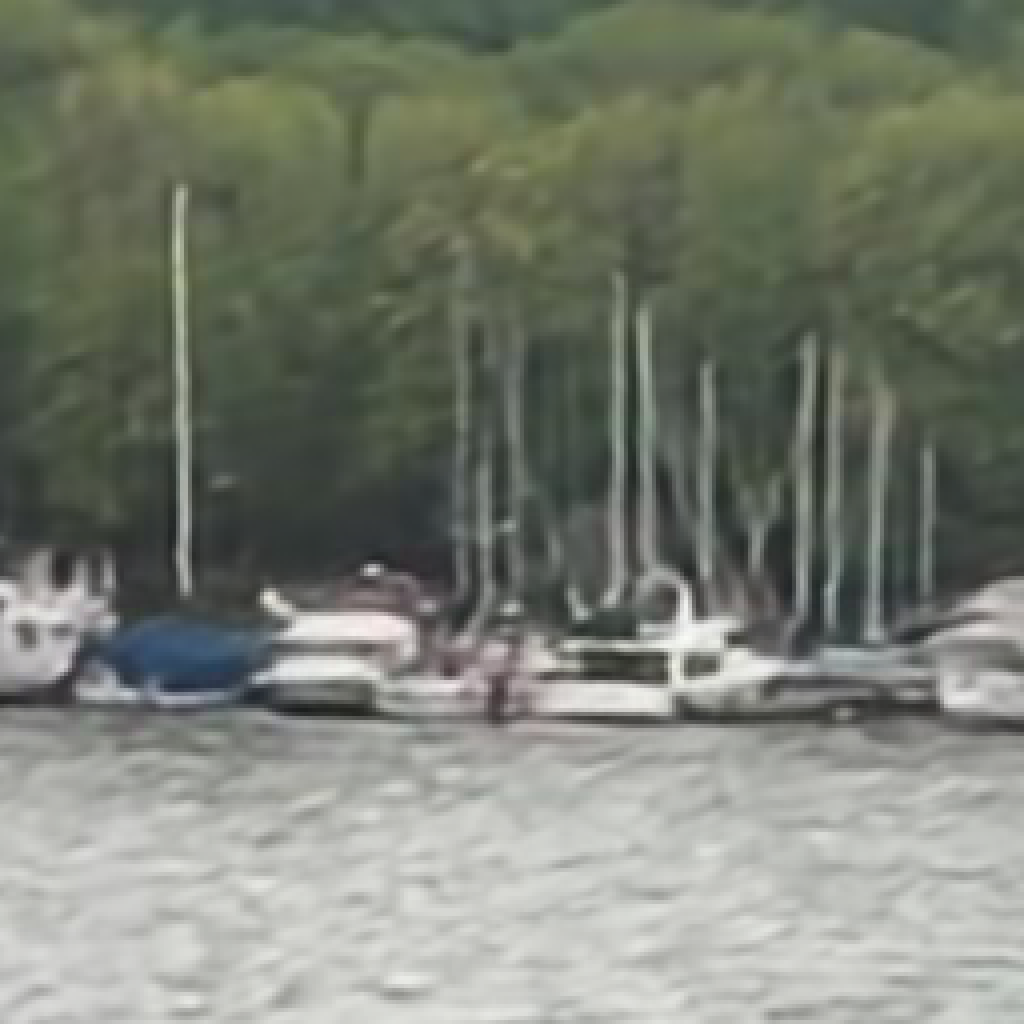}\\
\vspace{0.3em}
\end{minipage}}
%
%
%
\subfigure[gt]{
\label{fig:ablation-crops-gt}
\begin{minipage}{2.2cm}
\centering
\includegraphics[height=2.2cm]{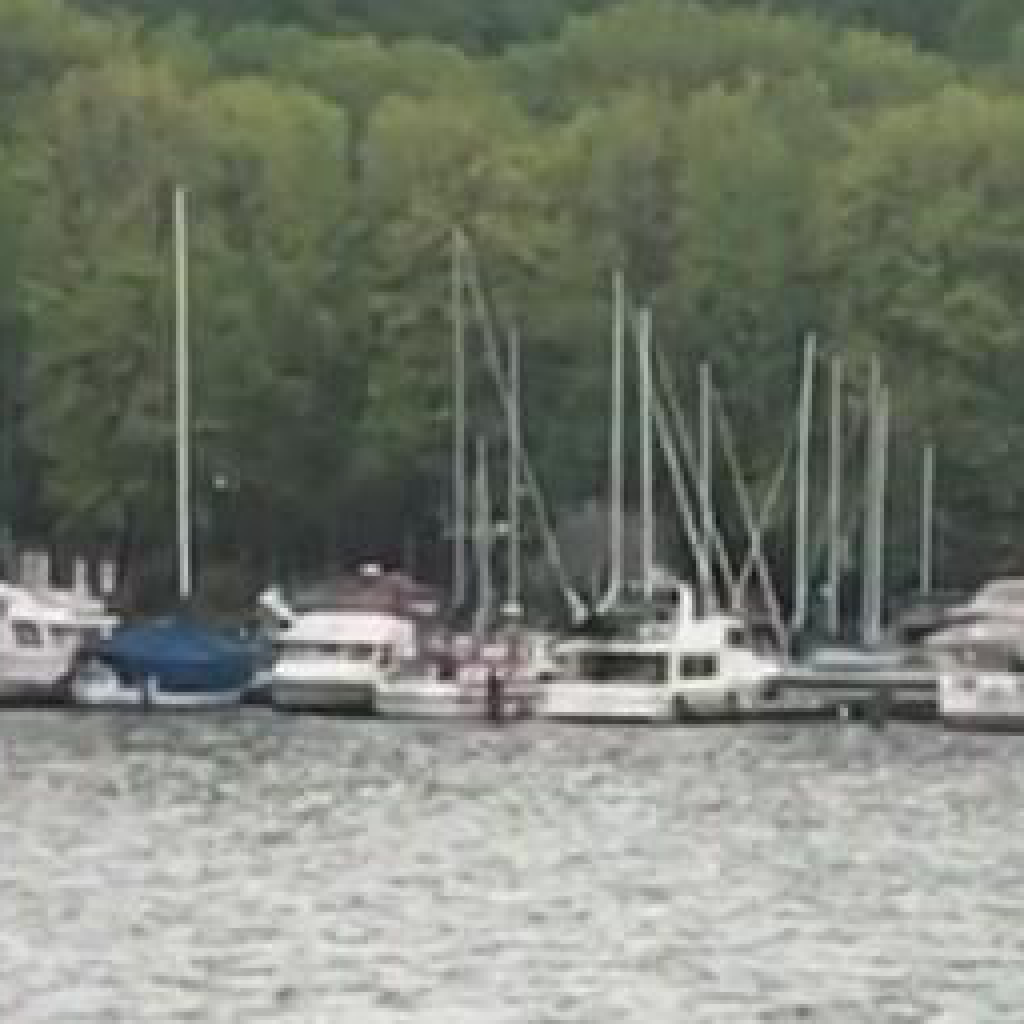}\\
\vspace{0.3em}
\end{minipage}}
\caption{{\bf Varying training crop size.}
Increasing the size of training patches is a simple, yet effective method to increase image quality.
Here we experiment with square patches of size $64\times64$ \subref{fig:ablation-crops-64} -- $192\times192$ \subref{fig:ablation-crops-192}.
The visual gain is biggest for smaller patch sizes.
Note how the left pole becomes sharper with increasing patch size.
}
\label{fig:ablation-crops}
\vspace{-0.5em}
\end{figure*}

We use four settings: No augmentations (other than random orientations and crops, \cref{tab:dbn-ablation}\red{(h)}), random photometric transformations (using PyTorch's popular random color jitter on hue, contrast, and saturation with p=0.5, \cref{tab:dbn-ablation}\red{(j)}), random scales (with a random scale factor in $[0.25, 1.0]$, \cref{tab:dbn-ablation}\red{(k)}), and with both augmentations (\cref{tab:dbn-ablation}\red{(l)}).
We find that these augmentations significantly hurt image quality; the quantitative difference between no and both augmentations (\cref{tab:dbn-ablation}\red{(h \vs l)}) amounts to a surprising 0.44dB.
Here, the photometric augmentations alone decrease the accuracy by 0.26dB (\cref{tab:dbn-ablation}\red{(h \vs j)}).
While we do not argue that any photometric augmentation will hurt accuracy, our results suggest that the common color jitter is counterproductive in deblurring; we attribute this to the fact that commonly applied photometric co-transforms obfuscate the ground truth signal for general non-uniform blur.
{\change
To illustrate this issue, let $\bm{P}$ be a photometric operator (applied to sharp images), and $\bm{K}$ be a non-uniform blur operator, respectively.
If $\bm{P}$ was linear, we could derive the appropriate photometric operator $\tilde{\bm{P}}$ for blurry images as%
\begin{align}
\tilde{\bm{P}} \bm{K} = \bm{K} \bm{P} \quad \Rightarrow \quad \tilde{\bm{P}} = \bm{K} \bm{P} \bm{K}^{-1}.
\end{align}
As there is no ground truth $\bm{K}$ available for the GOPRO datasets, the correct photometric transformation $\tilde{\bm{P}}$ to be applied to blurry images is not available.
}

{\change
The performance drop induced by random scales roots in a change of relative image statistics between blurry and sharp images.}
To that end, consider the gradient histogram statistics of 300 training image crops shown in \cref{fig:scaled-gradient-stats-unscaled} as well as the statistics for rescaled crops (scale factor 0.25) in \cref{fig:scaled-gradient-stats-scaled}.
The comparison reveals two points:
First, the original statistics are sparser than the rescaled ones.
Second, rescaling renders the gap of statistics between the blurry and sharp gradients less pronounced.
This difference manifests in a quantitative difference of 0.22dB, \cf \cref{tab:dbn-ablation}\red{(h \vs k)}.
Visually, the difference is most apparent for blob-like regions, \cf the leaves of the tree in \cref{fig:ablation-augmentations}.
Not applying any photometric or scale augmentation \subref{fig:ablation-augmentations-none} yields slightly clearer results than either random photometric transformations \subref{fig:ablation-augmentations-photometrics}, random scales \subref{fig:ablation-augmentations-scales}, or both \subref{fig:ablation-augmentations-photometrics-scales}.

\verdict
In contrast to other dense prediction problems, where photometric augmentations and random rescaling in training help to improve generalization, these augmentations can hurt the generalization performance of deblurring models.
One should thus be careful in choosing augmentation methods, as they may obfuscate the data statistics.

\myparagraph{Optical flow warping.}
Su~\etal \cite{Su:2017:DVD} experimented with pre-warping input images based on classic optical flow methods such as \cite{Perez:2013:TVL} to register them to the reference frame.
Surprisingly, they did not observe any empirical benefit, hence abandoned flow warping.
Yet, Chen~\etal \cite{Chen:2018:R2D} use a flow network after the deblurring network to predict an output sequence of sharp images, which is subsequently registered to the reference frame.
This consistency is worked into the loss function, which allows them to improve over the DBN baseline (\cf \cref{tab:gopro-su-short}).
Kim~\etal \cite{Kim:2018:STT} propose to put a spatio-temporal transformer network in front of the DBN baseline to transform 3D inputs (the stack of blurry input images) to the reference frame;
the synthesized images and the reference frame are then fed into the baseline network.
In contrast to \cite{Su:2017:DVD}, they observed the temporal correspondence to improve the deblurring accuracy.

While using a spatio-temporal transformer is elegant, we argue that the underlying correspondence estimation problem is itself very hard and requires a lot of engineering to achieve high accuracy \cite{Sun:2019:MMT}.
Hence, we consider pre-warping with the output from standard optical flow networks.
To avoid any efficiency concerns \cite{Kim:2018:STT}, we rely on pre-trained flow networks, which obviates backpropagating through them.
We experiment with two different backbones that we put in front of our baseline:
FlowNet1S (denoted as f1s) \cite{Dosovitskiy:2015:FN} and PWC-Net (denoted as pwc) \cite{Sun:2018:PWC}.
For both backbones, we warp the neighboring frames to the reference frame, and either input the reference frame along the replaced, warped neighbors (+ rep), or we concatenate the warped neighbors with the original input (+ cat).
{\change
Note that while concatenation allows the network to possibly overcome warping artifacts using the original inputs, this is not possible without the original input.
}
Our experiments in \cref{tab:dbn-ablation}\red{(m -- p)} show that, in contrast to the conclusions in \cite{Su:2017:DVD}, simple flow warping already helps  (0.15dB improvement in \red{(m -- n)} over the no-flow baseline \red{(i)}).
A more substantial benefit of $\sim$0.4dB comes from concatenating the warped images along the original inputs (\cref{tab:dbn-ablation}\red{(o -- p)}).
Perhaps surprisingly, the FlowNet1S backbone performs only slightly worse than PWC-Net.
The visual results in \cref{fig:ablation-flow} reveal that flow-based methods clearly improve upon the no-flow variant, which exhibits artifacts at the horizontal structures of the house.
Also note how the PWC-Net backbone is clearer in deblurring the horizontal structures than the FlowNet1S variant, despite the small quantitative difference.
Visually, pwc+cat further improves over pwc+rep, \eg note the boundaries of the windows.

\begin{table*}[t]
\caption{{\bf Comprehensive ablation study.}}
\label{tab:dbn-ablation}
\vspace{0.25em}
\centering
%
\footnotesize
%
\begin{tabularx}{\textwidth}{@{}lYYYYYYYYYS[table-format=2.2,table-number-alignment=right]@{}}
\toprule
\# & Output activation  & Initialization & Color space & Schedule & Random photom. & Random scales &  Flow &  Random crops & Sequence length & PSNR \\
\midrule
{\red a}  & $\sigmoid$ & $\fanout$    & RGB & short & \xmark   & \xmark  &  -  & $128^2$  & 5 &  29.04 \\
{\red b}  &        & $\fanin$     &     &       &          &         &     &          &   &  29.26 \\
{\red c}  &        & $\fanmax$     &     &       &          &         &     &          &   & \bfseries 30.00 \\

\midrule
{\red d} &$\linear$ & $\fanout$     & RGB  & short & \xmark   & \xmark  &  -  & $128^2$  & 5 & 30.09 \\
{\red e} &          & $\fanin$      &      &       &          &         &     &          &   &  30.31 \\
{\red f} &          & $\fanmax$     &      &       &          &         &     &          &   & \bfseries 31.07 \\
\midrule
{\red g} &$\linear$ & $\fanmax$ & YCbCr & short  & \xmark   & \xmark  &  -  & $128^2$  & 5 &   31.08     \\
{\red h} &          &          & RGB   & long   &          &         &     &          &   &    31.48     \\
{\red i} &         &          & YCbCr & long   &          &         &     &          &   &   \bfseries 31.50 \\
\midrule
{\red j} &$\linear$ & $\fanmax$     & RGB & long  & \cmark   & \xmark  &  -  & $128^2$  & 5 &  31.22 \\
{\red k} &          &              &     &       & \xmark   & \cmark  &     &          &   &  31.26 \\
{\red l} &          &              &     &       & \cmark   & \cmark  &     &          &   &  31.04 \\
\midrule
{\red m} &$\linear$  & $\fanmax$  & RBG & long &  \xmark  &  \xmark  & f1s~+~rep    & $128^2$  & 5 & 31.62  \\
{\red n} &           &           &     &      &          &          &  pwc~+~rep    &          &   & 31.67  \\
{\red o} &           &           &     &      &          &          &  f1s~+~cat    &          &   & 31.89   \\
{\red p} &           &           &     &      &          &          &  pwc~+~cat    &          &   & \bfseries 31.91  \\
\midrule
{\red q}&$\linear$  & $\fanmax$  & RBG & long &  \xmark  & \xmark    &  pwc~+~cat  & $64^2$   & 5 & 31.23 \\
{\red r}&           &           &     &      &          &           &             & $96^2$   &   & 31.71 \\
{\red s}&           &           &     &      &          &           &             & $160^2$  &   & 32.05 \\
{\red t}&           &           &     &      &          &           &             & $192^2$  &   &\bfseries 32.14 \\
\midrule
{\red u} &$\linear$  & $\fanmax$  & RBG & long &  \xmark  & \xmark    &  pwc~+~cat  & $128^2$   & 7  &  31.94  \\
{\red v}&           &           &     &      &          &           &               & $192^2$   &  & \bfseries 32.19  \\
\bottomrule
\end{tabularx}
\vspace{-0.5em}
\end{table*}

\verdict
While previous work proposes a sophisticated treatment of temporal features, we find that pre-trained optical flow networks perform quite well.
Concatenating warped neighbors to the inputs works significantly better than just replacing inputs.
While a good flow network may not quantitatively improve over a simple one, deblurred images may show subtle improvements upon visual inspection.

\myparagraph{Patch size and sequence length.}
Much of previous work \cite{Chen:2018:R2D,Kim:2018:STT,Su:2017:DVD,Zhang:2019:ASL} is trained on random crops of size $128^2$, yet the significance of this choice is not further justified.
{\change%
In general, larger crops are beneficial as they reduce the influence of boundaries, given the typically big receptive fields.}
Here we explore additional patch sizes of $64^2, 96^2, 160^2$, and $192^2$, which we apply when training our pwc+cat model.
\Cref{tab:dbn-ablation}\red{(q -- t)} reveals that the choice of patch size -- when comparing to the baseline patch size of $128^2$ -- is quite important with a relative performance difference spanning from $-0.68$dB when using the smallest patch size $64^2$ to $+0.23$dB when using the largest ($192^2$).
While the performance difference between patch sizes is more significant for smaller absolute sizes, the performance gain from very large patches is still substantial.
This can also be seen in the visual results in \cref{fig:ablation-crops}.
Note the clearer poles.
Overall, the relative visual improvement becomes smaller with larger patch sizes, yet is still apparent.

\cite{Su:2017:DVD} proposed to use input sequences with $5$ images, which is kept in follow up work \cite{Zhang:2019:ASL,Kim:2018:STT}.
We include one more dimension in our case study, and test whether longer sequences can help.
In \cref{tab:dbn-ablation}\red{(u--v)}, we increased the number of input images to $7$ and retrained our pwc+cat model (with patch sizes $128^2$ and $192^2$).
The results reveal that $5$ input images largely suffice; two additional input images only yield a small benefit of $\sim 0.05$dB.

\verdict
Training patches should be chosen as big as the hardware limitations allow, since larger patch sizes provide clear benefits in accuracy.
Future GPUs may allow training at full resolution and improve results further.
Inputting more than $5$ images currently yields only minimal benefit.

\def\dbnsmall{{$\text{DBN}_{128,5}$}\@\xspace}
\def\flowdbnsmall{{$\text{FlowDBN}_{128,5}$}\@\xspace}
\def\flowdbnlarge{{$\text{FlowDBN}_{192,7}$}\@\xspace}
\def\dbnlarge{{$\text{DBN}_{192,1}$}\@\xspace}
\begin{table}[b]
\vspace{-0.6em}
\caption{%
Deblurring performance on the GOPRO dataset of \cite{Su:2017:DVD}.
}
\label{tab:gopro-su-short}
\vspace{0.5em}
%
\footnotesize
\centering
\begin{tabularx}{\columnwidth}{@{} l S[table-format=2.2,table-alignment=right] @{\hskip 1.735em} | @{\hskip 1.735em} l S[table-format=2.2,table-alignment=right] @{}}
\toprule
Method & PSNR & Method & PSNR \\
\midrule
R2D+DBN\footnotemark[1] \cite{Chen:2018:R2D} & 30.15 & ASL\footnotemark[2] \cite{Zhang:2019:ASL} & 29.10 \\
IFI-RNN\footnotemark[1] \cite{Nah:2019:RNN} & 30.80 & DBN\footnotemark[2] \cite{Su:2017:DVD} & 30.08  \\
R2D+DeblurGAN\footnotemark[1]\,\cite{Chen:2018:R2D} & 31.37 & $\text{DBN}_{128,5}$ (ours) & 31.48  \\
STT+DBN\footnotemark[1] \cite{Kim:2018:STT}& 31.61 &  $\text{FlowDBN}_{128,5}$ (ours) & 31.91  \\
OVD\footnotemark[1] \cite{Kim:2017:OVD,Kim:2018:STT} & 32.28  & $\text{FlowDBN}_{192,7}$ (ours) & 32.19 \\
STT+OVD\footnotemark[1] \cite{Kim:2018:STT} & \bfseries 32.53 &   \\
\bottomrule
\end{tabularx}
\rule{0pt}{2.5ex}
\footnotemark[1] {\footnotesize Results as reported.}
\footnotemark[2] {\footnotesize Results from a provided model.}
\end{table}

\section{Experiments}
\label{sec:experiments}
\paragraph{Evaluation on GOPRO by Su~\etal\cite{Su:2017:DVD}.}
As shown in the previous section, the proposed changes to Su's baseline strikingly boosted its deblurring accuracy by over 3dB compared to our basic baseline implementation.
We next consider how the improved baseline fares against the state-of-the-art.
We evaluate three variants:
Our best model without an optical flow backbone, trained under the same patch size ($128^2$) and sequence length (5) as competing methods (\cref{tab:dbn-ablation}\red{(h)}), denoted as \dbnsmall.
Our improved baseline, which includes optical flow pre-warping (\cref{tab:dbn-ablation}\red{(p)}), denoted as \flowdbnsmall.
And our best performing model trained under large patches and two more input images (\cref{tab:dbn-ablation}\red{(v)}), denoted as \flowdbnlarge.
\Cref{tab:gopro-su-short} shows the quantitative evaluation on the GOPRO testing dataset of \cite{Su:2017:DVD}.
Surprisingly, even our \dbnsmall model without optical flow already beats the highly competitive methods from Chen~\etal \cite{Chen:2018:R2D} by 0.11dB, which utilizes optical flow.
{\change
Our variants including optical flow, \flowdbnsmall and \flowdbnlarge are also highly competitive \wrt the recurrent approach of Nah~\etal \cite{Nah:2019:RNN} and the spatio-temporal transformer (STT) networks \cite{Kim:2018:STT}, \ie~\flowdbnsmall yields a higher average PSNR than STT applied to the same DBN backbone.
}
Finally, we improve the authors' results of \cite{Su:2017:DVD} by more than 2dB.
While our best performing \flowdbnlarge cannot quite reach the accuracy of methods based on the OVD backbone \cite{Kim:2017:OVD}, the OVD model exploits a dynamic temporal blending layer and uses recurrent predictions from previous iterations.
In contrast, our model is based on the conceptually simpler DBN, a plain feed-forward CNN.
{\change
We expect similar improvements when applying our insights in training details to the OVD backbone.
}

\begin{table*}[t]
\caption{%
Deblurring performance on the GOPRO dataset of \cite{Nah:2017:DMC} reported as PSNR \cite{Koehler:2012:RPC} / MSSIM \cite{Wang:2003:MSS}.
}
\label{tab:gopro-nah-detailed}
\vspace{0.5em}
%
\scriptsize
\centering
\def\myfirstcolsep{\hspace{1.275\tabcolsep}} 
\def\mycolsep{\hspace{1\tabcolsep}} 
\centering
\begin{tabular*}{1.0\textwidth}{@{}>{\columncolor{white}[0pt][\tabcolsep]} l @{\myfirstcolsep} c @{\mycolsep} c @{\mycolsep} c @{\mycolsep} c @{\mycolsep} c @{\mycolsep} c @{\mycolsep} c @{\mycolsep} c @{\mycolsep} c @{\mycolsep} c @{\mycolsep} c @{\mycolsep} >{\columncolor{white}[\tabcolsep][0pt]} r @{}}
\toprule
Method & \#1 & \#2 & \#3 & \#4 & \#5 & \#6 & \#7 & \#8 & \#9 & \#10 & \#11 & avg\hspace{1.42em} \\ \midrule
Reference Input & 28.77/.938 & 27.76/.941 & 26.58/.881 & 29.83/.976 & 26.50/.863 & 23.48/.802 & 23.05/.820 & 22.83/.816 & 25.03/.818 & 23.08/.791 & 25.90/.894 & 25.79/.868 \\
\rowcolor[gray]{.9}
DeblurGAN\footnotemark[1] \cite{Kupyn:2018:DGB} & 31.02/.968 &    30.37/.969 &    29.62/.938 &    31.04/.984 &    27.71/.906 &    25.41/.877 &    24.55/.879 &    25.24/.899 &    26.93/.891 &    25.64/.881 &    28.88/.942 &    27.92/.922  \\
DMC\footnotemark[1] \cite{Nah:2017:DMC} & 31.16/.965 &     30.94/.971 &     30.57/.945 &     31.16/.984 &     28.81/.922 &     26.27/.898 &     25.38/.902 &     26.24/.916 &     27.82/.911 &     26.67/.907 &     30.62/.958 &     28.77/.935  \\
\rowcolor[gray]{.9}
SRN+color\footnotemark[1] \cite{Tao:2018:SRN} & 32.82/.978 &     32.38/.980 &     32.21/.962 &     32.06/.988 &     29.86/.944 &     28.57/.946 &     27.81/.948 &     28.77/.958 &     29.65/.946 &     28.87/.948 &     32.71/.977 &     30.56/.961 \\
SRN+lstm\footnotemark[1] \cite{Tao:2018:SRN} & 32.82/.976 &     32.45/.980 &     32.25/.961 &     32.12/.988 &     29.82/.943 &     28.60/.947 &     27.60/.946 &     29.03/.962 &     29.76/.948 &     28.93/.949 &     32.83/.978 &     30.60/.962 \\
\rowcolor[gray]{.9}
$\text{DBN}_{192,1}$ (ours)          & 32.97/.978 &     32.51/.980 &     32.51/.964 &     32.17/.988 &     30.99/.955 &     28.81/.948 &     28.20/.955 &     28.88/.961 &     30.12/.953 &     29.17/.950 &     33.14/.978 &     30.92/.965 \\
%
$\text{FlowDBN}_{128,5}$ (ours)          & 33.22/.982 &     32.71/.982 &     32.61/.965 &     32.78/.990 &     30.92/.955 &     28.78/.949 &     28.48/.959 &     28.81/.963 &     30.26/.954 &     29.03/.950 &     33.10/.978 &     31.02/.966 \\
\rowcolor[gray]{.9}
$\text{FlowDBN}_{192,7}$ (ours)       & {\bf33.56}/{\bf.983} &     {\bf32.95}/{\bf.983} &     {\bf33.03}/{\bf.968} &     {\bf32.96}/{\bf.991} &     {\bf31.32}/\bf{.960} &     {\bf29.24}/{\bf.954} &     {\bf28.97}/{\bf.964} &     {\bf29.31}/{\bf.968} &     {\bf30.66}/\bf{.959} &    {\bf29.51}/\bf{.956} &     {\bf33.58}/{\bf.981} &    {\bf31.42}/{\bf.969} \\
\bottomrule
\end{tabular*}
%
%
\rule{0pt}{2.5ex}
\footnotemark[1] {\footnotesize Results from a provided model.}
%
\end{table*}
\begin{figure*}[t]
\centering
\subfigure[Input]{%
\label{fig:comparison-nah-input}%
\begin{minipage}{4.45cm}
\centering
\includegraphics[height=2.5cm]{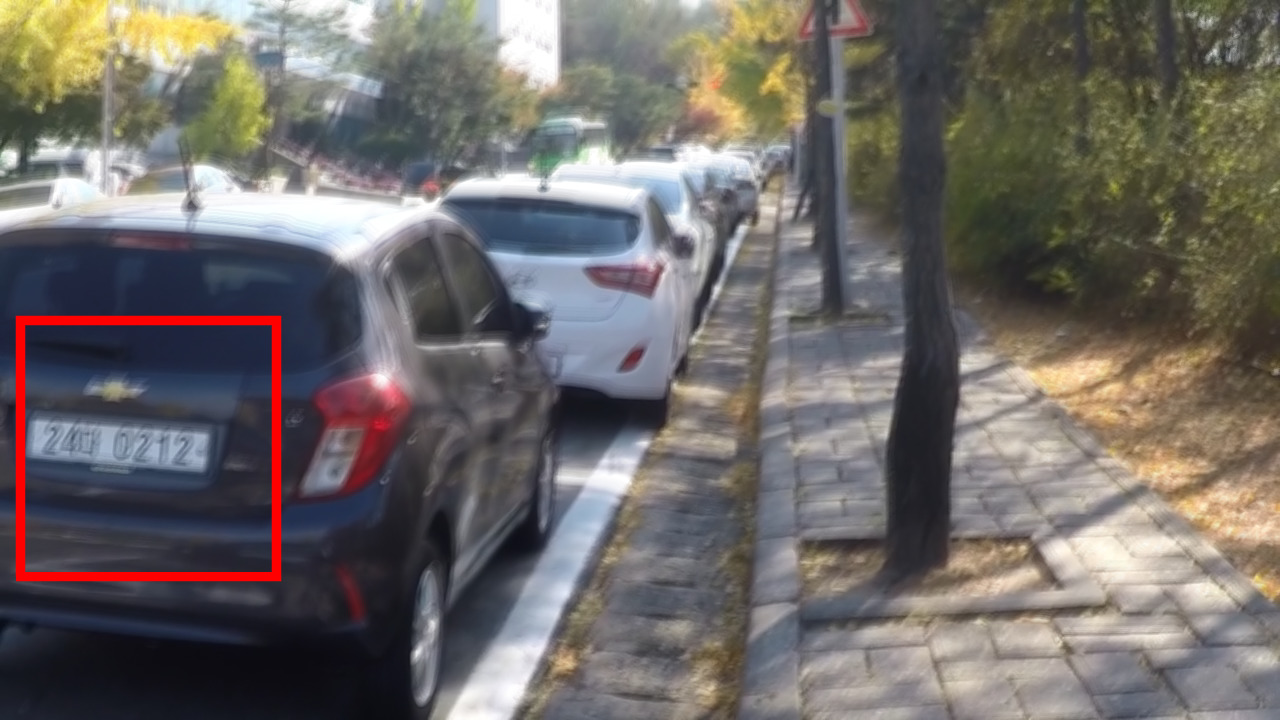} \\
\includegraphics[height=2.5cm]{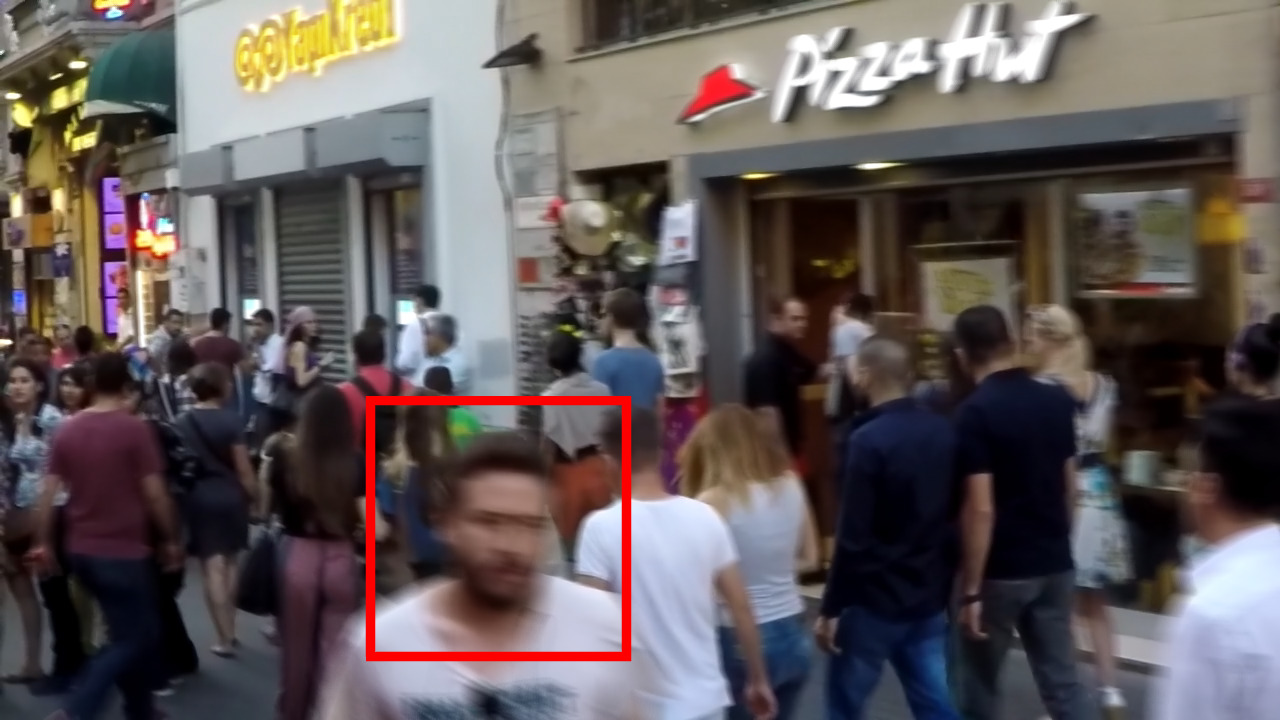} \\
\includegraphics[height=2.5cm]{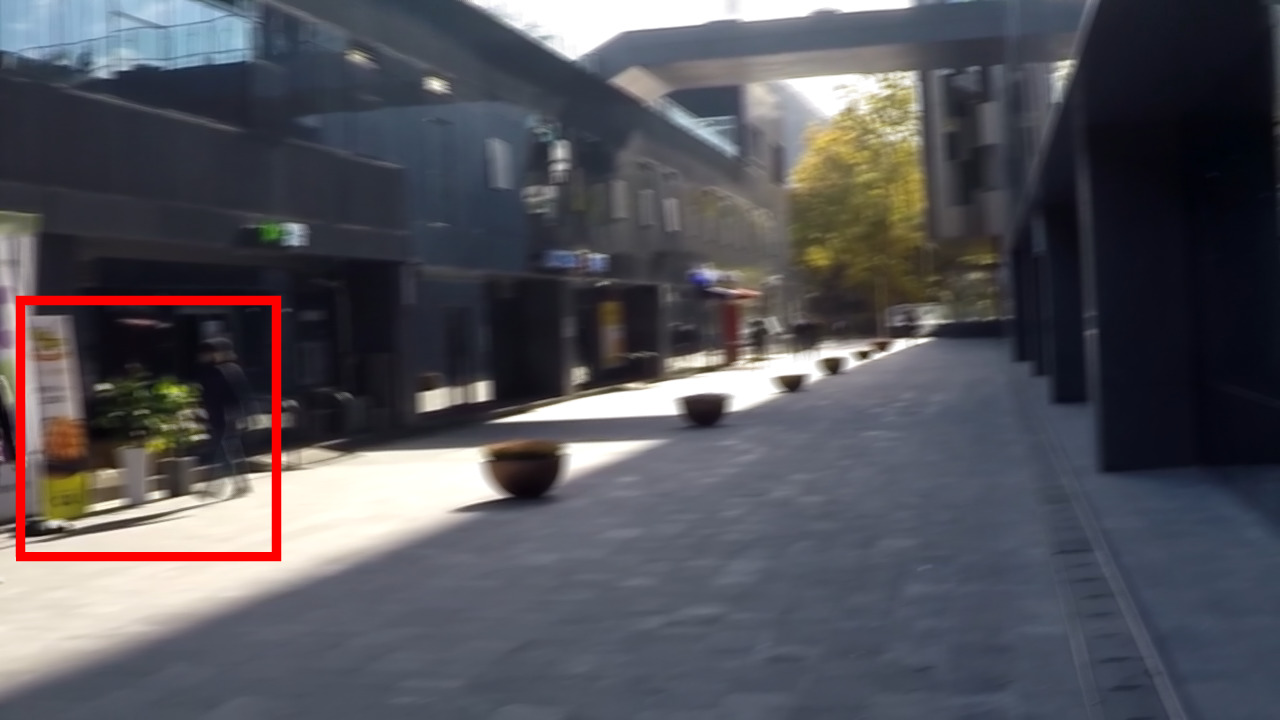} \\
\vspace{0.5em}
\end{minipage}}
\hspace{-0.8em}
\subfigure[DeblurGAN \cite{Kupyn:2018:DGB}]{
\label{fig:comparison-nah-deblurgan}
\begin{minipage}{2.5cm}
\centering
\includegraphics[height=2.5cm]{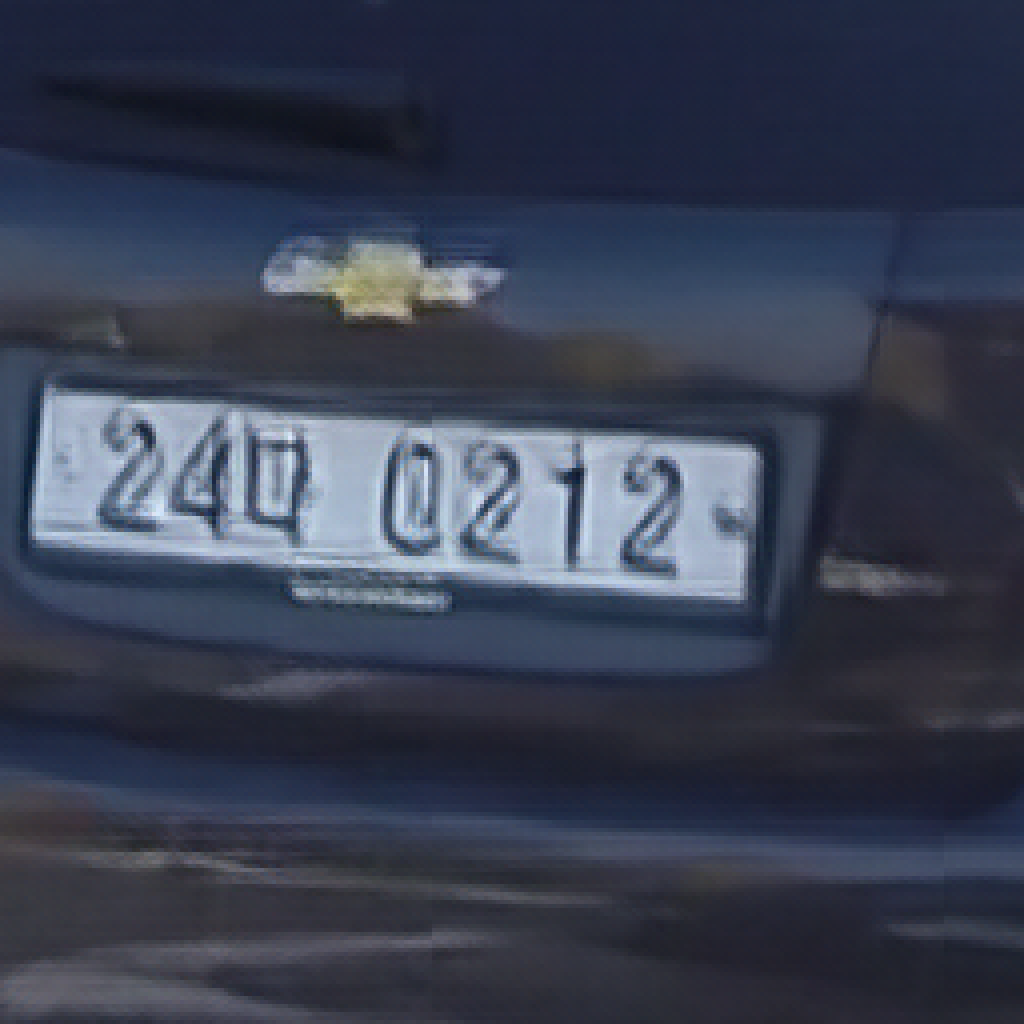} \\
\includegraphics[height=2.5cm]{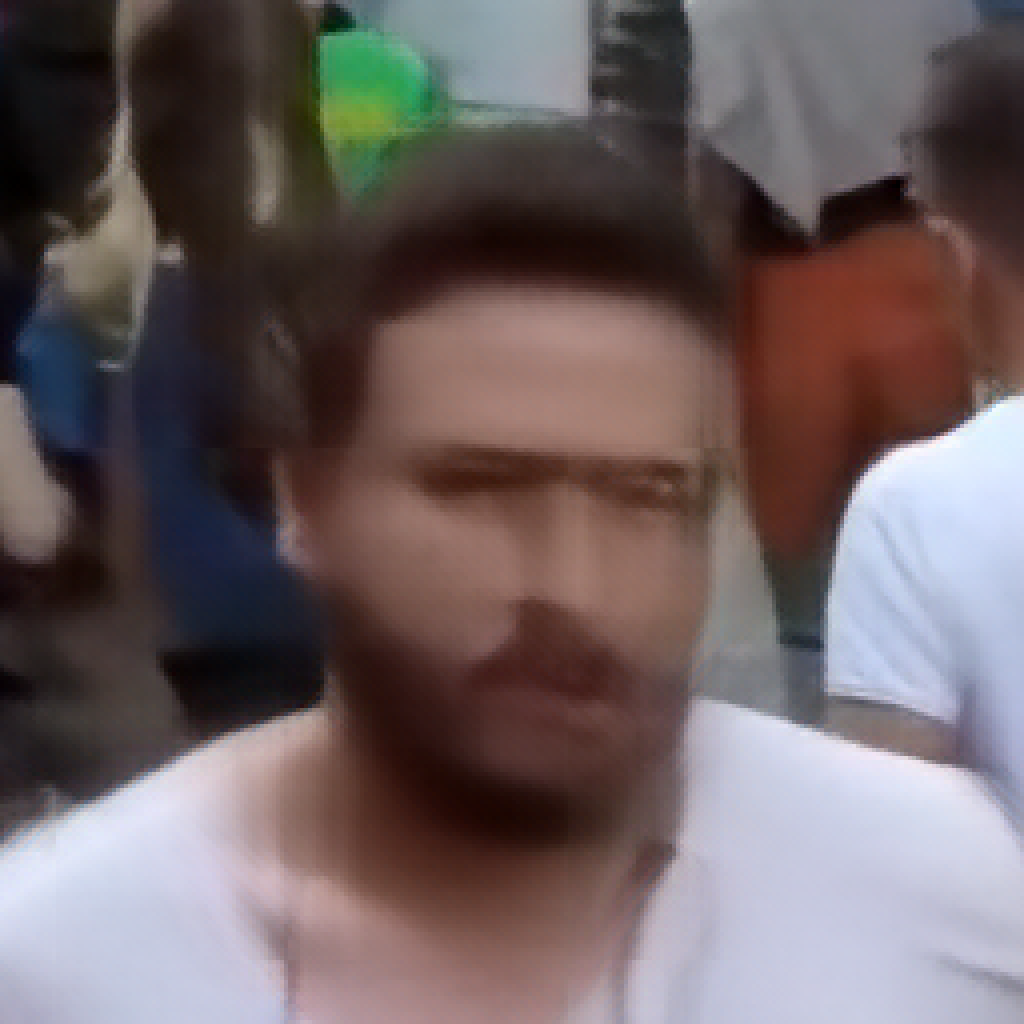} \\
\includegraphics[height=2.5cm]{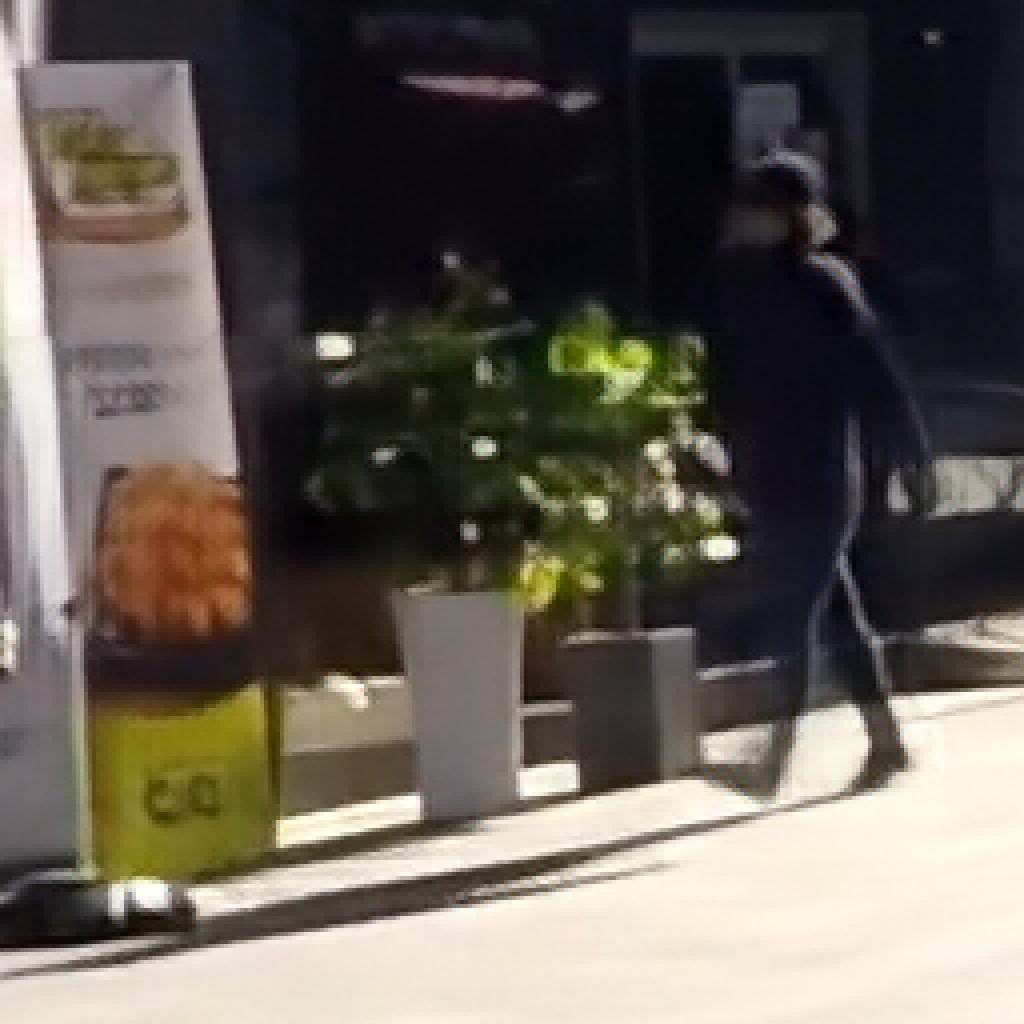} \\
\vspace{0.5em}
\end{minipage}}
\hspace{-0.8em}
\subfigure[DMC \cite{Nah:2017:DMC}]{
\label{fig:comparison-nah-dmc}
\begin{minipage}{2.5cm}
\centering
\includegraphics[height=2.5cm]{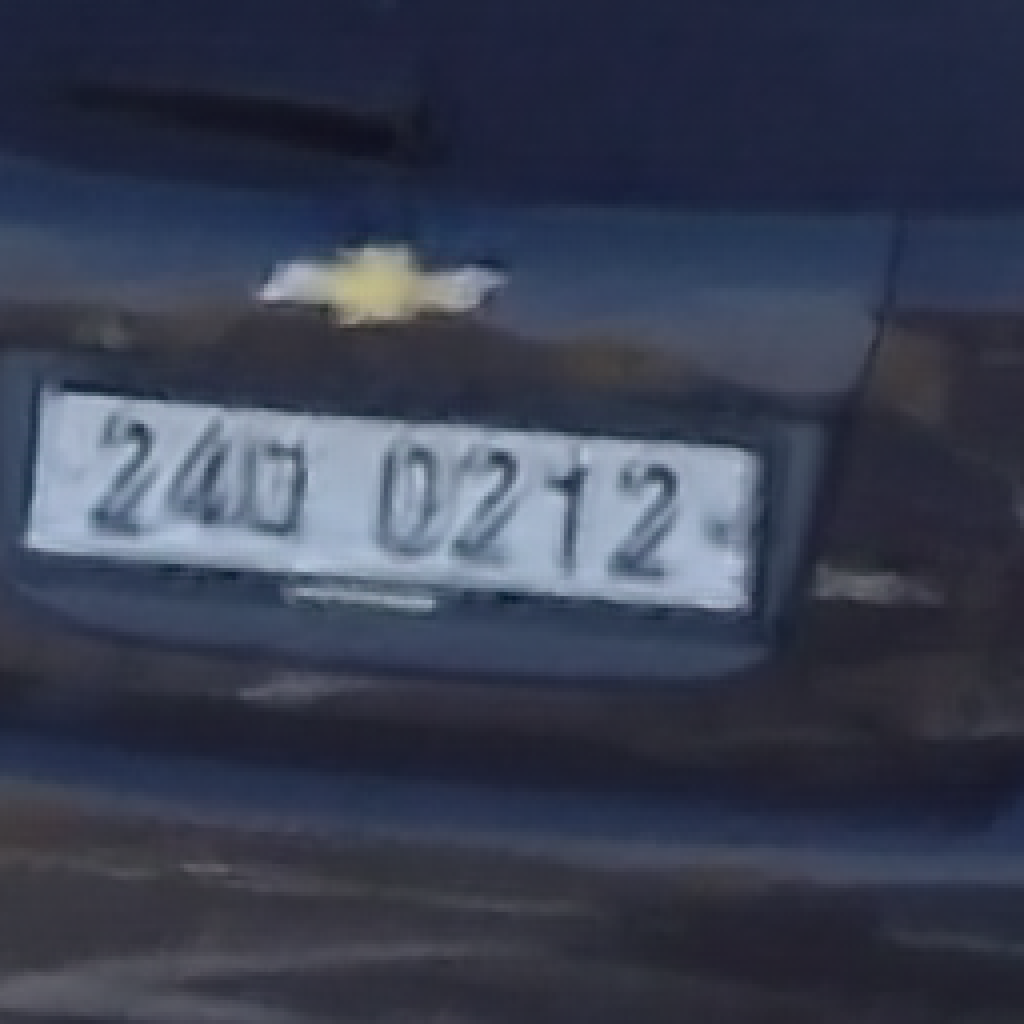} \\
\includegraphics[height=2.5cm]{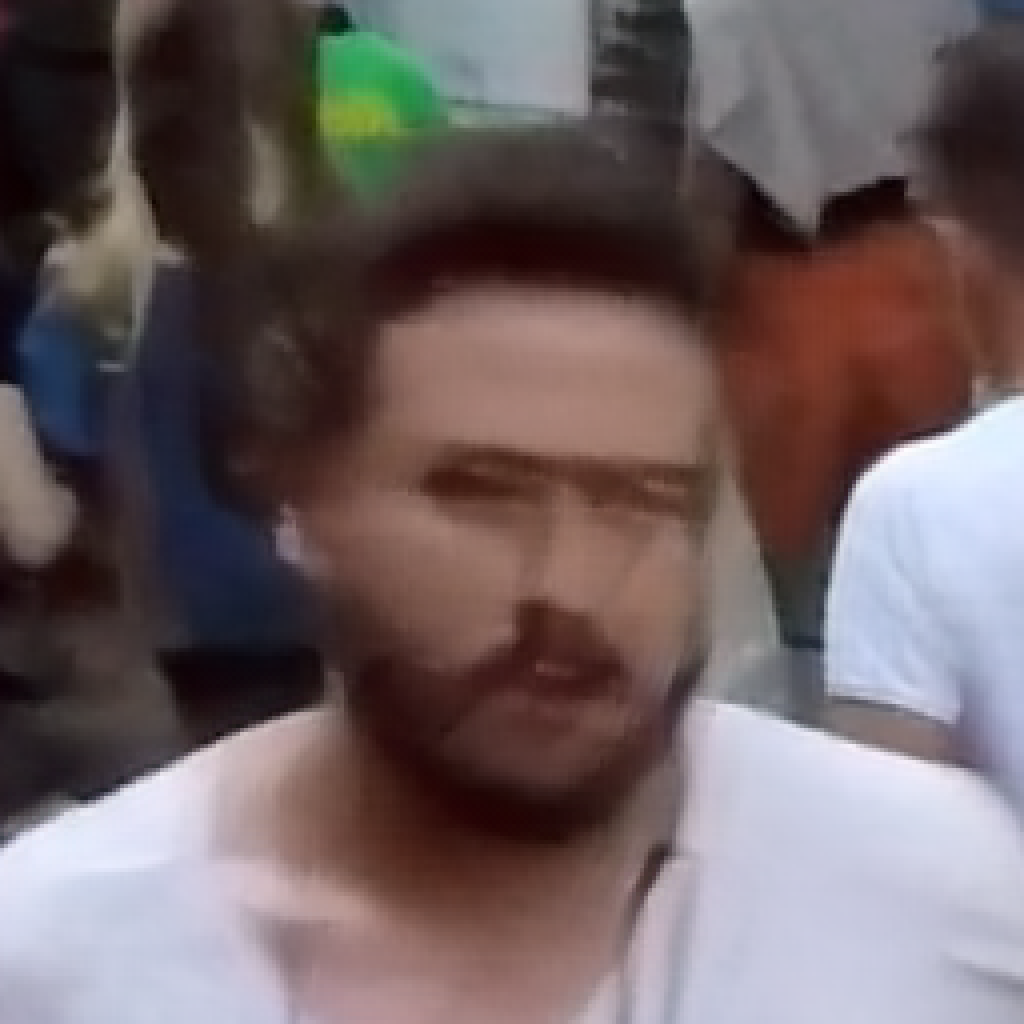} \\
\includegraphics[height=2.5cm]{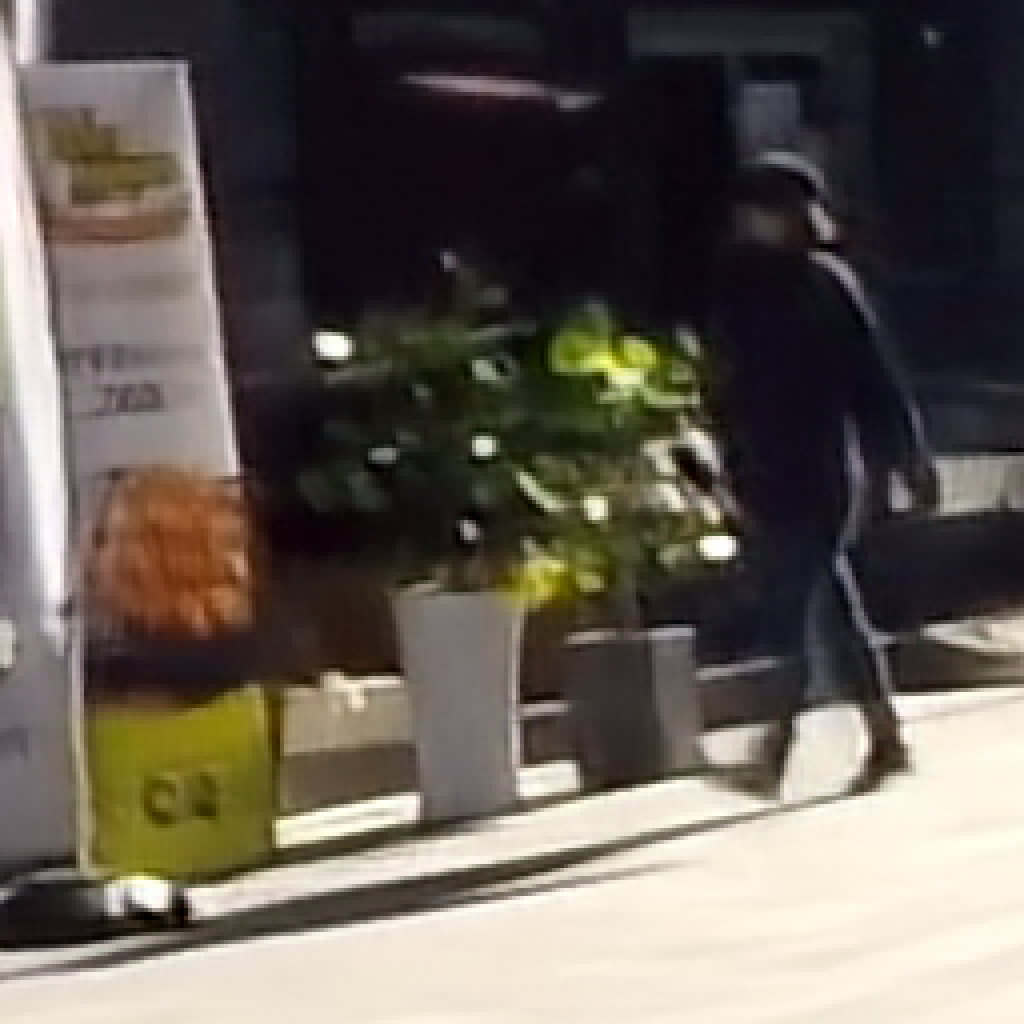} \\
\vspace{0.5em}
\end{minipage}}
\hspace{-0.8em}
\subfigure[SRN+lstm \cite{Tao:2018:SRN}]{
\label{fig:comparison-nah-srn}
\begin{minipage}{2.5cm}
\centering
\includegraphics[height=2.5cm]{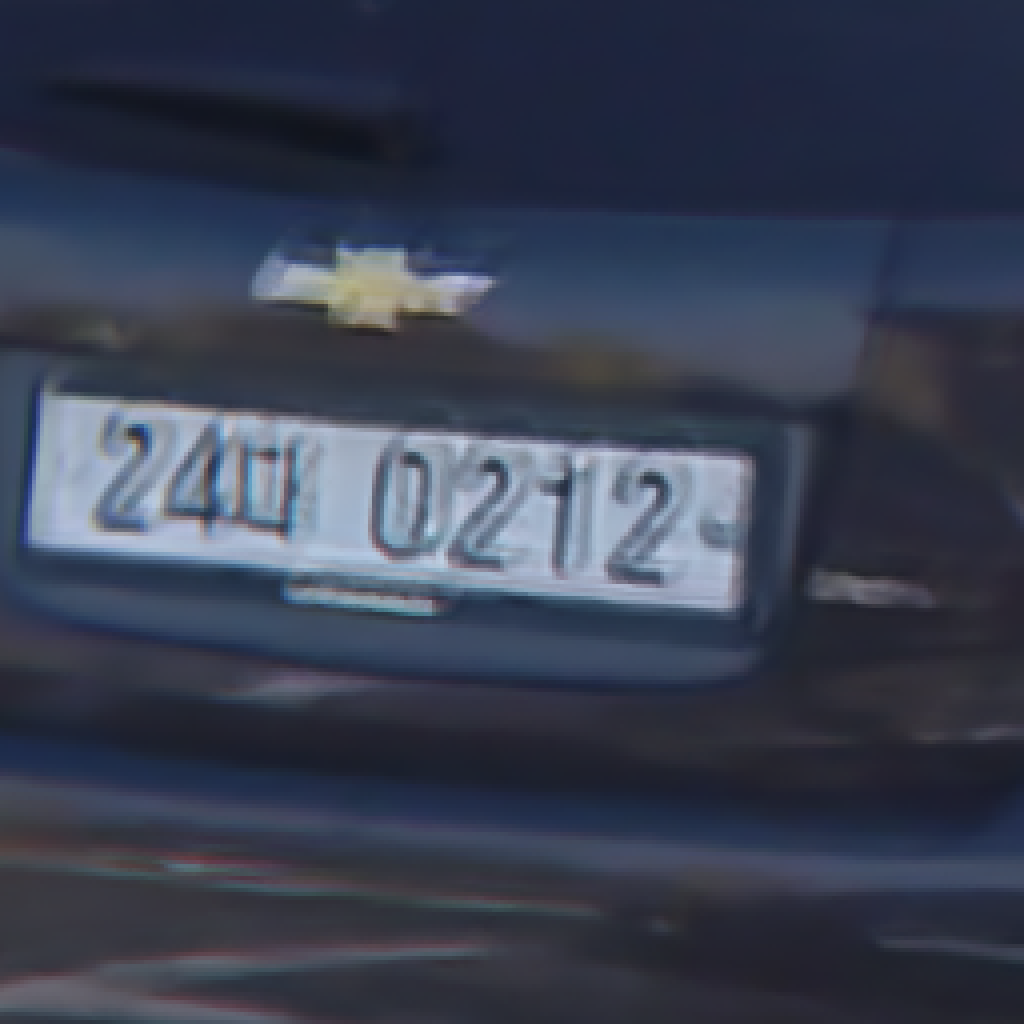} \\
\includegraphics[height=2.5cm]{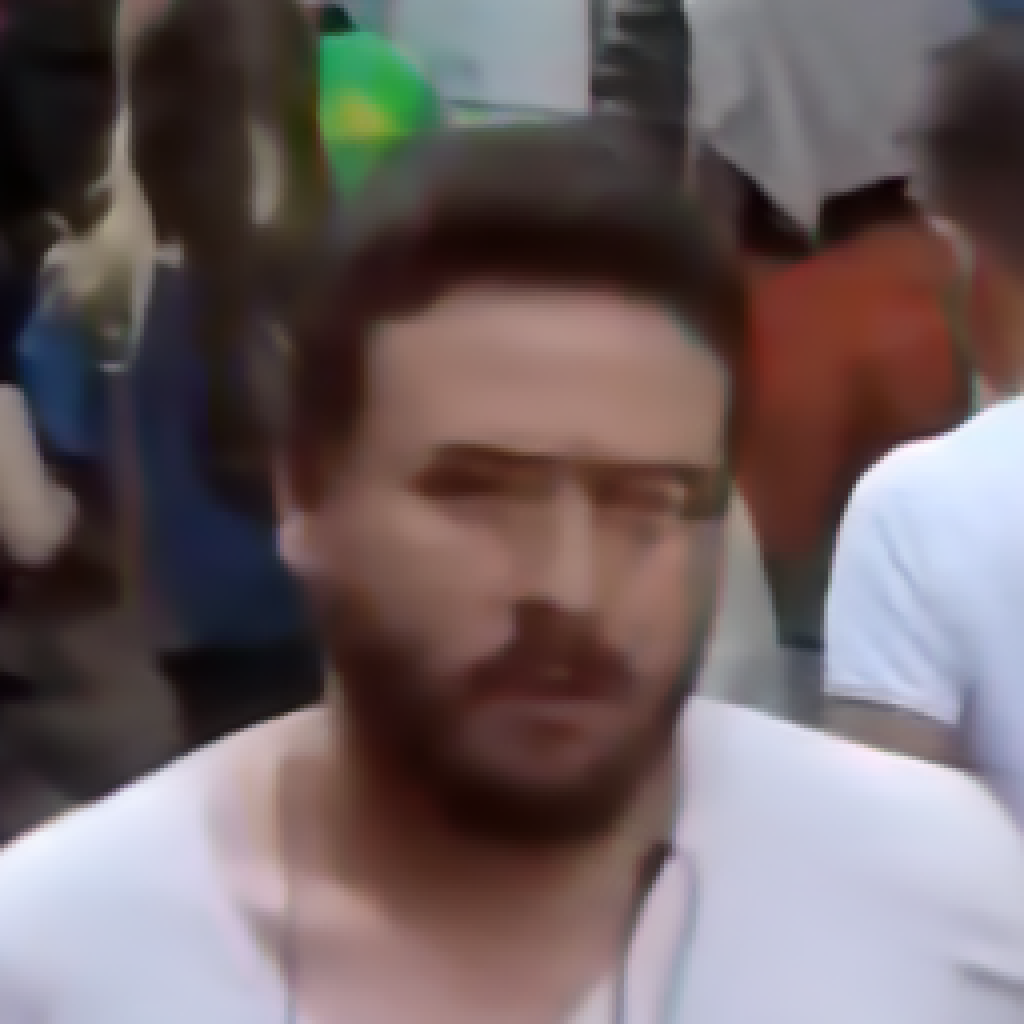} \\
\includegraphics[height=2.5cm]{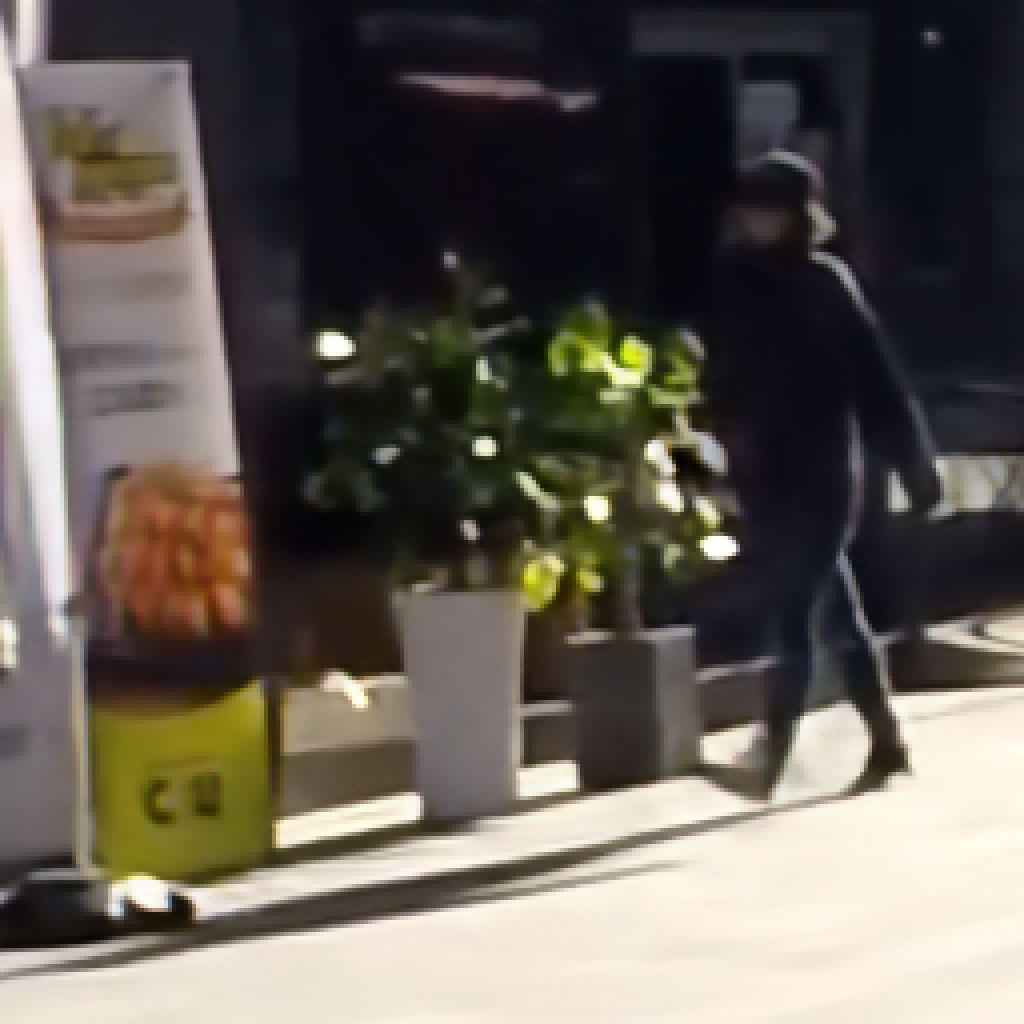} \\
\vspace{0.5em}
\end{minipage}}
\hspace{-0.8em}
\subfigure[$\text{FlowDBN}_{128,5}$]{
\label{fig:comparison-nah-flowdbn128}
\begin{minipage}{2.5cm}
\centering
\includegraphics[height=2.5cm]{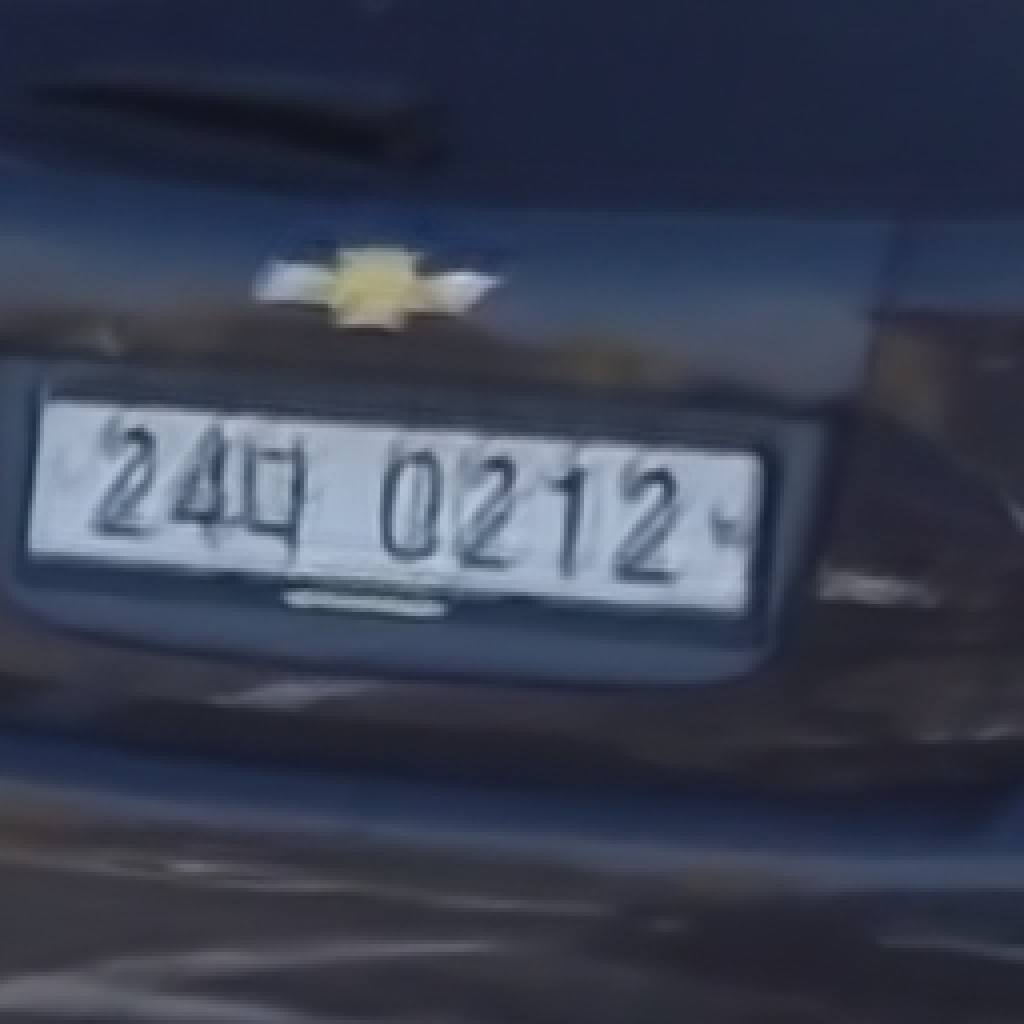} \\
\includegraphics[height=2.5cm]{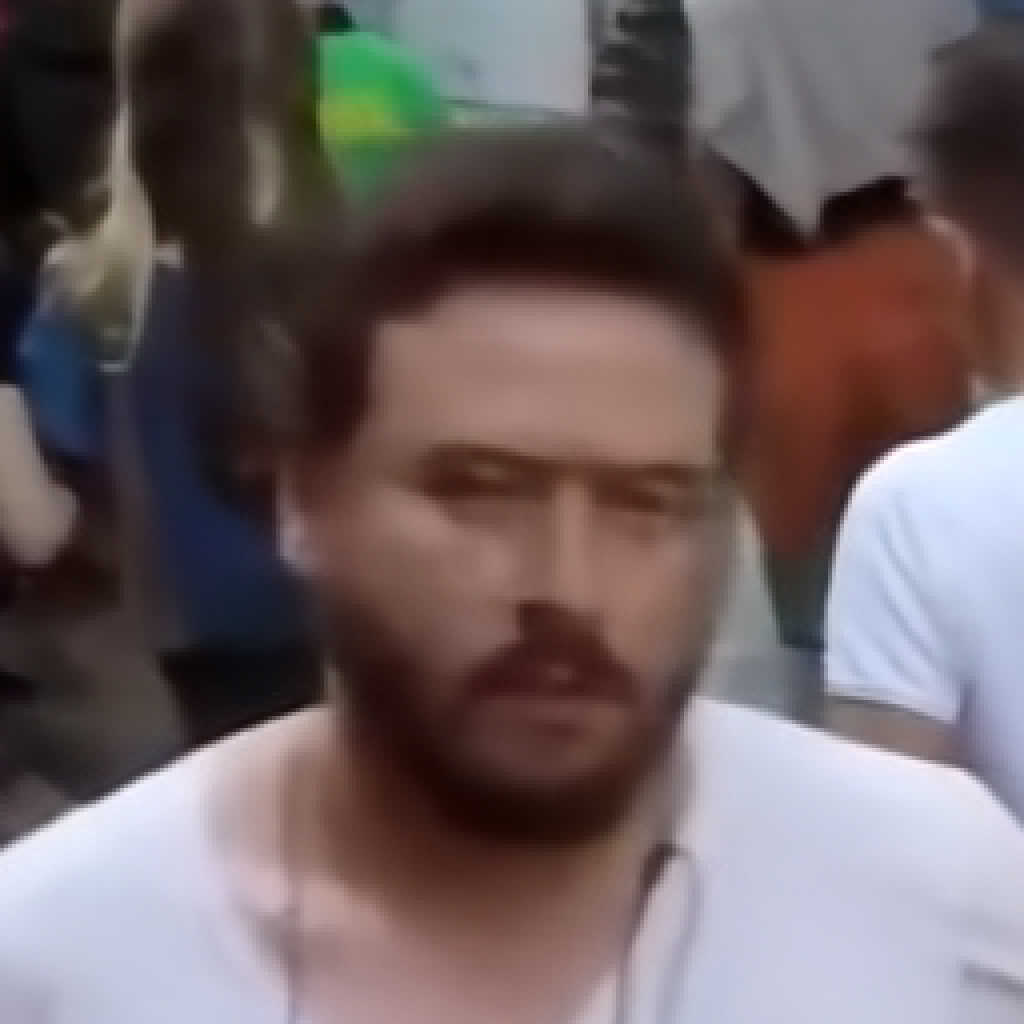} \\
\includegraphics[height=2.5cm]{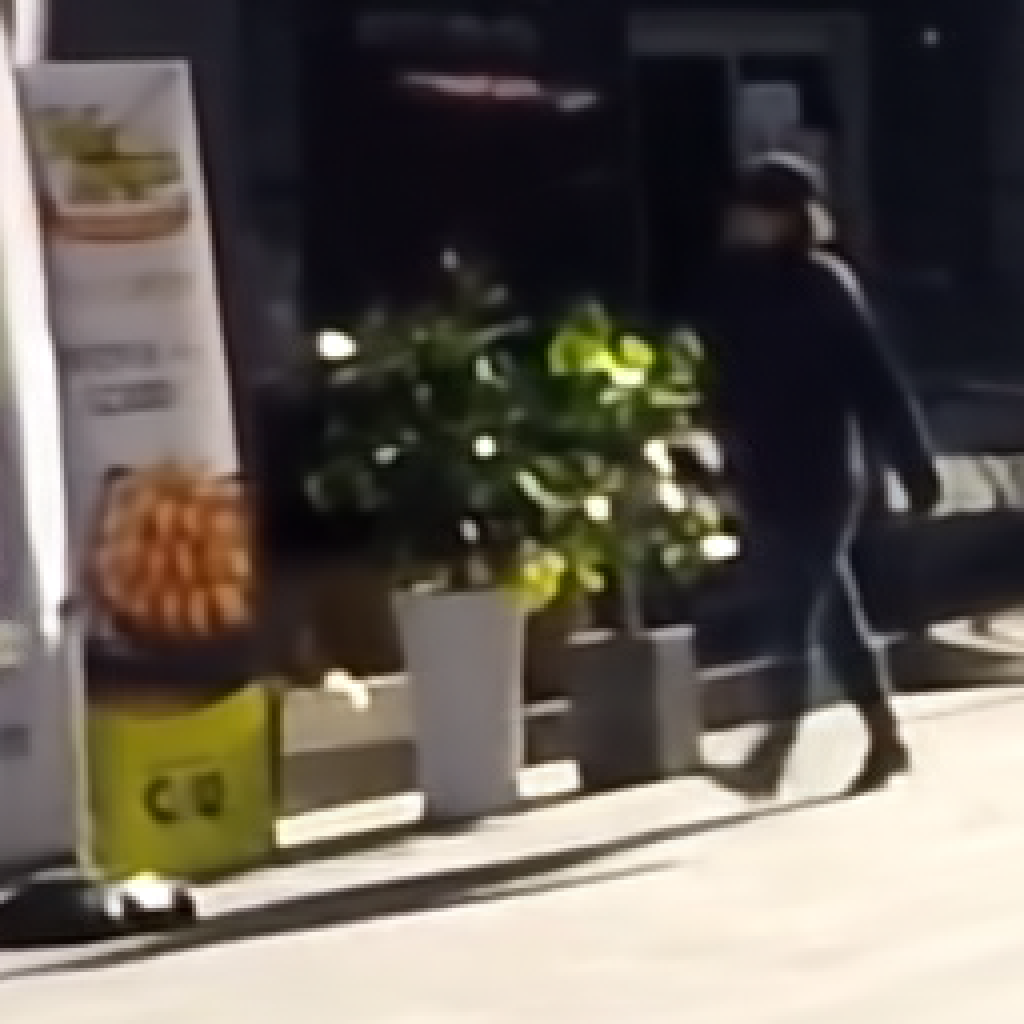} \\
\vspace{0.5em}
\end{minipage}}
\hspace{-0.8em}
\subfigure[$\text{FlowDBN}_{192,7}$]{
\label{fig:comparison-nah-flowdbn192}
\begin{minipage}{2.5cm}
\centering
\includegraphics[height=2.5cm]{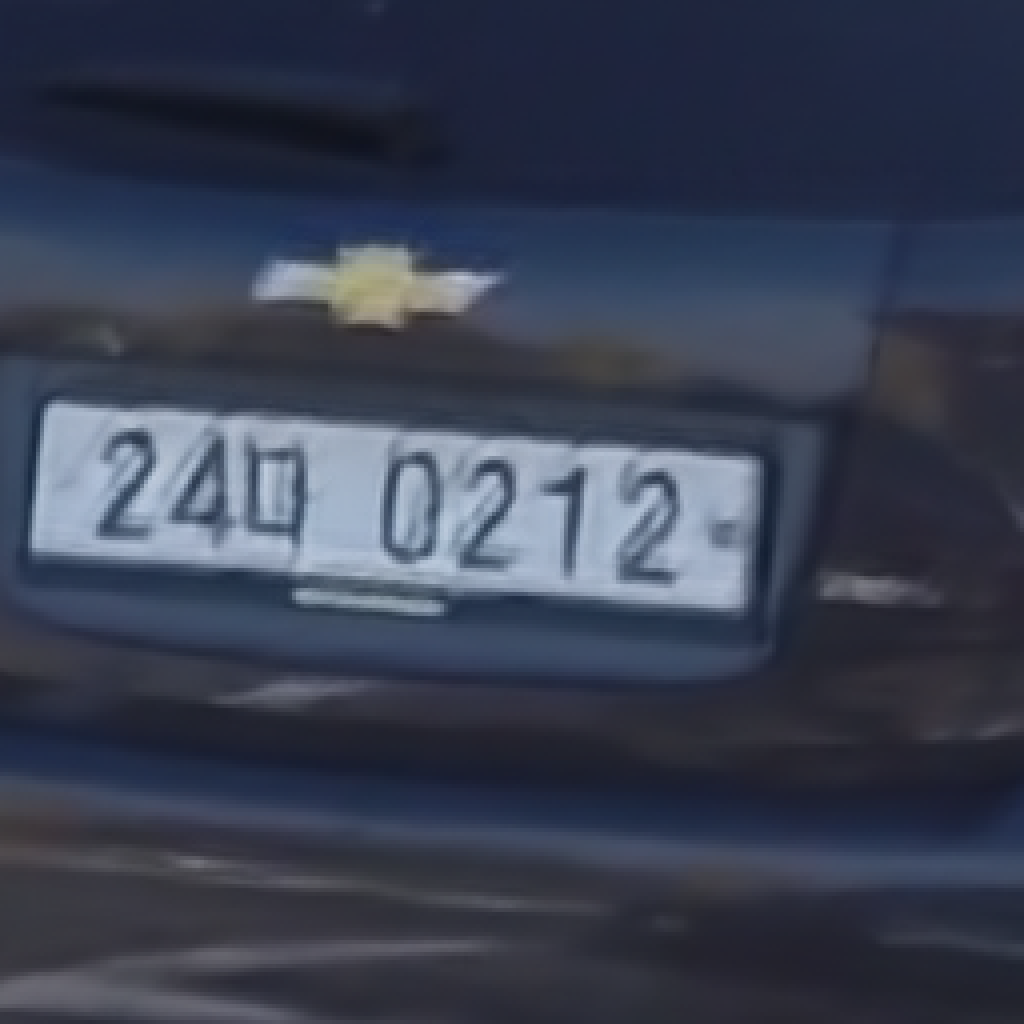} \\
\includegraphics[height=2.5cm]{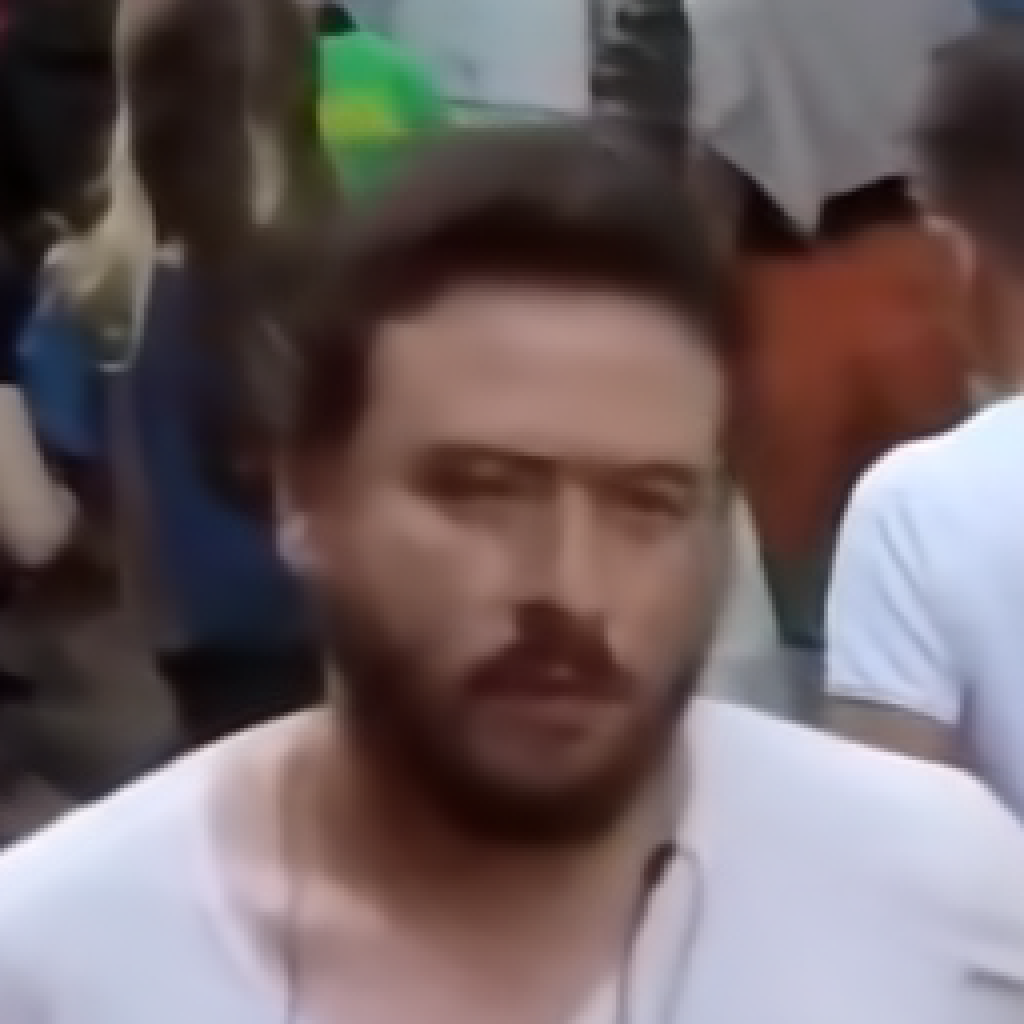} \\
\includegraphics[height=2.5cm]{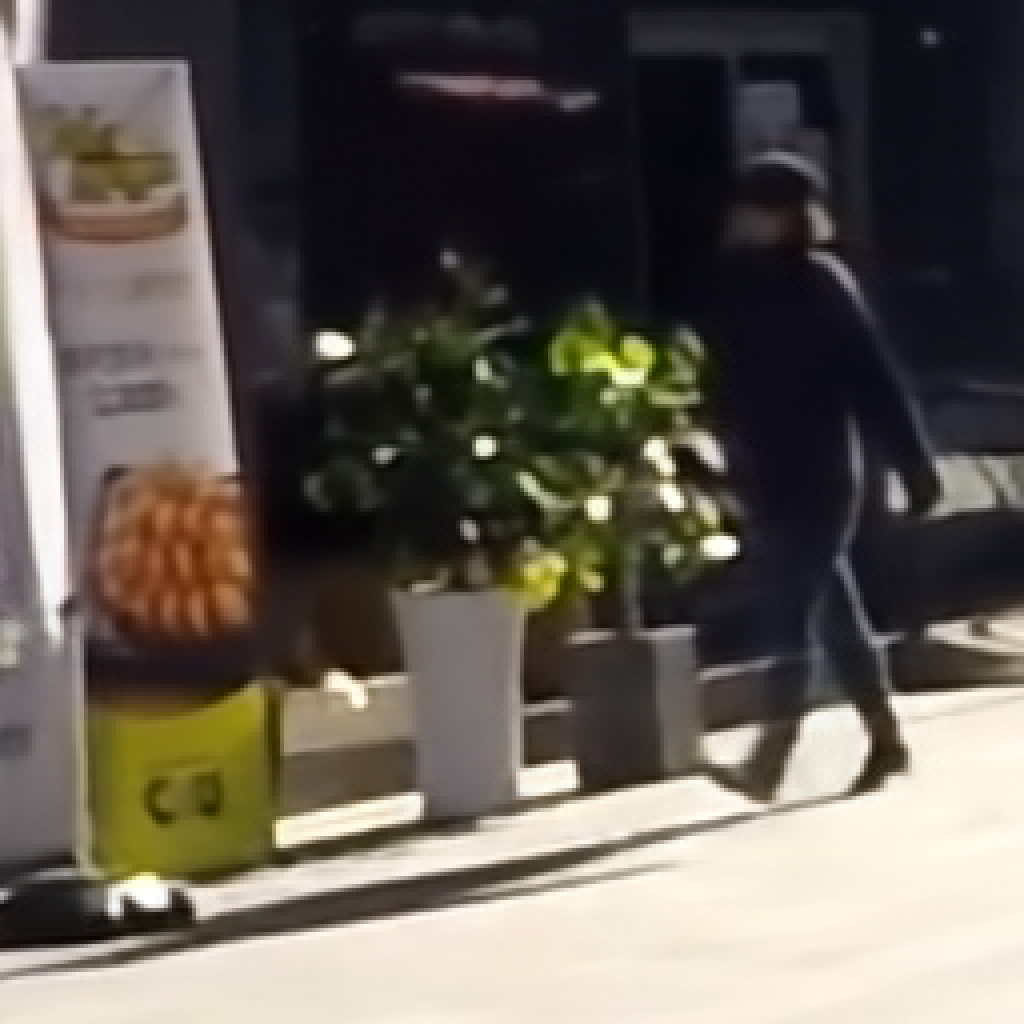} \\
\vspace{0.5em}
\end{minipage}}
\caption{%
{\bf Qualitative comparison.}
\subref{fig:comparison-nah-input} denotes the blurry input,
\subref{fig:comparison-nah-deblurgan} -- \subref{fig:comparison-nah-srn} competing methods.
Our FlowDBN models \subref{fig:comparison-nah-flowdbn128}, \subref{fig:comparison-nah-flowdbn192} exhibit clearer fonts in texts (\nth{1} row), fewer artifacts for small-scale details in face deblurring (\nth{2} row), and uncover more texture from blob-like structures (orange advertisement in the \nth{3} row).
}
\label{fig:comparison-nah}
%
\end{figure*}

\myparagraph{Evaluation on GOPRO by Nah~\etal \cite{Nah:2017:DMC}.}
To see whether the benefits we gain on our baseline generalize to other datasets, we also quantitatively evaluate on the GOPRO dataset of Nah~\etal \cite{Nah:2017:DMC}.
Note that the training set by \cite{Nah:2017:DMC} has roughly a third of the size of \cite{Su:2017:DVD}, hence our training schedule is three times as long, \ie 608 epochs and halving the learning rate at epochs $[308, 358, 408, 458, 508, 558]$.
The other details are as described in \cref{sec:baseline}.
We compare against DeblurGAN \cite{Kupyn:2018:DGB}, Nah~\etal's DMC baseline \cite{Nah:2017:DMC}, and the two highly competitive scale-recurrent models SRN+color/lstm by Tao~\etal \cite{Tao:2018:SRN}.
{\change
As these methods do not exploit multiple images, we additionally include \dbnlarge, a single-image variant of our baseline.}

The detailed results are shown in \cref{tab:gopro-nah-detailed}.
{\change
Interestingly, \dbnlarge already outperforms the highly competitive SRN+lstm model, a multiscale recurrent neural network, despite being trained on a smaller crop size (Tao~\etal \cite{Tao:2018:SRN} apply $256^2$ crops).
Both \flowdbnsmall and \flowdbnlarge perform even better, outperforming the best competing method by a very significant $\sim$0.8dB in PSNR.}

Qualitative results are shown in \cref{fig:comparison-nah}.
When inspecting the visual results, we find that both our FlowDBN models show perceptually better results, \eg they exhibit clearer text deblurring (\cf the plates in the \nth{1} row).
For moving people, faces can be problematic due to their small-scale details, as for instance shown in the results of the \nth{2} row, \ie  DeblurGAN, DMC, and SRN+LSTM all show artifacts in the face of the person.
While the results for both FlowDBN models are far from perfect, they show significantly fewer artifacts.
We observe another subtle improvement in blob-like structures such as the orange repetitive structure in the advertisement (last row).
{\change
Here, our FlowDBN models reconstruct a sharper texture than all competing methods.
}

\section{Conclusion}
\label{sec:conclusion}
In this paper we demonstrated how to create a highly competitive video deblurring model by revisiting details of an otherwise fairly standard CNN baseline architecture.
We show that despite a lot of effort being put into finding a good video deblurring architecture by the community, some benefits could possibly be even due to seemingly minor model and training details.
The resulting difference in terms of PSNR is surprisingly significant: In our study we improve the baseline network of \cite{Su:2017:DVD} by over 2dB compared to the original results in the paper, and 3.15dB over our initial implementation, which allows this simple network to outperform more recent and much more complex models.
This poses the question whether existing experimental comparisons in the deblurring literature actually uncover systematic accuracy differences from the architecture, or whether the differences may be down to detail engineering.
Future work thus needs to shed more light on this important point.

{\small
\balance
\bibliographystyle{ieee_fullname}
\bibliography{arxiv}

\begin{thebibliography}{10}\itemsep=-1pt

\bibitem{Chakrabarti:2016:ANA}
Ayan Chakrabarti.
\newblock A neural approach to blind motion deblurring.
\newblock In {\em ECCV}, volume~3, pages 221--235, 2016.

\bibitem{Chatfield:2014:RDD}
Ken Chatfield, Karen Simonyan, Andrea Vedaldi, and Andrew Zisserman.
\newblock Return of the devil in the details: {D}elving deep into convolutional
  nets.
\newblock In {\em BMVC}, 2014.

\bibitem{Chen:2018:R2D}
Huaijin Chen, Jinwei Gu, Orazio Gallo, Ming-Yu Liu, Ashok Veeraraghavan, and
  Jan Kautz.
\newblock {Reblur2Deblur}: {D}eblurring videos via self-supervised learning.
\newblock In {\em ICCP}, 2018.

\bibitem{Cho:2009:FMD}
Sunghyun Cho and Seungyong Lee.
\newblock Fast motion deblurring.
\newblock {\em ACM T. Graphics}, 28(5):145:1--145:8, Dec. 2009.

\bibitem{Cho:2012:VDH}
Sunghyun Cho, Jue Wang, and Seungyong Lee.
\newblock Video deblurring for hand-held cameras using patch-based synthesis.
\newblock {\em ACM T. Graphics}, 31(4):64:1--64:9, July 2012.

\bibitem{Couzinie:2013:LER}
Florent Couzini\'{e}-Devy, Jian Sun, Karteek Alahari, and Jean Ponce.
\newblock Learning to estimate and remove non-uniform image blur.
\newblock In {\em CVPR}, pages 1075--1082, 2013.

\bibitem{Delbracio:2015:HVD}
Mauricio Delbracio and Guillermo Sapiro.
\newblock Hand-held video deblurring via efficient {F}ourier aggregation.
\newblock {\em IEEE T. Comput. Imag.}, 1(4):270--283, Dec. 2015.

\bibitem{Dosovitskiy:2015:FN}
Alexey Dosovitskiy, Philipp Fischer, Eddy Ilg, Philip H{\"a}usser, Caner
  Haz{\i}rba\c{s}, Vladimir Golkov, Patrick van~der Smagt, Daniel Cremers, and
  Thomas Brox.
\newblock {FlowNet}: {L}earning optical flow with convolutional networks.
\newblock In {\em ICCV}, pages 2758--2766, 2015.

\bibitem{Fergus:2006:RCS}
Rob Fergus, Barun Singh, Aaron Hertzmann, Sam~T. Roweis, and William~T.
  Freeman.
\newblock Removing camera shake from a single photograph.
\newblock {\em ACM T. Graphics}, 25(3):787--794, July 2006.

\bibitem{Gast:2016:POM}
Jochen Gast, Anita Sellent, and Stefan Roth.
\newblock Parametric object motion from blur.
\newblock In {\em CVPR}, pages 1846--1854, 2016.

\bibitem{Gong:2017:FMB}
Dong Gong, Jie Yang, Lingqiao Liu, Yanning Zhang, Ian Reid, Chunhua Shen, Anton
  van~den Hengel, and Qinfeng Shi.
\newblock From motion blur to motion flow: {A} deep learning solution for
  removing heterogeneous motion blur.
\newblock In {\em CVPR}, pages 3806--3815, 2017.

\bibitem{Gupta:2010:SID}
Ankit Gupta, Neel Joshi, C.~Lawrence Zitnick, Michael Cohen, and Brian Curless.
\newblock Single image deblurring using motion density functions.
\newblock In {\em ECCV}, volume~1, pages 171--184, 2010.

\bibitem{Kaiming:2015:DDR}
Kaiming He, Xiangyu Zhang, Shaoqing Ren, and Jian Sun.
\newblock Delving deep into rectifiers: {S}urpassing human-level performance on
  {ImageNet} classification.
\newblock In {\em ICCV}, pages 1026--1034, 2015.

\bibitem{Hirsch:2010:FRN}
Michael Hirsch, Christian~J. Schuler, Stefan Harmeling, and Bernhard
  Sch{\"o}lkopf.
\newblock Fast removal of non-uniform camera shake.
\newblock In {\em ICCV}, pages 463--470, 2011.

\bibitem{Ilg:2017:FN2}
Eddy Ilg, Nikolaus Mayer, Tonmoy Saikia, Margret Keuper, Alexey Dosovitskiy,
  and Thomas Brox.
\newblock {FlowNet 2.0}: {E}volution of optical flow estimation with deep
  networks.
\newblock In {\em CVPR}, pages 1647--1655, 2017.

\bibitem{Ioffe:2015:BNA}
Sergey Ioffe and Christian Szegedy.
\newblock Batch normalization: {A}ccelerating deep network training by reducing
  internal covariate shift.
\newblock In {\em ICML}, pages 448--456, 2015.

\bibitem{Jin:2018:NBD}
Meiguang Jin, Stefan Roth, and Paolo Favaro.
\newblock Normalized blind deconvolution.
\newblock In {\em ECCV}, volume~7, pages 694--711, 2018.

\bibitem{Kim:2013:DSD}
Tae~Hyun Kim, Byeongjoo Ahn, and Kyoung~Mu Lee.
\newblock Dynamic scene deblurring.
\newblock In {\em ICCV}, pages 3160--3167, 2013.

\bibitem{Kim:2014:SFD}
Tae~Hyun Kim and Kyoung~Mu Lee.
\newblock Segmentation-free dynamic scene deblurring.
\newblock In {\em CVPR}, pages 2766--2773, 2014.

\bibitem{Kim:2015:GVD}
Tae~Hyun Kim and Kyoung~Mu Lee.
\newblock Generalized video deblurring for dynamic scenes.
\newblock In {\em CVPR}, pages 5426--5434, 2015.

\bibitem{Kim:2017:OVD}
Tae~Hyun Kim, Kyoung~Mu Lee, Bernhard Sch{\"o}lkopf, and Michael Hirsch.
\newblock Online video deblurring via dynamic temporal blending network.
\newblock In {\em ICCV}, pages 4058--4067, 2017.

\bibitem{Kim:2018:STT}
Tae~Hyun Kim, Mehdi S.~M. Sajjadi, Michael Hirsch, and Bernhard Sch{\"o}lkopf.
\newblock Spatio-temporal transformer network for video restoration.
\newblock In {\em ECCV}, volume~3, pages 111--127, 2018.

\bibitem{Kingma:2015:AAM}
Diederik~P. Kingma and Jimmy~Lei Ba.
\newblock Adam: {A} method for stochastic optimization.
\newblock In {\em ICLR}, 2015.

\bibitem{Koehler:2012:RPC}
Rolf K{\"o}hler, Michael Hirsch, Betty Mohler, Bernhard Sch{\"o}lkopf, and
  Stefan Harmeling.
\newblock Recording and playback of camera shake: {B}enchmarking blind
  deconvolution with a real-world database.
\newblock In {\em ECCV}, volume~7, pages 27--40, 2012.

\bibitem{Krishnan:2011:BDN}
Dilip Krishnan, Terence Tay, and Rob Fergus.
\newblock Blind deconvolution using a normalized sparsity measure.
\newblock In {\em CVPR}, pages 233--240, 2011.

\bibitem{Kupyn:2018:DGB}
Orest Kupyn, Volodymyr Budzan, Mykola Mykhailych, Dmytro Mishkin, and Ji{\v r}i
  Matas.
\newblock {DeblurGAN}: {B}lind motion deblurring using conditional adversarial
  networks.
\newblock In {\em CVPR}, pages 8183--8192, 2018.

\bibitem{Levin:2006:BMD}
Anat Levin.
\newblock Blind motion deblurring using image statistics.
\newblock In {\em NIPS*2006}, pages 841--848.

\bibitem{Levin:2009:UEB}
Anat Levin, Yair Weiss, Fredo Durand, and William~T. Freeman.
\newblock Understanding and evaluating blind deconvolution algorithms.
\newblock In {\em CVPR}, pages 1964--1971, 2009.

\bibitem{Levin:2011:EML}
Anat Levin, Yair Weiss, Fredo Durand, and William~T. Freeman.
\newblock Efficient marginal likelihood optimization in blind deconvolution.
\newblock In {\em CVPR}, pages 2657--2664, 2011.

\bibitem{Lucic:2018:AGC}
Mario Lucic, Karol Kurach, Marcin Michalski, Olivier Bousquet, and Sylvain
  Gelly.
\newblock Are {GAN}s created equal? {A} large-scale study.
\newblock In {\em NeurIPS*2018}, pages 700--709.

\bibitem{Matsushita:2006:FVS}
Yasuyuki Matsushita, Eyal Ofek, Weina Ge, Xiaoou Tang, and Heung-Yeung Shum.
\newblock Full-frame video stabilization with motion inpainting.
\newblock {\em IEEE T. Pattern Anal. Mach. Intell.}, 28(7):1150--1163, July
  2006.

\bibitem{Michaeli:2014:BDI}
Tomer Michaeli and Michal Irani.
\newblock Blind deblurring using internal patch recurrence.
\newblock In {\em ECCV}, volume~3, pages 783--798, 2014.

\bibitem{Miskin:2000:ELB}
James Miskin and David J.~C. MacKay.
\newblock Ensemble learning for blind image separation and deconvolution.
\newblock In Mark Girolami, editor, {\em Advances in Independent Component
  Analysis}, Perspectives in Neural Computing, chapter~7, pages 123--141.
  Springer London, 2000.

\bibitem{Nah:2017:DMC}
Seungjun Nah, Tae~Hyun Kim, and Kyoung~Mu Lee.
\newblock Deep multi-scale convolutional neural network for dynamic scene
  deblurring.
\newblock In {\em CVPR}, pages 257--265, 2017.

\bibitem{Nah:2019:RNN}
Seungjun Nah, Sanghyun Son, and Kyoung~Mu Lee.
\newblock Recurrent neural networks with intra-frame iterations for video
  deblurring.
\newblock In {\em CVPR}, pages 8102--8111, 2019.

\bibitem{Noroozi:2017:MDW}
Mehdi Noroozi, Paramanand Chandramouli, and Paolo Favaro.
\newblock Motion deblurring in the wild.
\newblock In {\em GCPR}, pages 65--77, 2017.

\bibitem{Pan:2016:BID}
Jinshan Pan, Deqing Sun, Hanspeter Pfister, and Ming-Hsuan Yang.
\newblock Blind image deblurring using dark channel prior.
\newblock In {\em CVPR}, pages 1628--1636, 2016.

\bibitem{Pascanu:2013:OTT}
Razvan Pascanu, Tomas Mikolov, and Yoshua Bengio.
\newblock On the difficulty of training recurrent neural networks.
\newblock In {\em ICML}, pages 1310--1318, 2013.

\bibitem{Perez:2013:TVL}
Javier~S{\'a}nchez P{\'e}rez, Enric Meinhardt-Llopis, and Gabriele Facciolo.
\newblock {TV}-{L1} optical flow estimation.
\newblock {\em Image Process. On Line}, 3:137--150, 2013.

\bibitem{Perrone:2014:TVB}
Daniele Perrone and Paolo Favaro.
\newblock Total variation blind deconvolution: {T}he devil is in the details.
\newblock In {\em CVPR}, pages 2909--2916, 2014.

\bibitem{Ramakrishnan:2017:DGF}
Sainandan Ramakrishnan, Shubham Pachori, Aalok Gangopadhyay, and Shanmuganathan
  Raman.
\newblock Deep generative filter for motion deblurring.
\newblock In {\em ICCV Workshops}, pages 2993--3000, 2017.

\bibitem{Ren:2017:VDS}
Wenqi Ren, Jinshan Pan, Xiaochun Cao, and Ming-Hsuan Yang.
\newblock Video deblurring via semantic segmentation and pixel-wise non-linear
  kernel.
\newblock In {\em ICCV}, pages 1086--1094, 2017.

\bibitem{Schelten:2015:IRT}
Kevin Schelten, Sebastian Nowozin, Jeremy Jancsary, Carsten Rother, and Stefan
  Roth.
\newblock Interleaved regression tree field cascades for blind image
  deconvolution.
\newblock In {\em WACV}, pages 494--501, 2015.

\bibitem{Schuler:2016:LTD}
Christian~J. Schuler, Michael Hirsch, Stefan Harmeling, and Bernhard
  Sch{\"o}lkopf.
\newblock Learning to deblur.
\newblock {\em IEEE T. Pattern Anal. Mach. Intell.}, 38(7):1439--1451, July
  2016.

\bibitem{Shan:2008:HQM}
Qi Shan, Jiaya Jia, and Aseem Agarwala.
\newblock High-quality motion deblurring from a single image.
\newblock {\em ACM T. Graphics}, 27(3):73:1--73:10, Aug. 2008.

\bibitem{Su:2017:DVD}
Shuochen Su, Mauricio Delbracio, Jue Wang, Guillermo Sapiro, Wolfgang Heidrich,
  and Oliver Wang.
\newblock Deep video deblurring for hand-held cameras.
\newblock In {\em CVPR}, pages 237--246, 2017.

\bibitem{Sun:2018:PWC}
Deqing Sun, Xiaodong Yang, Ming-Yu Liu, and Jan Kautz.
\newblock {PWC-Net}: {CNN}s for optical flow using pyramid, warping, and cost
  volume.
\newblock In {\em CVPR}, pages 8934--8943, 2018.

\bibitem{Sun:2019:MMT}
Deqing Sun, Xiaodong Yang, Ming-Yu Liu, and Jan Kautz.
\newblock Models matter, so does training: {A}n empirical study of {CNN}s for
  optical flow estimation.
\newblock {\em IEEE T. Pattern Anal. Mach. Intell.}, 2019, to appear.

\bibitem{Sun:2015:LCN}
Jian Sun, Wenfei Cao, Zongben Xu, and Jean Ponce.
\newblock Learning a convolutional neural network for non-uniform motion blur
  removal.
\newblock In {\em CVPR}, pages 769--777, 2015.

\bibitem{Sun:2013:EBK}
Libin Sun, Sunghyun Cho, Jue Wang, and James Hays.
\newblock Edge-based blur kernel estimation using patch priors.
\newblock In {\em ICCP}, 2013.

\bibitem{Tao:2018:SRN}
Xin Tao, Hongyun Gao, Xiaoyong Shen, Jue Wang, and Jiaya Jia.
\newblock Scale-recurrent network for deep image deblurring.
\newblock In {\em CVPR}, pages 8174--8182, 2018.

\bibitem{Wang:2019:VRE}
Xintao Wang, Kelvin C.~K. Chan, Ke Yu, Chao Dong, and Chen~Change Loy.
\newblock {EDVR}: {V}ideo restoration with enhanced deformable convolutional
  networks.
\newblock In {\em CVPR Workshops}, 2019.

\bibitem{Wang:2003:MSS}
Zhou Wang, Eero~P. Simoncelli, and Alan~C. Bovik.
\newblock Multi-scale structural similarity for image quality assessment.
\newblock In {\em ACSSC}, pages 1398--1402, 2003.

\bibitem{Whyte:2010:NDS}
Oliver Whyte, Josef Sivic, Andrew Zisserman, and Jean Ponce.
\newblock Non-uniform deblurring for shaken images.
\newblock In {\em CVPR}, pages 491--498, 2010.

\bibitem{Xu:2010:TPK}
Li Xu and Jiaya Jia.
\newblock Two-phase kernel estimation for robust motion deblurring.
\newblock In {\em ECCV}, volume~1, pages 157--170, 2010.

\bibitem{Xu:2013:ULS}
Li Xu, Shicheng Zheng, and Jiaya Jia.
\newblock Unnatural {$L_0$} sparse representation for natural image deblurring.
\newblock In {\em CVPR}, pages 1107--1114, 2013.

\bibitem{Zhang:2019:ASL}
Kaihao Zhang, Wenhan Luo, Yiran Zhong, Lin Ma, Wei Liu, and Hongdong Li.
\newblock Adversarial spatio-temporal learning for video deblurring.
\newblock {\em IEEE T. Image Process.}, 28(1):291--301, Jan. 2019.

\bibitem{Zheng:2013:FMD}
Shicheng Zheng, Li Xu, and Jiaya Jia.
\newblock Forward motion deblurring.
\newblock In {\em ICCV}, pages 1465--1472, 2013.

\end{thebibliography}
}

\end{document}